\documentclass[]{wan}

\usepackage[utf8]{inputenc}
\usepackage{url}
\usepackage{amsmath}
\usepackage{amssymb}
\usepackage{amsfonts}
\usepackage{amsthm}
\usepackage{nicefrac}
\usepackage{colortbl}
\usepackage{array}
\usepackage{enumitem}
\usepackage{float}
\usepackage{wrapfig}
\usepackage{tikz}
\usetikzlibrary{arrows.meta,positioning,shapes.geometric,fit,calc,backgrounds,decorations.pathreplacing}
\usepackage{algorithm}
\usepackage{algpseudocode}

\tikzset{every picture/.append style={font=\rmfamily}}

\providecommand{\sd}[1]{\,{\scriptsize$\pm$#1}}

\graphicspath{{figs/}{new_figs/}}

\title{ReCA: Multi-Shot Long Video Extrapolation via Recursive Context Allocation}

\author[1,2,3,\ddag]{Akide Liu}
\author[2,\dag]{Jinbo Xing}
\author[2]{Chaojie Mao}
\author[3]{Ye Li}
\author[3]{Zeyu Zhang}
\author[3]{Yefei He}
\author[3]{Weijie Wang}
\author[4]{Zihan Wang}
\author[2]{Yu Liu}
\author[1]{Gholamreza Haffari}
\author[3,\dag]{Bohan Zhuang}
\affiliation[1]{Monash University}
\affiliation[2]{Tongyi Lab, Alibaba Group}
\affiliation[3]{Zhejiang University}
\affiliation[4]{University of Queensland}
\contribution[\dag]{Corresponding Author}
\date{\today}
\note[\ddag]{Akide Liu is an intern at Tongyi Lab, Alibaba Group.}

\abstract{
Minute-scale cinematic video generation is a central challenge for generative video models. Existing paradigms address only fragments of this challenge: single-shot extrapolation preserves an anchor but lacks cinematic structure, while multi-shot storytelling imposes structure yet remains free to invent its visual states rather than continue an observed one. We define \emph{Multi-Shot Video Extrapolation} (MSVE), a task that extends an observed frame or clip into a sequence of cinematically structured shots while preserving anchor state and advancing narrative intent. This setting operates under the finite per-call generation budget of short-video models. We identify three coupled bottlenecks: (1)~global planners over-specify unsupported details from full screenplays; (2)~shot-level prompts dilute task-relevant state when carrying the complete story; (3)~temporal chaining turns generated frames into a lossy memory in which identity, scene, object, and action state decay. MSVE reveals that long-video failure is not merely a limitation of context length, but a failure of context allocation. We propose \emph{Recursive Context Allocation} (ReCA), an inference-time framework that allocates context hierarchically across planning and generation. ReCA recursively decomposes MSVE into context-bounded subproblems, invokes frozen generators at leaf nodes, and propagates structured state updates across time. To evaluate this setting, we further propose \emph{MSVE-Bench} and \emph{NB-Q}, a source-grounded protocol with prompts purpose-built for $3$--$5$~minute long-video generation, a regime not addressed by existing short-clip benchmarks. Compared to previous methods, ReCA improves average normalized score by 8–16\% over the strongest competing controller and improves multi-shot consistency metrics by 28–43\%. View the project page at \url{https://reca.vmv.re}.
}

\makeatletter
\newif\ifappendixtocrecording
\let\reca@addcontentsline\addcontentsline
\renewcommand{\addcontentsline}[3]{%
  \reca@addcontentsline{#1}{#2}{#3}%
  \ifappendixtocrecording
    \def\reca@tocext{toc}%
    \def\reca@sectiontype{section}%
    \def\reca@argext{#1}%
    \def\reca@argtype{#2}%
    \ifx\reca@argext\reca@tocext
      \ifx\reca@argtype\reca@sectiontype
        \begingroup
          \appendixtocrecordingfalse
          \addcontentsline{atoc}{#2}{#3}%
        \endgroup
      \fi
    \fi
  \fi
}

\newcommand{\printappendixfrontmatter}{%
  \begin{center}
    {\Large\bfseries Appendix: ReCA\par}
    {\large\bfseries Multi-Shot Long Video Extrapolation via Recursive Context Allocation\par}
  \end{center}
  \begingroup
  \renewcommand{\contentsname}{Appendix Contents}
  \setcounter{tocdepth}{1}
  \section*{\contentsname}
  \@starttoc{atoc}
  \endgroup
  \newpage
  \global\appendixtocrecordingtrue
}
\makeatother

\begin{document}

\thispagestyle{firstheader}
\maketitle
\pagestyle{empty}

\section{Introduction}
\label{sec:intro}

Minute-scale cinematic video generation is a central challenge for generative video models. Recent systems synthesize compelling short clips and extend them through image-to-video continuation, clip stitching, memory mechanisms, or language-based planning~\citep{ho2022videodiffusion,singer2022makeavideo,ho2022imagenvideo,hong2022cogvideo,videoldm2023,bartal2024lumiere,kondratyuk2024videopoet,yang2024cogvideox,polyak2024moviegen,lin2024opensoraplan}, but minute-scale cinematic videos are not merely longer clips: they are structured sequences of shots in which characters, objects, scenes, viewpoints, and narrative events evolve across cuts. Existing paradigms capture only part of this requirement: single-shot extrapolation preserves an observed anchor but produces one continuous take without cinematic structure~\citep{lfdm2023,seine2024,khachatryan2023text2videozero,xing2023dynamicrafter,guo2023sparsectrl,yin2023dragnuwa}, while multi-shot storytelling and text-to-multi-shot generation introduce shot structure but begin from text and are free to invent the visual world rather than continue an observed one~\citep{phenaki,videodirectorgpt,shotadapter2025,movieagent2025,mavis2026,videostudio2024,hong2023direct2v,zhou2024storydiffusion,gong2023talecrafter}. This leaves a task gap: current settings either preserve an observed world without multi-shot structure, or impose multi-shot structure without source-grounded extrapolation.

\begin{figure}[t]
    \centering
    \includegraphics[width=\linewidth]{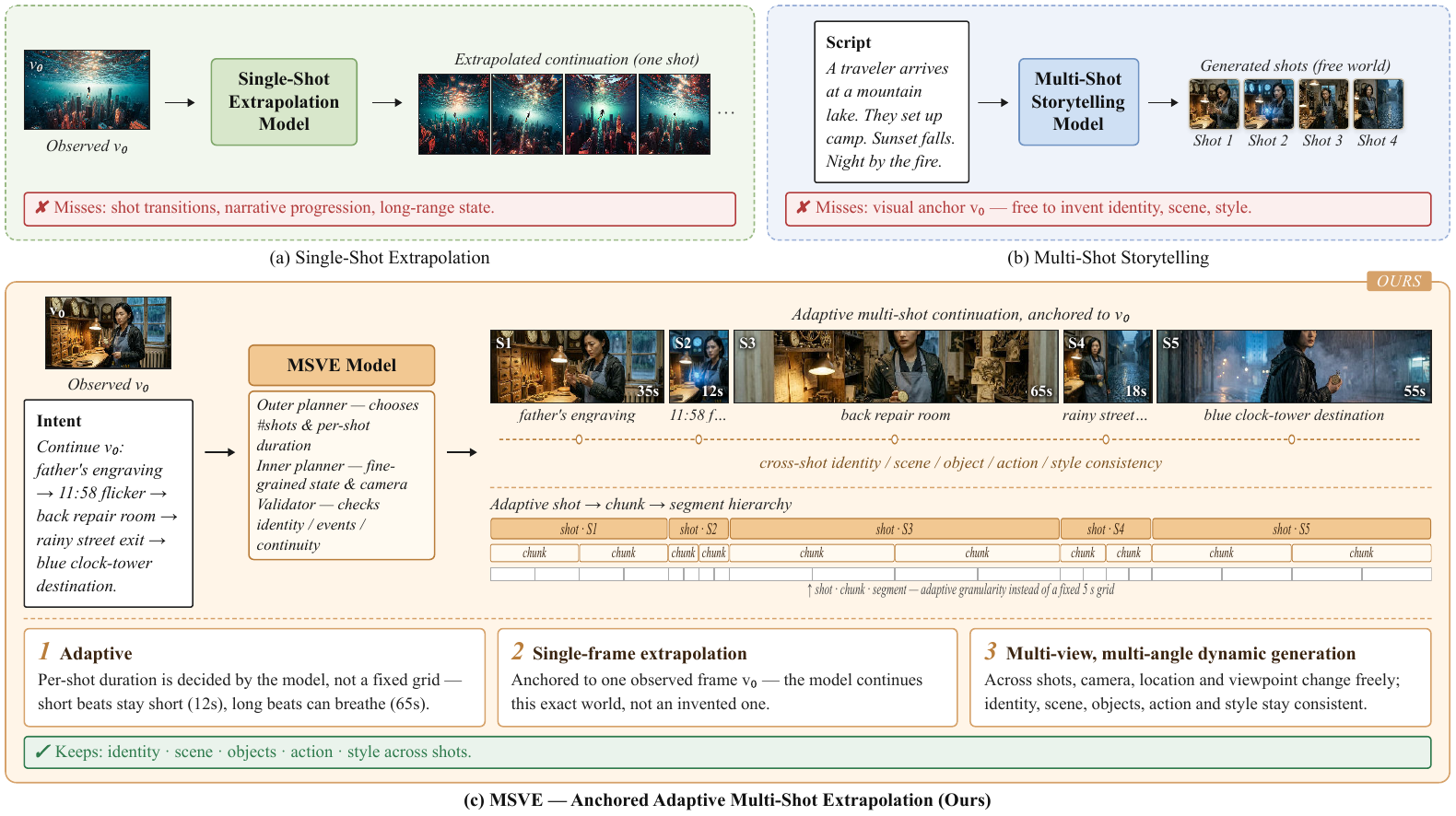}
    \caption{\textbf{Multi-Shot Video Extrapolation (MSVE).} \emph{(a)} Single-shot extrapolation extends an anchor into one continuous take but lacks cinematic structure. \emph{(b)} Multi-shot storytelling imposes shot structure but does not preserve an observed anchor. \emph{(c)} MSVE combines both: from $v_0$, $c$, and $T$ the system generates an adaptive shot sequence $\{s_i\}_{i=1}^{n}$ with $\sum_i d_i = T$ that preserves anchor state, advances the narrative, and produces plausible transitions under a finite per-call budget.}
    \label{fig:teaser}
\end{figure}

We introduce \textbf{Multi-Shot Video Extrapolation} (MSVE)\label{sec:msve}, a source-grounded long-video generation task that unifies these two requirements.
Given an observed visual anchor $v_0$, a narrative intent $c$, a target duration $T$, and a short-video generator $G$ with finite per-call duration and conditioning budgets, MSVE asks a system to generate
\[
    V=\{s_i\}_{i=1}^{n}, \qquad \sum_{i=1}^{n} d_i = T,
\]
where the number of shots, shot boundaries, shot durations, and per-shot conditioning states are chosen by the system.
A valid output must preserve the anchor state in $v_0$ (including identity, scene layout, objects, style, and physical configuration) while advancing the narrative intent, maintaining cross-shot consistency, and producing plausible transitions.
Thus, MSVE reduces to single-shot continuation when $n=1$, and to text-to-multi-shot storytelling when the observed anchor is removed; the full task requires both anchored extrapolation and adaptive multi-shot progression. Figure~\ref{fig:teaser} positions MSVE against single-shot extrapolation and multi-shot storytelling.

MSVE exposes a failure mode that is obscured when long video generation is framed only as increasing duration or context length.
\textbf{The failure of long video generation is not primarily a limitation of context length, but a failure of context allocation.}
More prompt text, more generated history, or more detailed plans do not necessarily provide more usable state to the next generation call.
This mirrors findings in long-context language modeling, where nominal context windows can diverge sharply from usable context
\citep{lostmiddle2024,ruler2024,oolong2025,bai2024longbench,zhang2024infinitebench,wang2024mmneedle,packer2023memgpt,jiang2023llmlingua,jiang2024longrag}.
In video generation, the problem is harder because the relevant state is multimodal and temporal: identity, layout, object status, action progress, camera state, narrative goals, and transition constraints must be selected and refreshed across calls.

We identify three coupled bottlenecks.
First, the \textbf{global planning bottleneck}: conditioning a planner on a full screenplay often leads to over-specified details unsupported by the observed anchor, causing the plan to drift away from the source world.
Second, the \textbf{shot-level context bottleneck}: passing the complete story or character bank into every shot dilutes the small set of state variables that the current shot must preserve.
Third, the \textbf{temporal chaining bottleneck}: using the previous generated frame or clip as the sole memory turns generation into a lossy state-propagation process, where identity, scene, object, and action consistency decay over time.
Together, these failures suggest that long-video generation should not maximize context exposure; it should allocate the right state to the right generation call.

We propose \textbf{Recursive Context Allocation} (ReCA), an inference-time framework for MSVE with frozen short-video generators.
ReCA decomposes long-video generation into context-bounded subproblems rather than submitting the full story to every call.
At the global level, it maintains compact persistent state, including visual memory, narrative memory, and transition memory.
At the shot level, it allocates only the state relevant to the current shot or video generator.
At the temporal level, it refreshes structured state after each generated segment, recording what changed, what must persist, and what should constrain the next call.
The leaf operation remains a standard call to a frozen generator; ReCA changes how context is selected, compressed, propagated, and re-injected.
This makes ReCA complementary to model-side long-video methods based on memory, sparse context routing, or long-context training
\citep{moc2025, longlive2025, videomemory2026}.

Evaluating MSVE requires a source-grounded protocol.
Existing video benchmarks~\citep{vbench_cvpr2024, vbench2, storyeval, narrlv, moviebench, vistorybench, msvbench2026} cover visual quality, text alignment, narrative completion, and multi-shot consistency only on tens-of-seconds clips, so they cannot test anchored multi-shot extrapolation at the minute scale where cross-shot identity, scene, and event-causality state begin to drift.
We therefore propose \textbf{MSVE-Bench} and the \textbf{NB-Q} score, a source-grounded protocol with prompts purpose-built for $3$--$5$~minute long-video generation that covers six capability groups (multi-shot transition consistency, character identity, scene/location handoff, event causality, cinematic realisation, artifact absence) under coverage-gated coherence.
We additionally report ReCA on a basket of established automatic metrics covering these axes (ViStory, VBench, MovieBench, StoryMem, MSVE-Bench) plus a human-rating study.
Across $20$ prompts and $3$ frozen generators, ReCA improves the average normalised score by $8$--$16\%$ over the strongest baseline (MovieAgent) on every backbone block of Table~\ref{tab:downstream_eval}; on the task-specific MSVE-Bench it gains $28$--$43\%$ over the strongest baseline and over $5\times$ versus the I2V-Extension baseline, while also achieving the lowest real-time factor (Table~\ref{tab:efficiency}), though MovieAgent posts a lower absolute wall-clock time due to shorter average generated shots.

\paragraph{Contributions.} 
\begin{itemize}[leftmargin=*,itemsep=0.15em,topsep=0.15em]
    \item We define \textbf{Multi-Shot Video Extrapolation}, a source-grounded long-video generation task that unifies anchored single-shot continuation and adaptive multi-shot storytelling.
    \item We identify \textbf{context allocation} as a core bottleneck in MSVE and propose \textbf{Recursive Context Allocation}, which addresses global planning, shot-level dilution, and temporal chaining failures with frozen video generators.
    \item We release \textbf{MSVE-Bench} and the \textbf{NB-Q} score as a source-grounded evaluation protocol for MSVE that decomposes the task into six capability groups, packaged with an instance schema, a six-group rubric (Table~\ref{tab:nbq-groups}), formal scoring equations with a coverage-gated coherence rule, $20$ MSVE prompts, and a reference NB-Q scorer; We evaluate ReCA on MSVE-Bench and a User Study.
\end{itemize}

\section{Related Work}
\label{sec:related}

 \textbf{Long-context models and effective context.} Long-context LLM benchmarks show that nominal and usable context length diverge sharply~\citep{ruler2024,oolong2025,lostmiddle2024,bai2024longbench,zhang2024infinitebench,wang2024mmneedle}, and memory- or compression-oriented language systems~\citep{packer2023memgpt,wang2023longmem,jiang2023llmlingua} treat long inputs as external state to be decomposed and refreshed; we ask whether an analogous behaviour governs T2V generation, where context is also temporal and planning-dependent. \textbf{Long video generation settings.} Existing long-video work clusters into three settings, each isolating one piece of the problem: \emph{duration extension and short-clip stitching}~\citep{wang2023genlvideo,qiu2023freenoise,henschel2025streamingt2v,tan2024videoinfinity} increases length but does not enforce cross-shot identity, scene, or narrative state; \emph{single-shot continuation} preserves local visual continuity from an observed anchor but does not require cinematic cuts or act-level progression; and \emph{text-to-multi-shot and prompt-to-plan systems}~\citep{videodirectorgpt,dreamfactory,mavis2026} introduce shot or scene structure, but start from text rather than from an observed visual state that must persist. \textbf{Memory, planning, and evaluation.} A complementary line modifies the generator through memory, sparse context routing, or autoregressive correction~\citep{longlive2025,moc2025,videomemory2026,svi2025}, while language-agent and memory systems externalise state across reasoning steps~\citep{packer2023memgpt,wang2023longmem,shinn2023reflexion}; we keep the generator frozen and ask how language-side state should be allocated across calls. Evaluation work~\citep{vbench_cvpr2024,vbench2,fetv2024,t2vcompbench,moviebench,narrlv,msvbench2026,liu2023evalcrafter,wu2024t2vscore,han2025videobench,meng2024phygenbench,liao2024devil} consistently argues that long video requires decomposed evaluation beyond frame-level quality. \textbf{Task gap.} None of these settings combines the two requirements at once: stay anchored to an observed visual state \emph{and} progress through multiple coherent shots under a finite per-call conditioning budget. Multi-Shot Video Extrapolation fills this gap. The full related is in Appendix~\ref{app:full_related}.

\section{Context Allocation, Not Context Length}
\label{sec:context_allocation}
\label{sec:contextness}

\begin{figure}[h]
    \centering
    \includegraphics[width=\linewidth]{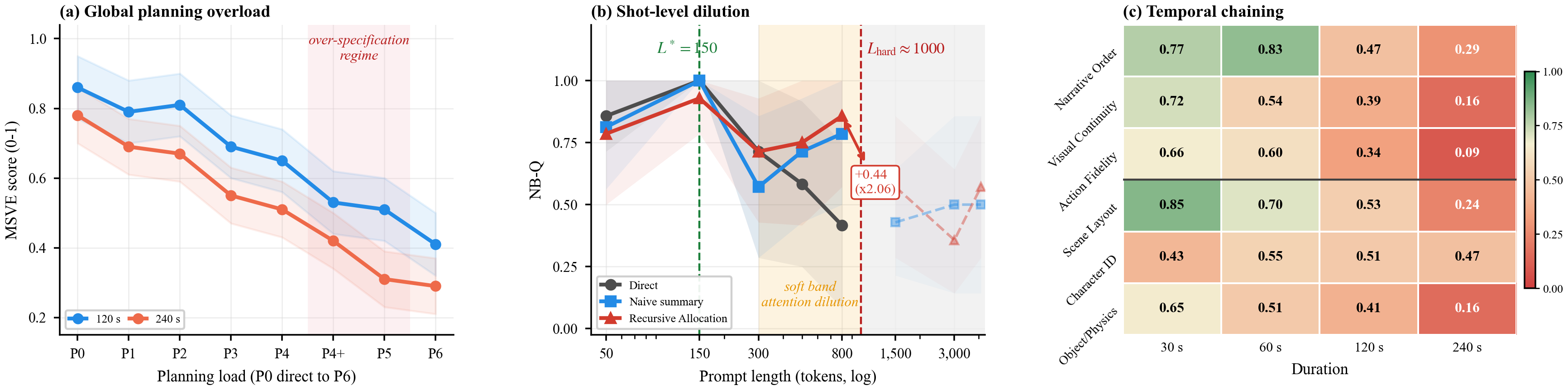}
    \caption{
    \textbf{Context allocation failures in MSVE.}
    The usable context of a frozen video generator is narrower than its nominal window. The three panels diagnose the three bottlenecks analysed in this section: \emph{(a) global planning overload}, where over-detailed plans introduce unsupported constraints; \emph{(b) shot-level dilution}, where longer per-shot prompts reduce the salience of state needed by the current call; and \emph{(c) temporal chaining}, where generated frames become a lossy and aging memory of identity, scene, object, and action state. Additional context helps only while it raises the density of \emph{usable} state. Per-panel numerical anchors are in Appendix~\ref{app:contextness-panels}.
    }
    \label{fig:contextness}
    \end{figure}

MSVE reveals a central limitation of current video generators: \textbf{more context does not necessarily improve generation quality}. The key bottleneck is not context length, but \emph{context allocation} under a bounded conditioning budget. This view is consistent with the information-bottleneck principle, which preserves task-relevant information rather than maximising total input~\citep{tishby2000ib}, and with recent long-context studies showing that usable context can be substantially smaller than the nominal window~\citep{lostmiddle2024, ruler2024, oolong2025}. We identify three recurring failure modes along different scaling axes (Figure~\ref{fig:contextness}); per-panel numerical anchors are deferred to Appendix~\ref{app:contextness-panels}.

\textbf{Global planning: unsupported state} (Fig.~\ref{fig:contextness}a). Increasing planning detail initially improves structure, but eventually harms performance: over-specified plans introduce constraints (camera motion, scene layout, intermediate events) that are not supported by the observed anchor, leading to drift, closely related to hallucination in generative models~\citep{ji2023survey, huang2025hallucination} and to the difficulty LLM-based planners have with grounded action reasoning under incomplete information~\citep{valmeekam2023planning}; global planning should therefore remain \emph{evidence-bounded}, selecting only anchor-supported state. \textbf{Shot-level prompting: diluted state} (Fig.~\ref{fig:contextness}b). Longer prompts degrade performance even before reaching hard limits because irrelevant details reduce the salience of task-critical state, an effect that aligns with ``lost-in-the-middle'' behaviour in long-context language models~\citep{lostmiddle2024, ruler2024}, so effective prompting requires \emph{compact, high-density context} focused on the current shot. \textbf{Temporal chaining: stale state} (Fig.~\ref{fig:contextness}c). Using generated frames as implicit memory leads to progressive degradation: visual history is lossy and cannot reliably encode identity, object state, or causal dependencies, producing accumulated drift consistent with observations in long-horizon autoregressive generation~\citep{longlive2025, videomemory2026, svi2025}, which motivates \emph{explicit state propagation} via memory refresh rather than purely implicit chaining. \textbf{Unified insight.} Across all three axes, additional context is beneficial only when it remains \emph{supported}, \emph{salient}, and \emph{fresh}; otherwise it introduces noise in the form of unsupported detail, diluted relevance, or stale memory, so long-video generation should optimise for \textbf{usable context density} rather than raw context length, motivating our Recursive Context Allocation (ReCA) framework, which explicitly controls how context is selected, structured, and propagated across generation steps.

\section{Recursive Context Allocation}
\label{sec:method}

Section~\ref{sec:context_allocation} shows that MSVE failures arise when context length does not become usable context, with three coupled bottlenecks: \emph{unsupported}, \emph{diluted}, and \emph{stale} state. We propose \textbf{Recursive Context Allocation} (ReCA), an inference-time framework with frozen short-video generators that addresses these three bottlenecks with three coupled operators, $\textsc{Plan}$, $\textsc{Allocate}$, and $\textsc{Refresh}$. ReCA maintains task state outside the generator, recursively decomposes the target video into context-bounded subproblems, and injects only the state that is supported by the anchor, salient for the current shot, and fresh at the current time.

\paragraph{External state.} Let $G$ be a frozen short-video generator with duration budget $\tau_G$ and conditioning budget $B_G$. Given an observed anchor $v_0$, narrative intent $c$, and target duration $T$, ReCA maintains an external state $S_k = (A_k, N_k, M_k, Q_k)$ comprising visual, narrative, transition, and per-variable-metadata fields; this is a compact non-parametric memory queried before each generation and updated after, with the per-field schema deferred to Appendix~\ref{app:method-framework}.

\begin{figure}[t]
\centering
\includegraphics[width=\linewidth]{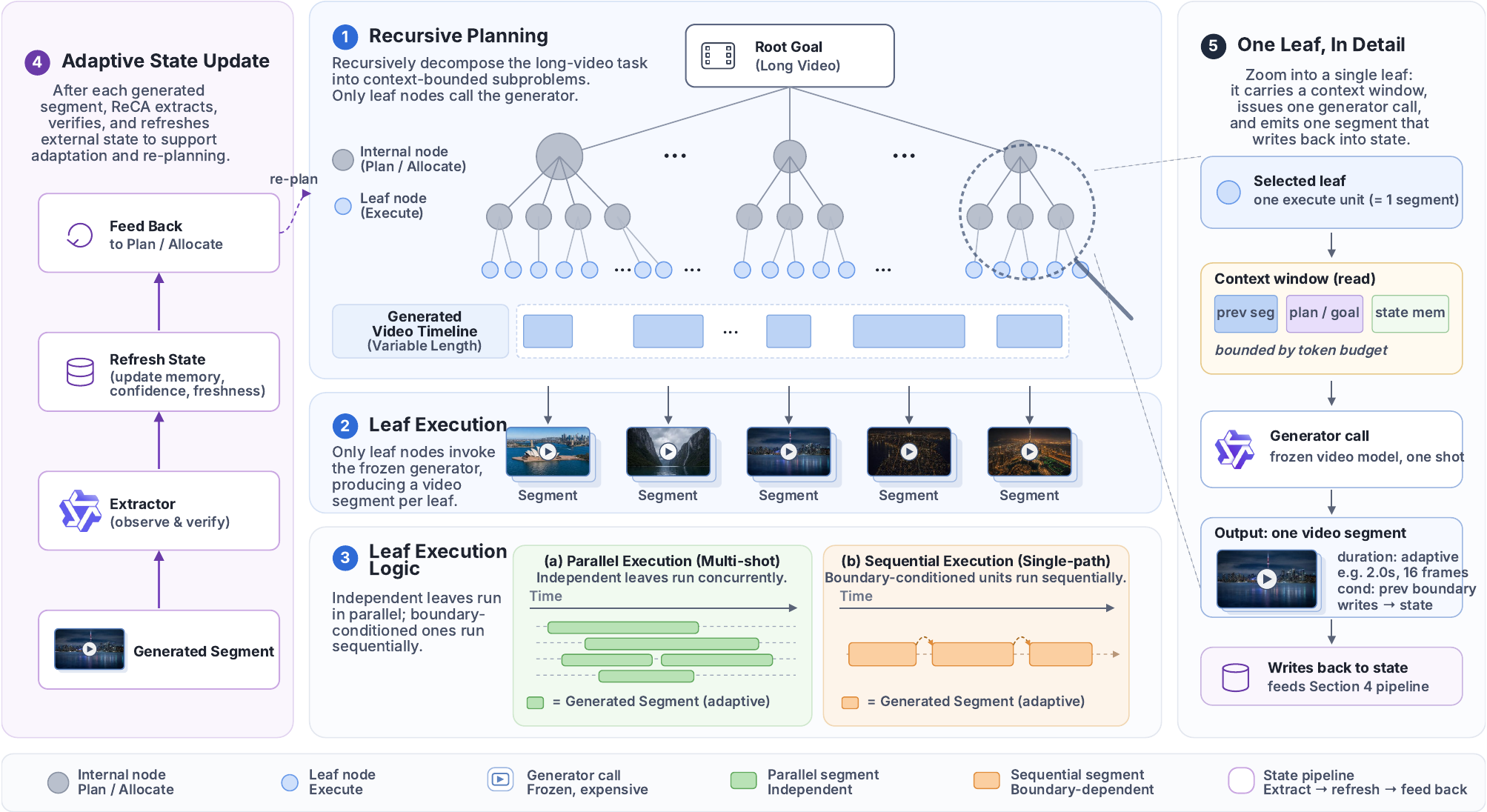}
\caption{\textbf{Recursive Context Allocation (ReCA).} \emph{(1) Recursive planning} decomposes the long-video task (root) into context-bounded subproblems; only leaf nodes (blue) call the frozen generator. The induced sequence of leaves produces a variable-length \emph{Generated Video Timeline}. \emph{(2) Leaf execution.} Each leaf invokes one generator call and produces a video segment. \emph{(3) Leaf execution logic.} Independent leaves can run concurrently as parallel segments (green); leaves that share a boundary frame are chained sequentially with handoff arrows (orange). Segment durations are adaptive to context and budget. \emph{(4) Adaptive state update.} Each generated segment is read by an extractor for observation and verification, refreshes the persistent state (memory, confidence, freshness), and feeds back into planning and allocation, closing the recursion.}
\label{fig:reca-main}
\end{figure}

\paragraph{Recursion.} ReCA recursively decomposes MSVE until every leaf call lies within both the duration budget $\tau_G$ and the conditioning budget $B_G$ (measured in prompt tokens). At each step, $\textsc{Plan}$ (Section~\ref{sec:method-global}) emits a sibling schedule $\{(g_{\ell},d_{\ell},b_{\ell})\}_{\ell=1}^{m}$ together with a dependency map $\mathrm{prev}:\{1,\dots,m\}\!\to\!\{1,\dots,m\}\cup\{\bot\}$ that fixes each sibling's input state $S^{(\ell)}$ (independent when $\mathrm{prev}(\ell)\!=\!\bot$, state-threaded otherwise). The recursion is then
\begin{equation}
\label{eq:reca}
\mathrm{ReCA}(S,g,d,b)=
\begin{cases}
G(p,\,b;\,d), & d \leq \tau_G,\ |p| \leq B_G,\\[2pt]
\bigoplus_{\ell=1}^{m}\mathrm{ReCA}(S^{(\ell)},g_{\ell},d_{\ell},b_{\ell}), & \text{otherwise},
\end{cases}
\end{equation}
where $p=\textsc{Compile}(\textsc{Allocate}(S,g,b))$ is the leaf prompt and $\oplus$ denotes temporal concatenation; the leaf case calls $\textsc{Allocate}$, $\textsc{Compile}$, $G$, and $\textsc{Refresh}$ (Sections~\ref{sec:method-local},~\ref{sec:method-refresh}). Figure~\ref{fig:reca-main} visualises the four phases this recursion induces. The leaf-case operator chain, the independent-versus-threaded dependency-map semantics, the 2-level implementation we use in practice, and the per-phase narration of Figure~\ref{fig:reca-main} are in Appendix~\ref{app:method-framework}.

\subsection{Evidence-Bounded Global Allocation}
\label{sec:method-global}

The first operator addresses the global planning bottleneck (Fig.~\ref{fig:reca-main}, phase~1). ReCA treats planning as \emph{evidence-bounded state allocation}: each variable $z$ is tagged with a provenance label $\pi(z)\!\in\!\{\textsc{anchor},\textsc{intent},\textsc{generated}\}$ and an anchor-support score $q_{\mathrm{sup}}(z\mid v_0)$, and anchor-provenance variables enter the persistent state only when $q_{\mathrm{sup}}\!\geq\!\epsilon$. The planner emits a shot schedule $P = \{(g_i,d_i,\kappa_i)\}_{i=1}^{n}$ that optimises a compact-plan objective:
\begin{equation}
\label{eq:plan-objective}
P^{\star}
=
\arg\max_P
\sum_i R_{\mathrm{nar}}(g_i,c)
+ \lambda_{\mathrm{tr}} R_{\mathrm{trans}}(g_i,g_{i+1})
- \lambda_{\mathrm{unsup}}
\sum_{z\in P}
[\epsilon-q_{\mathrm{sup}}(z\mid v_0)]_{+},
\end{equation}
balancing narrative progress, plausible shot transitions, and a penalty on unsupported visual commitments. The objective is a design principle, not a differentiable loss: preserve long-horizon variables the generator is likely to lose, and avoid details the anchor does not license. The full provenance-admission rule, schedule-field semantics, and the connection to information-bottleneck and hierarchical planning~\citep{tishby2000ib,sutton1999options,bercher2019hierarchical} are in Appendix~\ref{app:method-operators}.

\subsection{Context-Bounded Shot Decomposition}
\label{sec:method-local}

The second operator addresses the shot-level dilution bottleneck (Fig.~\ref{fig:reca-main}, phase~2). From the candidate pool $\Omega_k = S_k \cup g_k \cup b_k$ (persistent state, current shot goal, visual boundary), ReCA selects a compact slice
\begin{equation}
\label{eq:context-selector}
C_k^{\star}
=
\arg\max_{C\subseteq \Omega_k,\ |\textsc{Compile}(C)|\leq B_G}
\sum_{z\in C}
q_{\mathrm{sup}}(z\mid v_0)
\cdot
w_k(z;g_k)
\cdot
f_k(z)
-
\lambda_{\mathrm{red}}\mathrm{Red}(C),
\end{equation}
where $w_k$ is salience, $f_k$ is freshness, and $\mathrm{Red}(C)$ penalises redundant or conflicting descriptions; the budget is the same $B_G$ that bounds Eq.~\ref{eq:reca}. The local compiler turns $C_k^\star$ into a generator-facing prompt $p_k = \textsc{Compile}(C_k^\star)$ and, when a planned shot duration exceeds $\tau_G$, decomposes the shot into units $(u_{ij},\delta_{ij},\kappa_{ij})$ with $\delta_{ij}\!\leq\!\tau_G$, so each leaf call $r_{ij} = G(p_{ij}, b_{ij}; \delta_{ij})$ stays in the packed, high-salience regime of Section~\ref{sec:context_allocation}. The fixed prompt structure, the connection to prompt compression and retrieval-augmented generation~\citep{lewis2020rag,jiang2023llmlingua,jiang2024longrag}, and the formal unit-level recursion are in Appendix~\ref{app:method-operators}.

\subsection{Structured State Propagation}
\label{sec:method-refresh}

The third operator addresses the temporal chaining bottleneck (Fig.~\ref{fig:reca-main}, phase~4). Naive chaining sets $S^{\mathrm{chain}}_{k+1} = \phi(r_k)$, forcing the next call to infer identity, layout, and event progress from pixels alone. ReCA instead separates visual boundary from semantic memory: a lightweight extractor-verifier $\chi$ produces structured observations $\hat{o}_k = \chi(r_k, p_k, S_k)$ from each generated unit, and $S_{k+1} = \textsc{Refresh}(S_k,\hat{o}_k,g_k)$ updates the value, provenance, confidence, and last-refresh time of every affected variable. The same extractor doubles as a verifier and triggers local repair from a small action set $\{\textsc{RepackPrompt}, \textsc{ReanchorState}, \textsc{RegenerateUnit}, \textsc{SplitUnit}\}$ rather than resubmitting the full story. The variable-record schema, the freshness-based re-injection priority $\mathrm{Pri}_k(z)\!=\!w_k(z;g_{k+1})\!\cdot\!(1-f_k(z))$, and the relationship to language-agent and model-side memory methods~\citep{yao2023react,packer2023memgpt,shinn2023reflexion,moc2025,longlive2025,videomemory2026,svi2025} are in Appendix~\ref{app:method-operators}.

\subsection{Parallel Recursive Execution}
\label{sec:method-parallel}

A direct consequence of the recursive form in Eq.~\ref{eq:reca} is that ReCA is not inherently sequential (Fig.~\ref{fig:reca-main}, phase~3): the recursion produces a tree whose internal nodes are inexpensive context operators (\textsc{Plan}, \textsc{Allocate}, \textsc{Refresh}) and whose leaves are the only expensive calls. Sibling subtrees with $\mathrm{prev}(\ell)=\bot$ all read the same parent state $S$ and are independent, so they can be dispatched concurrently; only state-threaded siblings ($\mathrm{prev}(\ell)\neq\bot$) sit on the sequential critical path, and the serial fraction is set by the dependency map $\mathrm{prev}$ rather than by raw shot count. Letting $N$ be the number of leaf calls and $W$ the available concurrency, sequential ReCA has wall-clock cost $\Theta(N\,\tau_G)$ while parallel ReCA has cost $\Theta(\lceil N/W\rceil\,\tau_G + D\,\tau_{\mathrm{op}})$, where $D$ is the longest path in the leaf dependency graph induced by $\mathrm{prev}$ and $\tau_{\mathrm{op}}\!\ll\!\tau_G$ is the cost of an internal operator. The first term shrinks linearly with $W$ until the rate limit is saturated; the second term, set by the I2V handoff structure of the plan, is the only intrinsic floor, so the same allocation decisions that keep each leaf inside the prompt and duration budget also keep enough leaves independent to be issued in parallel. The three concrete levels of parallelism (leaf, intra-shot operator, and pipeline along a chain), the map-reduce cost model, and the parallel-safe repair semantics are detailed in Appendix~\ref{app:method-parallel}.

\section{Experiments}
\label{sec:exp-main}

\begin{table*}[t]
  \centering
  \small
  \setlength{\tabcolsep}{6pt}
  \renewcommand{\arraystretch}{1.15}
  \caption{
  \textbf{Quantitative Evaluation.}
  Models are grouped by rendering backbone; methods within a block share the generator, the per-call duration budget, and the long-video prompt package, so absolute scores are comparable inside a block. Each cell reports mean$\,\pm$\,standard error over the $20$ MSVE prompts. We report averaged VBench scores, NB-Q for MSVE-Bench, ViStory-Self and ViStory-Cross for single-shot and cross-frame consistency, and MovieBench for hierarchical shot alignment.The \emph{Normalisation for Avg.} column carries no error bar because it aggregates six metrics with different per-cell variances; for significance use the per-metric SE columns directly.
  }
  
  \label{tab:downstream_eval}
  \resizebox{\textwidth}{!}{
  \begin{tabular}{llccccccc}
  \toprule
  \multirow{2}{*}{\textbf{Model}} & \multirow{2}{*}{\textbf{Method}}
  & \multicolumn{3}{c}{\textbf{Inner-shot}}
  & \multicolumn{2}{c}{\textbf{Cross-shot}}
  & \textbf{Global}
  & \multirow{2}{*}{\textbf{Avg.}} \\
  \cmidrule(lr){3-5} \cmidrule(lr){6-7} \cmidrule(lr){8-8}
  &
  & \textbf{VBench}
  & \textbf{StoryMem}
  & \textbf{ViStory-Self}
  & \textbf{ViStory-Cross}
  & \textbf{MovieBench}
  & \textbf{MSVE-Bench}
  &  \\
  \midrule
  
  \rowcolor{orange!12}
  \multicolumn{9}{c}{\textbf{Open-Source Models}} \\
  \midrule
  
  \multirow{5}{*}{\textsc{Wan~2.2}~\citep{wan2025}}
  & I2V Extension
  & 0.853\sd{0.012}
  & 0.943\sd{0.014}
  & 0.512\sd{0.026}
  & 0.127\sd{0.018}
  & 19.327\sd{1.34}
  & 0.094\sd{0.022}
  & 0.454 \\

  & VGoT~\citep{vgot2025}
  & 0.871\sd{0.010}
  & 0.988\sd{0.008}
  & 0.631\sd{0.024}
  & 0.263\sd{0.027}
  & 25.144\sd{1.56}
  & 0.113\sd{0.024}
  & 0.520 \\

  & Mora~\citep{mora2024}
  & \underline{0.885}\sd{0.011}
  & \underline{0.990}\sd{0.008}
  & 0.666\sd{0.025}
  & \underline{0.303}\sd{0.028}
  & 23.726\sd{1.41}
  & 0.187\sd{0.029}
  & 0.545 \\

  & MovieAgent~\citep{movieagent2025}
  & 0.878\sd{0.013}
  & 0.945\sd{0.014}
  & \underline{0.753}\sd{0.022}
  & 0.232\sd{0.025}
  & \textbf{26.513}\sd{1.18}
  & \underline{0.572}\sd{0.041}
  & \underline{0.608} \\

  & \textbf{ReCA (Ours)}
  & \textbf{0.891}\sd{0.009}
  & \textbf{0.992}\sd{0.005}
  & \textbf{0.755}\sd{0.021}
  & \textbf{0.378}\sd{0.030}
  & \underline{26.062}\sd{1.27}
  & \textbf{0.819}\sd{0.038}
  & \textbf{0.683} \\
  
  \midrule
  
  \rowcolor{gray!15}
  \multicolumn{9}{c}{\textbf{Proprietary Models}} \\
  \midrule
  
  \multirow{5}{*}{\textsc{Wan~2.7}~\citep{wan27}}
  & I2V Extension
  & 0.951\sd{0.008}
  & 0.952\sd{0.013}
  & 0.587\sd{0.030}
  & 0.142\sd{0.020}
  & 19.881\sd{1.42}
  & 0.123\sd{0.025}
  & 0.492 \\

  & VGoT~\citep{vgot2025}
  & \underline{0.978}\sd{0.006}
  & 0.965\sd{0.011}
  & 0.798\sd{0.022}
  & 0.274\sd{0.026}
  & 21.959\sd{1.38}
  & 0.182\sd{0.023}
  & 0.569 \\

  & Mora~\citep{mora2024}
  & 0.972\sd{0.007}
  & \underline{0.988}\sd{0.007}
  & 0.855\sd{0.020}
  & 0.263\sd{0.029}
  & \underline{26.062}\sd{1.21}
  & 0.276\sd{0.031}
  & 0.602 \\

  & MovieAgent~\citep{movieagent2025}
  & 0.974\sd{0.006}
  & 0.978\sd{0.009}
  & \underline{0.913}\sd{0.018}
  & \underline{0.286}\sd{0.024}
  & 25.467\sd{1.34}
  & \underline{0.738}\sd{0.039}
  & \underline{0.691} \\

  & \textbf{ReCA (Ours)}
  & \textbf{0.984}\sd{0.005}
  & \textbf{0.992}\sd{0.005}
  & \textbf{0.936}\sd{0.015}
  & \textbf{0.324}\sd{0.022}
  & \textbf{31.454}\sd{1.06}
  & \textbf{0.942}\sd{0.030}
  & \textbf{0.749} \\
  
  \midrule
  
  \multirow{5}{*}{\textsc{HappyHorse~1.0}~\citep{happyhorse2026model}}
  & I2V Extension
  & 0.948\sd{0.009}
  & 0.961\sd{0.012}
  & 0.539\sd{0.029}
  & 0.158\sd{0.019}
  & 20.214\sd{1.37}
  & 0.102\sd{0.026}
  & 0.485 \\

  & VGoT~\citep{vgot2025}
  & \underline{0.978}\sd{0.007}
  & 0.978\sd{0.010}
  & 0.710\sd{0.024}
  & \underline{0.370}\sd{0.029}
  & 24.470\sd{1.49}
  & 0.153\sd{0.025}
  & 0.572 \\

  & Mora~\citep{mora2024}
  & 0.974\sd{0.008}
  & 0.975\sd{0.011}
  & \underline{0.798}\sd{0.022}
  & 0.331\sd{0.027}
  & 25.165\sd{1.31}
  & 0.273\sd{0.033}
  & 0.600 \\

  & MovieAgent~\citep{movieagent2025}
  & 0.976\sd{0.007}
  & \underline{0.980}\sd{0.009}
  & 0.610\sd{0.026}
  & 0.284\sd{0.024}
  & \underline{27.962}\sd{1.45}
  & \underline{0.687}\sd{0.043}
  & \underline{0.636} \\

  & \textbf{ReCA (Ours)}
  & \textbf{0.985}\sd{0.006}
  & \textbf{0.990}\sd{0.006}
  & \textbf{0.866}\sd{0.019}
  & \textbf{0.383}\sd{0.025}
  & \textbf{28.517}\sd{1.18}
  & \textbf{0.913}\sd{0.034}
  & \textbf{0.737} \\
  
  \bottomrule
  \end{tabular}
  }
  \end{table*}

  \begin{figure}[t]
    \centering
    \includegraphics[width=\linewidth]{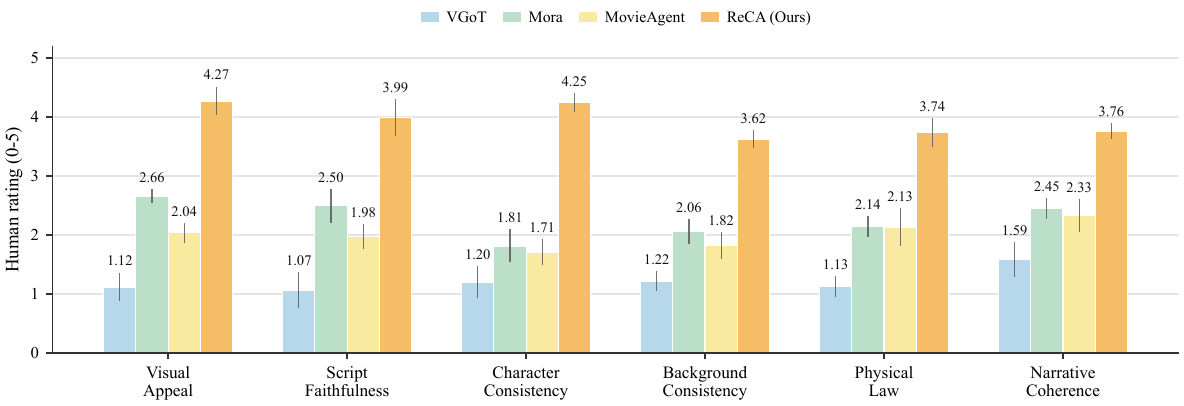}
    \caption{\textbf{User-study ratings for automated movie generation.} Grouped bars compare VGoT, Mora, MovieAgent, and ReCA on six human-rating criteria, with Background Consistency inserted before Physical Law. Whiskers report the per-cell standard error of the mean across raters; bars use a light colorblind-friendly palette; higher is better.}
    \label{fig:user-study-movieagent}
\end{figure}

\paragraph{Setup.}
We evaluate ReCA as an inference-time controller for frozen video generators against three long-video controllers (VGoT~\citep{vgot2025}, Mora~\citep{mora2024}, MovieAgent~\citep{movieagent2025}) and a single-shot \textbf{I2V Extension} baseline that repeatedly extends the anchor by last-frame handoff. Table~\ref{tab:downstream_eval} reports results on the open-source \textsc{Wan~2.2} family~\citep{wan2025}, our primary reproducibility target since reviewers can re-run it locally from public weights, and on two proprietary backbones, \textsc{Wan~2.7}~\citep{wan27} and \textsc{HappyHorse~1.0}~\citep{happyhorse2026model} (Appendix~\ref{app:happyhorse}), as generalisation checks across hosted backbones. Within each block all four methods share the same frozen backbone, per-call duration budget, and long-video prompt package; ReCA differs only in how it allocates global plan, shot-local prompt, and refreshed temporal state. Its planner, allocator, prompt compiler, and verifier-extractor $\chi$ share one frozen Qwen~3.6 Plus model~\citep{qwen2026qwen36plus}, while the headline NB-Q metric is scored by an independent GPT-$5.5$ API call (a stronger, non-Qwen vision-language judge) so the metric and the controller never share a model family. Full baseline descriptions are in Appendix~\ref{app:exp-setup}.

\paragraph{MSVE-Bench protocol.}
We further propose \emph{MSVE-Bench}, a source-grounded evaluation protocol purpose-built for $3$--$5$~minute multi-shot extrapolation; each instance is a hand-engineered prompt package paired with an \emph{NB-Q} score that decomposes preservation of the observed world and advancement of the intended narrative into six capability groups (transition consistency, character identity, scene/location handoff, event causality, cinematic realisation, artifact absence). Prior video benchmarks~\citep{vbench_cvpr2024,vbench2,storyeval,narrlv,moviebench,vistorybench,msvbench2026} confine evaluation to short clips and cannot expose the cross-shot state that emerges only at minute scale, so MSVE-Bench is the first protocol whose prompts are purpose-built for that regime. The packaged release ships an instance schema, the six-group rubric (Table~\ref{tab:nbq-groups}), formal scoring equations with a coverage-gated coherence rule, the $20$ MSVE prompts used in the user study, and a reference NB-Q scorer running on the GPT-$5.5$ API as a non-Qwen judge that keeps the metric model-family-independent from the Qwen-side ReCA controller; the full schema, scoring equations, and the cross-validation mapping the six groups onto Table~\ref{tab:downstream_eval} and the user-study criteria are in Appendix~\ref{sec:mlve} and~\ref{app:msve-bench-detail}.

\paragraph{Quantitative results.}
Table~\ref{tab:downstream_eval} shows that ReCA is the strongest method across every backbone block. The clearest signal lies in MSVE-Bench, the only column purpose-built for multi-shot long-video evaluation, where ReCA's gain over the strongest baseline is the largest in the table on every backbone. The remaining columns (VBench, StoryMem, ViStory) and the Average are short-clip scorers whose dynamic range saturates once methods share a competent backbone, so their margins are inherently compressed; ReCA still leads under this compression. Our User Study (Figure~\ref{fig:user-study-movieagent}) corroborates this picture, with a wide perceptual gap on exactly the cross-shot dimensions the saturated metrics cannot distinguish.

\paragraph{User study and NB-Q agreement.}
We compare ReCA against VGoT, Mora, and MovieAgent on the frozen \textsc{Wan~2.7} backbone over the $20$ MSVE prompts ($80$ long videos, $3$--$5$~min) using six $0$--$5$ Likert criteria (visual appeal, script faithfulness, character consistency, background consistency, physical law, narrative coherence); each rater scores a random subset of $5$--$10$ videos with method identity hidden and per-prompt order shuffled. ReCA wins every criterion (Figure~\ref{fig:user-study-movieagent}: $4.27$, $3.99$, $4.25$, $3.62$, $3.74$, $3.76$ in that order), with the largest gaps on state-preservation, where the next-best baseline stays below $2.7$ on visual appeal, $2.5$ on script faithfulness, and $2.1$ on character and background consistency. The GPT-$5.5$-based NB-Q metric tracks these human means at the prompt$\,\times\,$method level (Spearman $\rho\!\approx\!0.81$, Kendall $\tau\!\approx\!0.66$) and induces an identical per-method ranking on every backbone (ReCA $>$ MovieAgent $>$ Mora/VGoT $>$ I2V Extension), so the headline ordering is corroborated by an independent non-Qwen judge and by human raters rather than a self-reinforcing one. Full mechanics (per-participant load, dropout handling, inter-rater agreement, external-pool corroboration) are deferred to Appendix~\ref{app:user-study}.

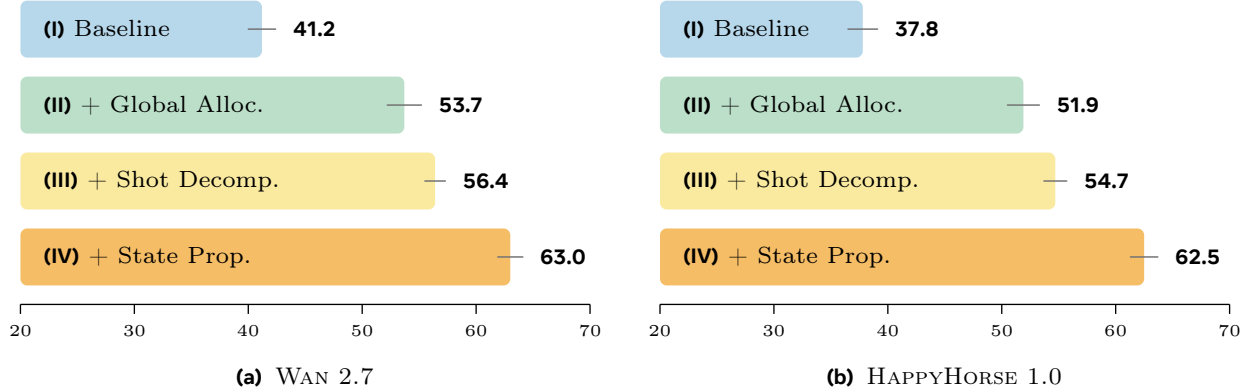
\begin{figure}[t]
  \centering
  \begin{minipage}{\linewidth}
  \centering

  \definecolor{ablI}{HTML}{B7D8EB}
  \definecolor{ablII}{HTML}{BCDFC9}
  \definecolor{ablIII}{HTML}{FAEAA0}
  \definecolor{ablIV}{HTML}{F5BD66}

  \begin{subfigure}[t]{0.49\linewidth}
  \centering
  \resizebox{\linewidth}{!}{%
  \begin{tikzpicture}[font=\scriptsize, x=1.2mm, y=1mm,
      errbar/.style={draw=black!55, line width=0.45pt, line cap=round}]
    \fill[ablI, rounded corners=2pt] (20, 0) rectangle (41.2, 6);
    \draw[errbar] (40.0, 3) -- (42.4, 3);
    \node[anchor=west] at (21, 3) {\textbf{(I)} Baseline};
    \node[anchor=west] at (42.9, 3) {\textbf{41.2}};
    \fill[ablII, rounded corners=2pt] (20, -8) rectangle (53.7, -2);
    \draw[errbar] (52.2, -5) -- (55.2, -5);
    \node[anchor=west] at (21, -5) {\textbf{(II)} + Global Alloc.};
    \node[anchor=west] at (55.7, -5) {\textbf{53.7}};
    \fill[ablIII, rounded corners=2pt] (20, -16) rectangle (56.4, -10);
    \draw[errbar] (55.5, -13) -- (57.3, -13);
    \node[anchor=west] at (21, -13) {\textbf{(III)} + Shot Decomp.};
    \node[anchor=west] at (57.8, -13) {\textbf{56.4}};
    \fill[ablIV, rounded corners=2pt] (20, -24) rectangle (63.0, -18);
    \draw[errbar] (61.9, -21) -- (64.1, -21);
    \node[anchor=west] at (21, -21) {\textbf{(IV)} + State Prop.};
    \node[anchor=west] at (64.6, -21) {\textbf{63.0}};
    \draw (20, -26) -- (70, -26);
    \foreach \x in {20, 30, 40, 50, 60, 70} {
      \draw (\x, -26) -- (\x, -27);
      \node[anchor=north, font=\tiny] at (\x, -27) {\x};
    }
  \end{tikzpicture}%
  }
  \caption{\textsc{Wan~2.7}}
  \label{fig:reca-ablation-wan27}
  \end{subfigure}
  \hfill
  \begin{subfigure}[t]{0.49\linewidth}
  \centering
  \resizebox{\linewidth}{!}{%
  \begin{tikzpicture}[font=\scriptsize, x=1.2mm, y=1mm,
      errbar/.style={draw=black!55, line width=0.45pt, line cap=round}]
    \fill[ablI, rounded corners=2pt] (20, 0) rectangle (37.8, 6);
    \draw[errbar] (36.5, 3) -- (39.1, 3);
    \node[anchor=west] at (21, 3) {\textbf{(I)} Baseline};
    \node[anchor=west] at (39.6, 3) {\textbf{37.8}};
    \fill[ablII, rounded corners=2pt] (20, -8) rectangle (51.9, -2);
    \draw[errbar] (50.5, -5) -- (53.3, -5);
    \node[anchor=west] at (21, -5) {\textbf{(II)} + Global Alloc.};
    \node[anchor=west] at (53.8, -5) {\textbf{51.9}};
    \fill[ablIII, rounded corners=2pt] (20, -16) rectangle (54.7, -10);
    \draw[errbar] (53.7, -13) -- (55.7, -13);
    \node[anchor=west] at (21, -13) {\textbf{(III)} + Shot Decomp.};
    \node[anchor=west] at (56.2, -13) {\textbf{54.7}};
    \fill[ablIV, rounded corners=2pt] (20, -24) rectangle (62.5, -18);
    \draw[errbar] (61.3, -21) -- (63.7, -21);
    \node[anchor=west] at (21, -21) {\textbf{(IV)} + State Prop.};
    \node[anchor=west] at (64.2, -21) {\textbf{62.5}};
  
    \draw (20, -26) -- (70, -26);
    \foreach \x in {20, 30, 40, 50, 60, 70} {
      \draw (\x, -26) -- (\x, -27);
      \node[anchor=north, font=\tiny] at (\x, -27) {\x};
    }
  \end{tikzpicture}%
  }
  \caption{\textsc{HappyHorse~1.0}}
  \label{fig:reca-ablation-happyhorse}
  \end{subfigure}
  \end{minipage}
  \caption{\textbf{Component ablation of ReCA on ViStory.} Each bar incrementally adds one ReCA component on top of the bar above it: (II) \emph{Evidence-Bounded Global Allocation}, (III) \emph{Context-Bounded Shot Decomposition}, (IV) \emph{Structured State Propagation}. Scores are the average of ViStory-Self and ViStory-Cross ($\times 100$); whiskers report the per-bar standard error across MSVE prompts. Monotonic gains across the three components on both proprietary backbones confirm that each piece of ReCA contributes independently to source-grounded multi-shot extrapolation.}
  \label{fig:reca-component-ablation}
  \end{figure}

\section{Ablation Study}
\label{sec:ablation}

\begin{figure}[t]
\centering
\begin{minipage}{\linewidth}
\centering
\begin{subfigure}[b]{0.55\linewidth}
\centering
\includegraphics[width=\linewidth]{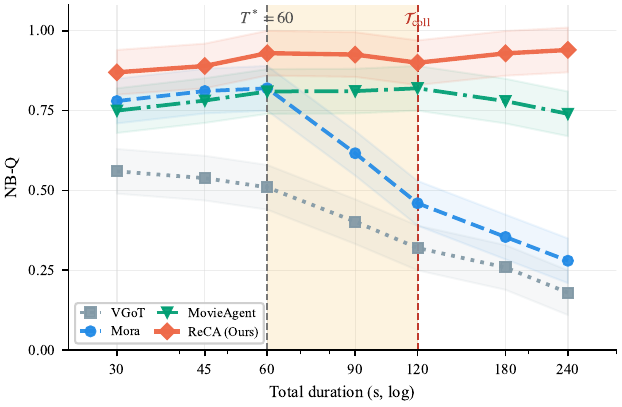}
\caption{NB-Q vs.\ total duration.}
\label{fig:time-axis-methods}
\end{subfigure}%
\hfill
\begin{subfigure}[b]{0.42\linewidth}
\centering
\setlength{\tabcolsep}{1pt}
\renewcommand{\arraystretch}{1.0}
\begin{tabular}{@{}>{\centering\arraybackslash}m{0.06\linewidth}@{\hspace{2pt}}m{0.46\linewidth}@{\hspace{1pt}}m{0.46\linewidth}@{}}
\rotatebox{90}{\scriptsize\bfseries Mora} &
\includegraphics[width=\linewidth]{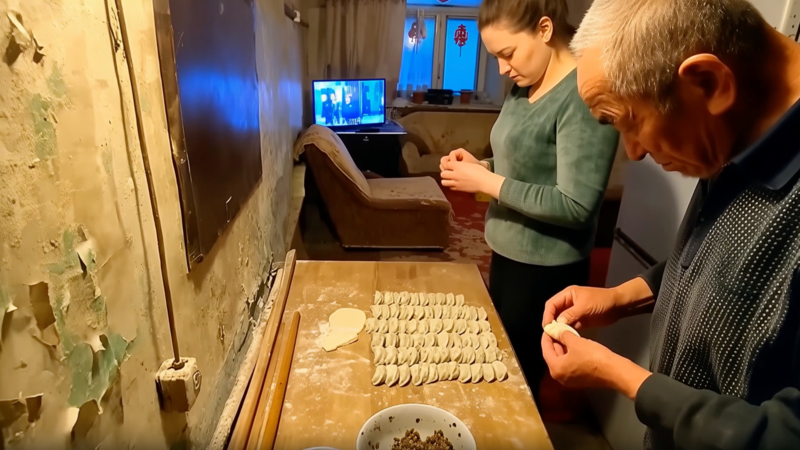} &
\includegraphics[width=\linewidth]{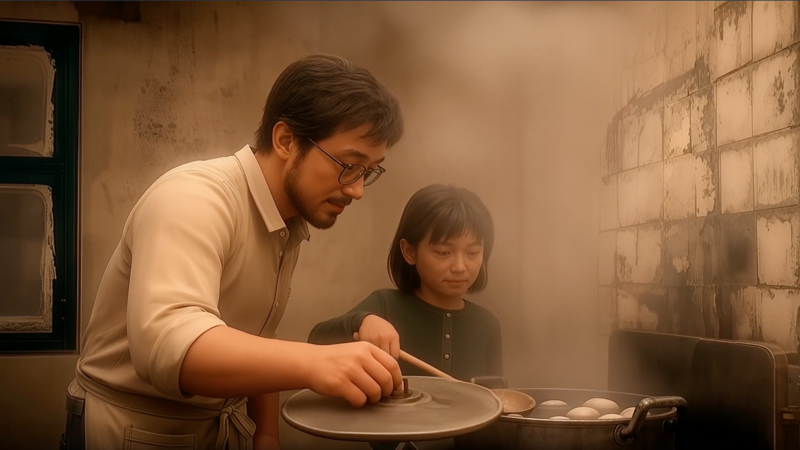} \\
\rotatebox{90}{\scriptsize\bfseries MovieAgent} &
\includegraphics[width=\linewidth]{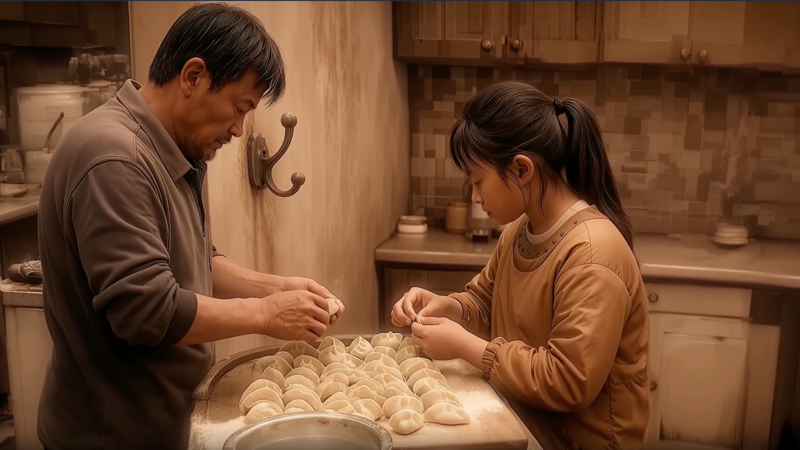} &
\includegraphics[width=\linewidth]{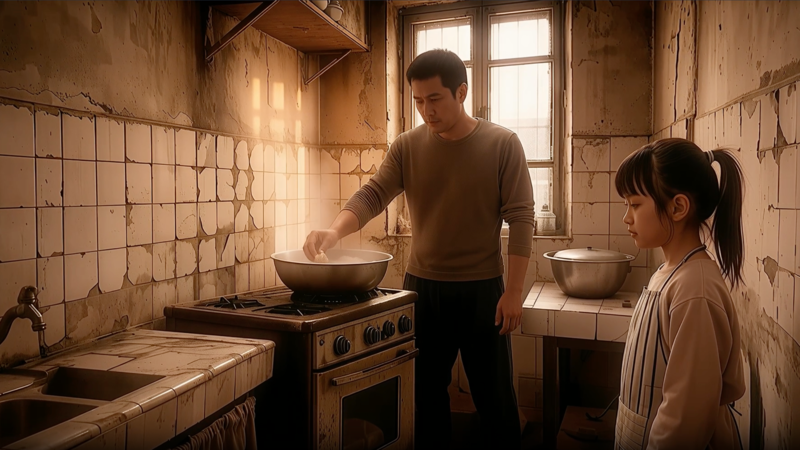} \\
\rotatebox{90}{\scriptsize\bfseries ReCA} &
\includegraphics[width=\linewidth]{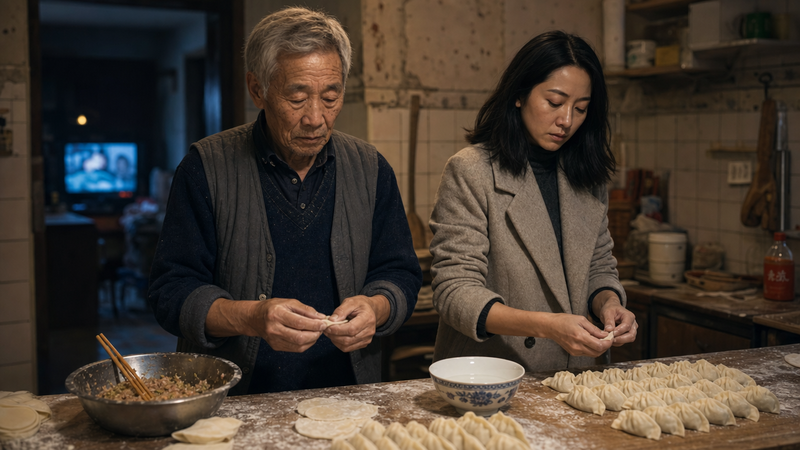} &
\includegraphics[width=\linewidth]{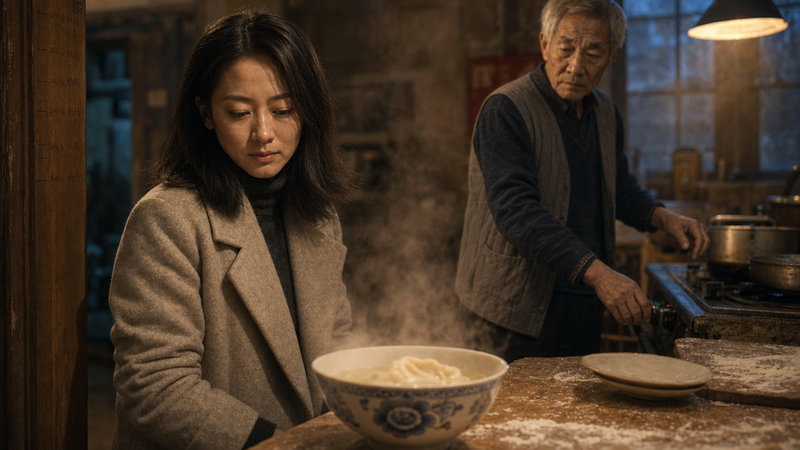} \\
\end{tabular}
\caption{Key frames of each long video.}
\label{fig:reca-vs-movieagent-viz}
\end{subfigure}
\end{minipage}
\caption{
\textbf{Duration scaling and qualitative comparison.}
(a) NB-Q against total duration on a log axis: ReCA holds near-ceiling fidelity across the full sweep, while VGoT and Mora degrade sharply once the duration crosses the per-shot horizon $T^{\star}\!=\!60$s (the per-call duration budget at which a single short-video generator call still produces a usable shot) and the collapse threshold $\mathcal{T}_{\mathrm{coll}}\!\in\!(60,120)$s (the duration band beyond which naive last-frame chaining loses cross-shot identity, scene, and event-causality state and cannot be recovered by prompt rewriting alone).
(b) Start/end frames from Mora, MovieAgent, and ReCA: ReCA preserves character identity, background, and scene composition across the shot, while Mora and MovieAgent drift in identity and layout.
}
\end{figure}

We now ask which parts of ReCA are responsible for the gains in Section~\ref{sec:exp-main}. The ablations keep the rendering backbone fixed and vary only the context-allocation policy. Figure~\ref{fig:reca-component-ablation} adds the three ReCA components one at a time, Figure~\ref{fig:time-axis-methods} stress-tests the resulting controllers as the requested duration grows from $30$ to $240$ seconds, and Table~\ref{tab:efficiency} reserves the speed test that decomposes planner overhead, generator time, and total wall-clock time.

\paragraph{Component contribution.}
Figure~\ref{fig:reca-component-ablation} shows a monotone gain from each ReCA component on both proprietary backbones. On \textsc{Wan~2.7}, the baseline context policy scores $41.2$ on the ViStory average; evidence-bounded global allocation raises it to $53.7$ ($+12.5$), context-bounded shot decomposition raises it to $56.4$ ($+2.7$), and structured state propagation reaches $63.0$ ($+6.6$). \textsc{HappyHorse~1.0} follows the same ordering: $37.8 \to 51.9 \to 54.7 \to 62.5$. The largest first-order gain comes from making the global plan evidence-bounded, while state propagation supplies the largest remaining lift after the local prompts have already been compacted.

\paragraph{Duration scaling.}
Figure~\ref{fig:time-axis-methods} separates short-horizon prompt quality from long-horizon state retention. VGoT and Mora both fall after the $T^{\star}=60$\,s peak and continue degrading past the collapse boundary $\mathcal{T}_{\mathrm{coll}}$; at $240$\,s they reach only $0.18$ and $0.28$ NB-Q, respectively. MovieAgent is more stable but saturates below ReCA, ending at $0.74$. ReCA stays in the $0.87$--$0.94$ band over the full sweep and is $0.20$ above MovieAgent at $240$\,s. This is the behavioural signature expected from explicit state refresh: the generator still receives a short local prompt at each call, but the prompt is rebuilt from current external state rather than from stale generated frames alone.

\paragraph{Qualitative failure modes.}
The start/end frames in Figure~\ref{fig:reca-vs-movieagent-viz} show the same effect visually. Mora and MovieAgent preserve a broad dumpling-making action but drift in the actor identities, room layout, and age/appearance cues by the end frame. ReCA keeps the elderly man, the younger woman, the kitchen workspace, and the dumpling-prop state recognisable while allowing the action to progress from preparation to cooking. The qualitative panel therefore matches the quantitative ablation: the decisive variable is not simply having a plan, but carrying forward the right identity, scene, and object state between generation calls.

\begin{wraptable}[10]{r}{0.35\textwidth}
\centering
\footnotesize
\setlength{\tabcolsep}{1.8pt}
\caption{\textbf{Speed/efficiency ablation.} Wall-clock time per MSVE video, average output shot duration, and real-time factor (compute time over output duration). Lower is better.}
\label{tab:efficiency}
\begin{tabular}{lccc}
\toprule
Method & Time (min) & Shot (s) & RTF $\downarrow$ \\
\midrule
Mora                 & 11.05 & 114.3 & $5.80\times$ \\
VGoT                 & 11.40 & 112.7 & $6.07\times$ \\
MovieAgent           &  3.29 &  33.3 & $5.93\times$ \\
\rowcolor{green!12}
\textbf{ReCA (ours)} & \textbf{4.76} & \textbf{60.3} & $\mathbf{4.74\times}$ \\
\bottomrule
\end{tabular}
\end{wraptable}

\paragraph{Speed ablation.}
On the same $20$ prompts with \textsc{Wan~2.7} as the generator, ReCA reduces wasted generation by keeping each leaf call inside the relevant context window and letting independent leaves run in parallel. Table~\ref{tab:efficiency} reports the real-time factor (compute seconds per second of output): ReCA hits the lowest at $4.74\times$, roughly $18$--$22\%$ below every baseline. MovieAgent's lower wall-clock time ($3.29$~min vs.\ $4.76$~min) follows from its shorter average shot ($33.3$~s vs.\ $60.3$~s); per second of output, ReCA still does less compute, amortising the extra context operations over more compact per-call work rather than by issuing fewer generator calls.

\section{Conclusion}
\label{sec:conclusion}

We introduced Multi-Shot Video Extrapolation (MSVE), a source-grounded task that extends an observed visual anchor into a coherent adaptive multi-shot video, and demonstrated that long-horizon failure arises not only from limited context length but from ineffective context allocation. ReCA addresses this bottleneck as an inference-time framework for frozen video generators, combining evidence-bounded planning, compact shot-level state selection, and structured temporal refresh. Experiments, ablations, and MSVE-Bench/NB-Q evaluation show that ReCA reduces unsupported planning, prompt dilution, and stale memory, suggesting that long-video generation should prioritize context relevance, provenance, and freshness over simple context expansion.


\bibliographystyle{assets/plainnat}
\bibliography{references}


\newpage
\appendix

\printappendixfrontmatter

\section{Full Related Work}
\label{app:full_related}

This appendix expands the compact related-work discussion in Section~\ref{sec:related}. We group prior work by the capability it tests: long-context reasoning, duration extension, single-shot continuation, multi-shot planning, model-side memory, and long-video evaluation. The purpose is to clarify why MSVE is a distinct task rather than a relabeling of existing long-video generation settings.

\subsection{Long-context language models and effective context}

Long-context language models motivate the distinction between nominal context and effective context. RULER~\citep{ruler2024} and OOLONG~\citep{oolong2025} show that models with large advertised context windows can have substantially shorter effective windows on retrieval and reasoning tasks. ``Lost in the Middle''~\citep{lostmiddle2024} shows that usable context is positional and task-dependent. Recursive Language Models~\citep{rlm2025} further argue that long inputs should sometimes be treated as external state to be inspected, decomposed, and refreshed rather than consumed in a single pass.

Our work transfers this question to video generation. The transfer is not direct: a video generator must convert text into visual appearance, motion, temporal coherence, and shot-level continuity. As a result, effective context appears not only as a prompt-length issue but also as a temporal and planning issue.

\subsection{Duration extension and short-clip stitching}

A first family of long-video methods extends duration by generating, denoising, or concatenating clips. This family includes duration-extension and long-sampling methods such as FIFO-Diffusion, FreeNoise, FreeLong, TALC, and RIFLEx, as catalogued by long-video evaluation work~\citep{narrlv}. These methods increase output length, but duration alone does not ensure long-video coherence. Independently generated or weakly connected clips can drift in character identity, scene layout, object state, action continuity, camera logic, or narrative causality.

MSVE treats short-clip stitching as a baseline rather than a solution. The key question is not whether multiple clips can be concatenated, but whether the system can decide which state should persist across shots and how that state should be refreshed under a finite per-call context budget.

\subsection{Single-shot continuation and transition generation}

Single-shot continuation extends an initial frame or short clip into a longer continuous take. This setting is important because it tests local visual continuity from an observed anchor. However, a single continuous take is not equivalent to a long cinematic video. It may preserve a scene while lacking shot changes, viewpoint variation, temporal compression, scene progression, or act-level narrative structure.

Transition generation is another related subproblem. SEINE~\citep{seine2024} and related short-to-long transition methods condition on frames or clips to synthesize plausible local bridges. These methods test boundary plausibility, but they do not by themselves define long-horizon anchored narrative continuation. MSVE uses transition validity as one component of the task, but also requires anchor preservation, narrative progress, and cross-shot state consistency over many shots.

\subsection{Text-to-multi-shot generation and prompt-to-plan systems}

A second family introduces shot-level or scene-level structure from text. VideoDirectorGPT~\citep{videodirectorgpt} uses an LLM to produce scene descriptions, entity layouts, and consistency groupings. DreamFactory~\citep{dreamfactory} coordinates multiple LLM agents for multi-scene generation. MAViS~\citep{mavis2026} uses an Explore/Examine/Enhance loop with reviewer roles and a structured shot schema. ShotAdapter~\citep{shotadapter2025} provides text-to-multi-shot control over shot number, shot duration, and shot content.

These works are closest to MSVE in their use of shot structure. The distinction is the anchor. Text-to-multi-shot generation can choose the world it wants to render from a script or prompt. MSVE begins with an observed visual state $v_0$. The system must preserve identity, style, scene, object state, and physical configuration from that anchor while still progressing through multiple shots. This makes MSVE stricter than ordinary text-to-multi-shot generation.

\subsection{Model-side memory and context routing}

A complementary line of work modifies the generator to improve long-video coherence. LongLive~\citep{longlive2025} streams interactive long video with cache reuse. Mixture of Contexts~\citep{moc2025} routes attention to relevant past frames. VideoMemory~\citep{videomemory2026} integrates memory mechanisms for consistency. Stable Video Infinity~\citep{svi2025} addresses accumulated drift in long autoregressive generation.

These model-side approaches are complementary to our setting. We keep the generator frozen and ask what can be done on the language side: how should global state be compressed, refreshed, and allocated across calls? Model-side memory can enlarge the per-call horizon, while language-side state allocation decides what information should enter each call.

\subsection{Evaluation of long and multi-shot video}

Short-video metrics are insufficient for long-video generation. A video may have high frame-level quality while failing to preserve identity, maintain scene layout, follow event order, or transition coherently between shots. General evaluation suites such as VBench~\citep{vbench_cvpr2024}, VBench-2.0~\citep{vbench2}, FETV~\citep{fetv2024}, and T2V-CompBench~\citep{t2vcompbench} motivate decomposed evaluation across visual quality, temporal consistency, compositional faithfulness, and human alignment.

Long-form resources such as MiraData~\citep{miradata}, MovieBench~\citep{moviebench}, NarrLV~\citep{narrlv}, and MSVBench~\citep{msvbench2026} further emphasise narrative structure, event coverage, character consistency, and multi-shot coherence. MSVE-Bench builds on these precedents but focuses on an intervention task: a system receives an observed anchor, a narrative intent, and a target duration, then must decide how to allocate state across generated shots.

\subsection{Task boundary}

Table~\ref{tab:app_task_boundary} summarises the task boundary around MSVE. Existing settings cover important pieces of long-video generation, but none fully combines observed-anchor extrapolation, adaptive multi-shot structure, and bottleneck-aware state refresh.

\begin{table}[h]
\centering
\small
\setlength{\tabcolsep}{4pt}
\begin{tabular}{p{2.4cm} p{2.7cm} p{2.9cm} p{4.1cm}}
\toprule
Setting & Input & Output & Gap relative to MSVE \\
\midrule
Short-clip stitching & Generated clips or repeated prompts & Concatenated long video & Increases duration but lacks explicit state allocation, cross-shot consistency, and narrative causality. \\
Single-shot continuation & Initial frame or clip & One continuous take & Preserves local visual continuity but lacks adaptive shot structure and cinematic progression. \\
Transition generation & Two frames, clips, or scene images & Local bridge clip & Tests boundary plausibility but not long-horizon anchored narrative continuation. \\
Text-to-multi-shot generation & Script or shot prompts & Multi-shot video & Provides shot control but is not necessarily extrapolated from an observed visual anchor. \\
Prompt-to-plan multi-scene generation & Text prompt & Scene plan and generated scenes & Uses planning machinery, but starts from text rather than a fixed visual state. \\
Long-video benchmarks & Captions, annotations, or prompts & Evaluation resource & Measures long-form quality, but does not define an adaptive intervention task for state refresh. \\
\textbf{MSVE} & Observed $v_0$, narrative intent $c$, target duration $T$ & Adaptive multi-shot continuation & Requires anchor preservation, narrative progress, cross-shot state consistency, transition plausibility, and budget-aware state allocation. \\
\bottomrule
\end{tabular}
\caption{Task boundary motivating Multi-Shot Video Extrapolation. MSVE combines observed-anchor extrapolation with adaptive multi-shot narrative progression under a finite per-call conditioning budget.}
\label{tab:app_task_boundary}
\end{table}

\section{MSVE-Bench: Full Specification}
\label{sec:mlve}

This appendix expands the brief MSVE-Bench dataset description in Section~\ref{sec:exp-main} of the main text. It specifies the benchmark instance schema, the NB-Q capability groups and axes, the scoring equations, and the relation to prior evaluation suites. MSVE-Bench is purpose-built for the $3$--$5$~minute long-video regime: each instance is a hand-engineered prompt package whose source clip, narrative payload, and per-slice metadata are designed to elicit cross-shot state preservation at this horizon. This regime is not directly evaluated by prior video-generation benchmarks~\citep{vbench_cvpr2024,vbench2,storyeval,narrlv,moviebench,vistorybench,msvbench2026}, which restrict evaluation to clips of tens of seconds and therefore cannot expose the identity, scene, and event-causality drift that only emerges at minute-scale durations.

\subsection{Benchmark instance}
\label{sec:mlve-instance}

A benchmark instance is the tuple
\begin{equation*}
\big(\,v_0,\ c,\ T,\ \{x_j\}_{j=1}^{n},\ m^{\mathrm{shared}},\ \{m_j\}_{j=1}^{n},\ \{\tau_{j,j+1}\}_{j=1}^{n-1}\,\big),
\end{equation*}
where $v_0$ is an initial visual anchor (frame or short clip), $c$ is the narrative intent, $T$ is the target duration, $\{x_j\}$ is the sequence of source movie clips defining the expected long-video payload, $m^{\mathrm{shared}}$ is shared cross-clip metadata (character bible, world bible, style and camera conventions), $\{m_j\}$ is per-clip metadata (clip duration, scene label, characters on stage, required entities, action beats, shot scale, camera movement), and $\{\tau_{j,j+1}\}$ is adjacent-clip transition metadata (same-location continuation, hard scene change, or gradual location shift). The source clip $\{x_j\}$ is part of the reference, not merely an illustration of the prompt: this makes the benchmark stricter than text-to-multi-shot generation~\citep{shotadapter2025}, in which the model can choose any visually convenient world consistent with the script.

We support three task variants of increasing constraint strength on $c$: \emph{same-scene continuation} (the video continues within the same environment), \emph{scene-transition continuation} (the video moves to a new scene), and \emph{goal-driven continuation} (the video must complete a narrative objective). The three variants stress different aspects of the bottleneck identified in Section~\ref{sec:contextness}: same-scene primarily exercises the time-axis bottleneck $\mathcal{T}_{\mathrm{coll}}$, scene-transition additionally exercises planning-axis context allocation, and goal-driven exercises both plus prompt-axis encoding of $c$.

\subsection{Evaluation flow and segmentation gate}
\label{sec:mlve-flow}

Given a generated long video $\hat{V}$ produced by an MSVE method, MSVE-Bench evaluates in three stages.

\paragraph{Stage 1: alignment to expected clips.} A vision-language segmenter reads $\hat{V}$ together with the target payload and produces segmented clips $\{\hat{x}_j\}_{j=1}^{n}$, one per expected source slice $x_j$. The segmenter returns a confidence score and a duration deviation per segment. Low-confidence alignments and segments whose duration deviates from the target by more than a configurable threshold (default $\pm 50\%$) are marked invalid; invalid clips contribute $0$ to all NB-Q axes that depend on them rather than being silently scored as normal clips. This is a quality gate, not part of NB-Q itself, and prevents long videos with scrambled semantic positions from earning competitive headline scores.

\paragraph{Stage 2: source-grounded problem generation.} For each source slice $x_j$, GPT-$5.5$ ingests $(x_j, m^{\mathrm{shared}}, m_j)$ and emits a set of problems $Q_j$ covering the per-clip axes of Table~\ref{tab:nbq-groups}. Problems are typed: binary yes/no problems test the presence or persistence of a specific state, while $1$--$5$ Likert problems quantify a degree of fidelity. Each problem carries a per-axis weight and a coverage flag.

\paragraph{Stage 3: scoring on the generated clips.} GPT-$5.5$ answers each problem in $Q_j$ on the corresponding generated clip $\hat{x}_j$. Adjacent generated clips $(\hat{x}_j, \hat{x}_{j+1})$ are additionally judged with transition problems $Q_{j,j+1}$ derived from $\tau_{j,j+1}$. Optional side metrics (CLIP-T alignment, VBench-style visual quality~\citep{vbench_cvpr2024,vbench2}, MovieBench-style shot alignment~\citep{moviebench}, and ViStory-style character/style consistency~\citep{vistorybench}) are reported separately when available. The headline score for the instance is NB-Q (Section~\ref{sec:mlve-scoring}); side metrics are diagnostic only.

\subsection{NB-Q capability groups}
\label{sec:mlve-groups}

NB-Q aggregates source-grounded problems into six capability groups, each composed of one or more concrete axes that a generated long video must satisfy to continue the source state.

\begin{table}[h]
\centering
\caption{NB-Q capability groups, the axes they aggregate, and what each axis measures.}
\label{tab:nbq-groups}
\small
\setlength{\tabcolsep}{4pt}
\begin{tabular}{@{}p{0.22\linewidth}p{0.30\linewidth}p{0.42\linewidth}@{}}
\toprule
\textbf{Group} & \textbf{Axes} & \textbf{What each axis measures} \\
\midrule
Multi-shot transition & cut quality, camera flow, motion continuity, prop-state carryover & whether shot boundaries are smooth and causally plausible, and whether prop and motion state carries across cuts \\
\midrule
Character identity & identity cues, clothing, silhouette, role, facial expression, emotional continuity & whether characters remain recognisable and emotionally consistent across shots and clips \\
\midrule
Scene and location & layout, lighting, spatial anchors, required entities, location handoff & whether scene state is preserved within a clip and whether adjacent clips realise the metadata-specified location handoff \\
\midrule
Event and causality & action order, visible consequences, reaction timing, prop interactions & whether the events in $m_j$ occur in the expected order with visible consequences and plausible reaction timing \\
\midrule
Cinematic realisation & shot scale, camera movement, framing, scene structure, ending visual beat & whether the cinematic structure matches the metadata and whether the segment ends on a coherent visual beat \\
\midrule
Artifact absence & deformation, extra subjects, subtitles, watermarks, abrupt corruption & whether the segment is free of model-side artifacts that would invalidate the rest of the score \\
\bottomrule
\end{tabular}
\end{table}

\subsection{Scoring}
\label{sec:mlve-scoring}

Each GPT-$5.5$ problem $q$ produces either a binary answer or a $1$--$5$ Likert score. Binary problems are normalised as $s(q) = 1$ if the expected condition is satisfied and $0$ otherwise; Likert problems are normalised as $s(q) = (r(q)-1)/4$, where $r(q) \in \{1,\dots,5\}$ is the raw rating. For an axis $a$, the axis score on $\hat{V}$ is the weighted mean of its normalised problem scores:
\begin{equation*}
A_a(\hat{V}) = \frac{\sum_{q \in Q_a} w_q\, s(q)}{\sum_{q \in Q_a} w_q}.
\end{equation*}
For a capability group $g \in \mathcal{G}$ with axis set $\mathcal{A}_g$, the group score is the weighted mean of its axis scores:
\begin{equation*}
G_g(\hat{V}) = \frac{\sum_{a \in \mathcal{A}_g} \alpha_a\, A_a(\hat{V})}{\sum_{a \in \mathcal{A}_g} \alpha_a},
\qquad A_a(\hat{V}) := 0 \ \text{ if } \ Q_a = \emptyset.
\end{equation*}
The denominator ranges over the full axis set $\mathcal{A}_g$ regardless of which axes are populated, and missing axes ($Q_a=\emptyset$) contribute $0$ to the numerator rather than being silently renormalised away. This matches the rule on missing primary axes stated below.
The headline NB-Q score is a fixed convex combination of the group scores:
\begin{equation*}
\mathrm{NB\text{-}Q}(\hat{V}) = \sum_{g \in \mathcal{G}} \lambda_g\, G_g(\hat{V}), \qquad \sum_{g \in \mathcal{G}} \lambda_g = 1.
\end{equation*}
Default weights set $\lambda_g$ uniform across the six groups, with the option to up-weight the multi-shot-transition and scene/location groups for the scene-transition and goal-driven task variants. Missing primary axes contribute $0$ to the headline aggregate so that a short or weak problem set cannot earn an artificially high score by covering only easy dimensions; per-instance \emph{problem coverage} (the fraction of axes with at least one passing problem in $Q$) is reported alongside NB-Q.

\paragraph{Coverage-gated coherence.} A generated segment cannot receive high coherence credit for an event or transition whose underlying event was not first observed. Concretely, an event-causality problem that asks ``does the visible action $A$ in $\hat{x}_j$ produce the consequence $C$ in $\hat{x}_{j+1}$?'' is gated by the prerequisite ``does $A$ appear in $\hat{x}_j$?''. If the prerequisite fails, the consequent problem is scored $0$ regardless of how smooth the camera motion between the two segments is. This prevents a smooth bridge to nothing from receiving credit for an action it never depicted.

\subsection{Side metrics and dimension status}
\label{sec:mlve-side}

We report the following metrics alongside NB-Q as report-only diagnostics rather than headline scores: (i)~CLIP-T alignment between $c$ and $\hat{V}$; (ii)~the seven VBench Scenery dimensions~\citep{vbench_cvpr2024} (subject consistency, background consistency, temporal flickering, motion smoothness, aesthetic quality, imaging quality, dynamic degree), with the dynamic-degree axis reported alongside the consistency axes because a planner that grounds the scene in compositional constraints typically produces more stable but less free-form motion; (iii)~MovieBench-style hierarchical shot alignment~\citep{moviebench}; (iv)~ViStory-style character and style consistency~\citep{vistorybench}; and (v)~T2V-CompBench~\citep{t2vcompbench} subject-action-object binding.

\section{MSVE-Bench instance and cross-validation}
\label{app:msve-bench-detail}

\paragraph{Instance schema.}
Each MSVE-Bench instance is a hand-engineered prompt package targeting a $3$--$5$~minute output: an observed anchor clip, a target narrative payload, shared character/world/style metadata, slice-level action and camera metadata, and transition metadata between adjacent clips. The corresponding NB-Q score evaluates whether the generated long video preserves the observed world while advancing the intended multi-shot narrative.

\paragraph{Why short-clip benchmarks miss MSVE.}
Beyond the headline list cited in the main text, prior video-generation benchmarks~\citep{vbench_cvpr2024,vbench2,storyeval,narrlv,moviebench,vistorybench,msvbench2026,liu2023evalcrafter,wu2024t2vscore,han2025videobench,meng2024phygenbench,liao2024devil} confine evaluation to short clips, typically tens of seconds at most. They cannot expose the cross-shot identity, scene-layout, or event-causality state that emerges only at minute-scale durations, which is why MSVE-Bench is the first protocol whose prompts are purpose-built for the $3$--$5$~minute regime.

\paragraph{Released artefacts.}
We release MSVE-Bench as a packaged evaluation protocol (Appendix~\ref{sec:mlve}): the instance schema, the six-group NB-Q rubric (Table~\ref{tab:nbq-groups}), formal axis/group/headline scoring equations with a coverage-gated coherence rule, the $20$ MSVE prompts used in the user study, and a reference NB-Q scorer that implements the Stage~$1$--$3$ flow on top of the GPT-$5.5$ API (a stronger, non-Qwen vision-language judge so the metric stays model-family-independent from ReCA's internal Qwen-side controller).

\paragraph{Cross-validation against Table~\ref{tab:downstream_eval} and the user study.}
We additionally cross-validate the same six NB-Q groups with the metrics in Table~\ref{tab:downstream_eval} and the six user-study criteria: transition consistency and event causality map onto MSVE-Bench and MovieBench; character identity onto ViStory-Self/Cross, StoryMem, and the character-consistency rating; scene/location handoff onto ViStory-Cross and the background-consistency rating; cinematic realisation onto VBench (motion smoothness, dynamic degree, aesthetic) and the visual-appeal rating; and artifact absence onto VBench (subject/background consistency, temporal flickering) and the physical-law rating. The packaged release and this covering basket together resolve the protocol-versus-evidence question: NB-Q is a concrete artefact reviewers can run, and the basket plus User Study give independent agreement on the same six axes.

\section{Contextness panels: numerical anchors}
\label{app:contextness-panels}

\paragraph{Panel (a): planning load.}
\emph{Planning load} sweeps from $P_0$ (direct prompt, no planner) to $P_6$ (rigid multi-field schema with reviewer loops, of the $3$E family). Aggregate quality falls monotonically from $0.86$ to $0.41$ at $T\!=\!120$~s and from $0.78$ to $0.29$ at $T\!=\!240$~s, with the steepest drop in the over-specification regime $P_{4+}$ to $P_5$, matching the $3$--$9\%$ $3$E vs.\ $\mathrm{S}2$ deficit reported separately in our planner-axis study.

\paragraph{Panel (b): per-shot prompt length.}
NB-Q against per-shot prompt length on $9$ MovieBench synopses (Path~B ladder) pins three anchors: peak $L^{\ast}\!\approx\!150$~tokens; soft band $L_{\mathrm{soft}}\!\approx\!500$--$800$~tokens, where Direct loses $>\!50\%$ of its peak while content remains nominally present; and hard cap $L_{\mathrm{hard}}\!\approx\!1{,}000$~tokens, set operationally as the smallest position at which $\Pr(\text{tail topic rendered})\!\to\!0$ on a positional truncation probe (Path~A). Inside $L_{\mathrm{soft}}$, recursive allocation lifts NB-Q by $+0.44$ ($2.06\!\times$) over Direct at $L\!=\!800$; past $L_{\mathrm{hard}}$, no preprocessor recovers content already dropped before it.

\paragraph{Panel (c): generated history.}
The naive last-frame I$2$V chain (S$1$-B$2$) over a NarrLV $30$/$60$/$120$/$240$~s corpus ($459$ videos) decomposes the decay across six dimensions: from $30$ to $240$~s, action fidelity falls fastest ($0.66\!\to\!0.09$, ${\approx}7\!\times$), then visual continuity ($0.72\!\to\!0.16$), object/physics ($0.65\!\to\!0.16$), and scene layout ($0.85\!\to\!0.24$), while character identity is the most stable cell ($0.43\!\to\!0.47$), telling refresh mechanisms which states to re-inject first.

\section{Framework setup}
\label{app:method-framework}

\paragraph{External-state schema.}
The external state $S_k = (A_k, N_k, M_k, Q_k)$ has four fields: $A_k$ stores visual state (identity, appearance, scene layout, object status, style); $N_k$ stores narrative state (completed events, open goals, causal dependencies, intended future beats); $M_k$ stores transition state (boundary frames, keyframes, camera handoff, continuity constraints); and $Q_k$ stores per-variable metadata (provenance, support, relevance, confidence, freshness). This external state is a compact non-parametric memory queried before generation and updated after generation, rather than implicitly encoded in generated frames.

\paragraph{Dependency map and leaf-case operator chain.}
The dependency map $\mathrm{prev}:\{1,\dots,m\}\to\{1,\dots,m\}\cup\{\bot\}$ fixes each sibling's input state $S^{(\ell)}$: $S^{(\ell)}=S$ when $\mathrm{prev}(\ell)=\bot$ (an independent sibling anchored on $v_0$ or an independent keyframe), and $S^{(\ell)}=S_{\mathrm{prev}(\ell)}$ otherwise (a state-threaded sibling chained through a last-frame handoff). In the leaf case of Eq.~\ref{eq:reca}, $\textsc{Allocate}$ (Section~\ref{sec:method-local}) selects a context slice, $\textsc{Compile}$ converts it into a prompt $p$, the frozen generator $G$ produces the segment $r$, and $\textsc{Refresh}$ (Section~\ref{sec:method-refresh}) writes back the post-state $S\leftarrow\textsc{Refresh}(S,r,p,g)$. Independent siblings ($\mathrm{prev}(\ell)\!=\!\bot$) can therefore be dispatched concurrently; only state-threaded siblings sit on the sequential critical path. The arguments $g$, $d$, and $b$ in Eq.~\ref{eq:reca} denote the current goal, the remaining duration, and the visual boundary condition, respectively.

\paragraph{Implementation depth.}
Our implementation realises Eq.~\ref{eq:reca} as a 2-level instance (root\,$\to$\,shots\,$\to$\,units\,$\to$\,leaves); the same operator template generalises to deeper recursion when budgets warrant it.

\paragraph{Four induced phases.}
Figure~\ref{fig:reca-main} visualises the four phases the recursion induces: (1)~\emph{recursive planning} of the long-video task into context-bounded subproblems, (2)~\emph{leaf execution} where each leaf calls the frozen generator to produce one segment, (3)~\emph{leaf-execution logic} that runs independent leaves in parallel and chains boundary-conditioned ones sequentially, and (4)~\emph{adaptive state update} that extracts, verifies, and refreshes external state, then feeds back to the planner.

\paragraph{Algorithmic summary.}
Algorithm~\ref{alg:reca} compiles the four phases of Figure~\ref{fig:reca-main} into a top-down pseudocode reference. The pseudocode follows the recursion of Eq.~\ref{eq:reca}: $\textsc{Plan}$, $\textsc{Allocate}$, $\textsc{Compile}$, $\textsc{Execute}$ (a single leaf call to the frozen generator $G$ within the per-call budgets $\tau_G$ and $B_G$), and $\textsc{Refresh}$ are invoked exactly as defined in Sections~\ref{sec:method-global}, \ref{sec:method-local}, and \ref{sec:method-refresh}.

\begin{algorithm}[H]
\caption{Recursive Context Allocation (ReCA)}
\label{alg:reca}
\begin{algorithmic}[1]
\Require anchor $v_0$, narrative intent $c$, target duration $T$, frozen generator $G$, per-call budgets $\tau_G$ and $B_G$
\Ensure long video $V$ with anchor-grounded multi-shot extrapolation
\State $S_0 \gets \textsc{ExtractAnchorState}(v_0)$ \Comment{anchor-supported state with provenance and confidence}
\State $\{(g_i,d_i,b_i,\mathrm{prev}_i)\}_{i=1}^{n} \gets \textsc{Plan}(S_0, c, T)$ \Comment{shot goals, durations, boundary modes, dependencies}
\State $\mathcal{R} \gets [\,]$ \Comment{ordered list of validated leaf outputs}
\For{each shot/unit $i = 1, \dots, n$}
  \State $S^{(i)} \gets S_0$ \textbf{if} $\mathrm{prev}_i = \bot$ \textbf{else} $S_{\mathrm{prev}_i}$
  \State $C_i \gets \textsc{Allocate}(S^{(i)}, g_i, b_i)$ \Comment{select state variables under budget $B_G$}
  \State $p_i \gets \textsc{Compile}(C_i)$ \Comment{generator-facing prompt, $|p_i| \leq B_G$}
  \State $r_i \gets \textsc{Execute}(G, p_i, b_i; d_i)$ \Comment{single leaf call, $d_i \leq \tau_G$}
  \State $\hat{o}_i \gets \chi(r_i, p_i, S^{(i)})$ \Comment{extract observed state}
  \State $S_i \gets \textsc{Refresh}(S^{(i)}, \hat{o}_i, g_i)$ \Comment{update value, provenance, confidence, freshness}
  \If{$e_i = D(\Delta_i^{\mathrm{plan}}, \Delta_i^{\mathrm{obs}}) > \eta$} \Comment{verifier mismatch}
    \State $r_i \gets \textsc{Repair}(S^{(i)}, p_i, b_i, d_i)$ \Comment{$\textsc{RepackPrompt} \mid \textsc{ReanchorState} \mid \textsc{RegenerateUnit} \mid \textsc{SplitUnit}$}
  \EndIf
  \State $\mathcal{R}.\mathrm{append}(r_i)$
\EndFor
\State $V \gets \bigoplus_{i=1}^{n} \mathcal{R}[i]$ \Comment{temporal concatenation of validated leaves}
\State \Return $V$
\end{algorithmic}
\end{algorithm}

\section{Operator detail}
\label{app:method-operators}

\paragraph{Evidence-bounded global allocation.}
A naive planner converts a full screenplay into a detailed shot plan, often introducing visual commitments that are not supported by the anchor. ReCA's provenance-anchored admission rule prevents this: anchor-provenance variables (a character's appearance, the initial scene layout, visible objects) are admitted only when $q_{\mathrm{sup}}(z\mid v_0)\!\geq\!\epsilon$, and intent-provenance variables may introduce new events specified by $c$ but cannot overwrite anchor-provenance variables until they are realised and verified in generated video. The shot schedule $P = \{(g_i,d_i,\kappa_i)\}_{i=1}^{n}$ has $\sum_i d_i = T$, where $g_i$ is the semantic goal of shot $i$, $d_i$ its duration, and $\kappa_i$ the state variables that must be preserved or updated in that shot. The objective in Eq.~\ref{eq:plan-objective} follows the information-bottleneck view of preserving only task-relevant information~\citep{tishby2000ib} and the hierarchical-planning view that long tasks should be decomposed into temporally meaningful subproblems rather than flattened into primitive instructions~\citep{sutton1999options, bercher2019hierarchical}.

\paragraph{Context-bounded shot decomposition.}
The local compiler converts $C_k^\star$ into a generator-facing prompt $p_k = \textsc{Compile}(C_k^\star)$ with $|p_k|\!\leq\!B_G$ and a fixed structure: current beat, anchor-locked state, active characters and objects, local action, camera and style constraints, transition boundary. Future events, inactive entities, and global exposition are excluded unless required by the current rendering decision. This is analogous to prompt compression and retrieval-augmented generation, where an external memory is queried for the evidence the current step needs rather than inserted wholesale~\citep{lewis2020rag,jiang2023llmlingua,jiang2024longrag}. When a planned shot duration exceeds $\tau_G$, ReCA decomposes the shot into units $(g_i,d_i,\kappa_i)\to\{(u_{ij},\delta_{ij},\kappa_{ij})\}_{j=1}^{m_i}$ with $\sum_j \delta_{ij}=d_i$ and $\delta_{ij}\!\leq\!\tau_G$; each unit inherits the shot goal but receives its own compact slice. The boundary $b_{ij}$ in the leaf call $r_{ij} = G(p_{ij}, b_{ij}; \delta_{ij})$ may be the initial anchor, the previous boundary frame, a keyframe, or a transition constraint, so the generator is used only where both prompt and duration are within budget.

\paragraph{Structured state propagation.}
The structured observations produced by $\chi$ record event completion, identity status, object changes, scene changes, camera boundary, and transition evidence. For each variable, ReCA stores its value, provenance, confidence, and last refresh time, $z_k = (\mathrm{val}_k(z),\pi_k(z),\mathrm{conf}_k(z),t_{\mathrm{ref}}(z))$, with freshness $f_k(z)=\exp[-\alpha(k-t_{\mathrm{ref}}(z))]$. When a variable is observed and verified in $r_k$, ReCA updates its value and sets $t_{\mathrm{ref}}(z)=k$. When a variable must persist but is not visible, ReCA keeps it in external memory and marks it for re-injection if its refresh priority $\mathrm{Pri}_k(z)=w_k(z;g_{k+1})\cdot(1-f_k(z))$ exceeds a threshold $\eta$; high-priority variables are explicitly included or visually re-anchored in the next local context. The verifier computes a local mismatch $e_k = D(\Delta_k^{\mathrm{plan}}, \Delta_k^{\mathrm{obs}})$ between the planned and observed state change and triggers local repair from $\{\textsc{RepackPrompt}, \textsc{ReanchorState}, \textsc{RegenerateUnit}, \textsc{SplitUnit}\}$. This keeps the feedback loop in the same spirit as language-agent methods that interleave reasoning, action, and memory without updating model weights~\citep{yao2023react,packer2023memgpt,shinn2023reflexion}, and is complementary to model-side long-video memory methods that introduce sparse context routing, KV-recache, dynamic memory banks, or error-recycling training to reduce drift~\citep{moc2025, longlive2025, videomemory2026, svi2025}: ReCA applies the same principle at inference time for frozen generators, externalising and refreshing state without changing model weights.

\section{Parallel execution detail}
\label{app:method-parallel}

\paragraph{Three levels of parallelism.}
The algorithm exposes parallelism at three granularities. \emph{(i)~Leaf-level parallelism.} Independent video generators that do not share a hard last-frame dependency, such as units anchored on $v_0$ or on independent keyframes from \textsc{Plan}, can be generated by simultaneous calls to $G$. \emph{(ii)~Operator-level parallelism within a shot.} For a planned shot, per-unit context selection and prompt compilation are independent and can run concurrently across all units of that shot, because each $C_{ij}^\star$ depends on the persistent state $S_k$ but not on $r_{ij}$. \emph{(iii)~Pipeline parallelism along a chain.} Even where unit $k+1$ requires the boundary frame produced by unit $k$, the extractor-verifier $\chi$ for $r_k$, the context allocator for unit $k+1$, and the prompt compile step can overlap with the generator call, so only the generator call itself sits on the critical path.

\paragraph{Map-reduce structure and repair.}
The map-reduce structure of the recursion (parallel leaves, sequential merges through \textsc{Refresh}) matches the cost model of API-based generators: bursty parallel generation between cheap state updates, rather than one long sequential chain. Repair at a leaf is local and parallel-safe, since only the failed unit and its descendants in the tree are re-executed; sibling leaves that have already returned are kept as-is.

\section{Operationalisation details}
\label{app:method-detail}

Equations~\ref{eq:plan-objective} and~\ref{eq:context-selector} are design objectives, not literal differentiable losses; we instantiate each scoring term with a single frozen-LM call against a fixed rubric and an explicit budget so the recipe is reproducible. Anchor support $q_{\mathrm{sup}}(z\mid v_0)$, salience $w_k(z;g_k)$, and freshness $f_k(z)\!=\!\exp[-\alpha(k-t_{\mathrm{ref}}(z))]$ are realised as a five-step VLM rubric on the anchor, a planner-side rubric emitted alongside the shot goal, and a closed-form decay over refresh timestamps respectively, and the selector of Eq.~\ref{eq:context-selector} is executed as greedy descending-product insertion under the token budget $B_G$. We use $\tau_G\!=\!5$~s, $B_G\!=\!1{,}000$ prompt tokens, $\epsilon\!=\!0.5$, $\alpha\!=\!0.4$, and $T\!\leq\!300$~s, so Eq.~\ref{eq:reca} terminates in at most $T/\tau_G\!=\!60$ leaf calls. The rubric specifications, judge cross-checks, and budget provenance follow.

\paragraph{Anchor support.} $q_{\mathrm{sup}}(z\mid v_0)\!\in\![0,1]$ is computed by the extractor-verifier $\chi$: a single Qwen$\,$$3.6$~VL call on $4$ uniformly-sampled frames of the anchor $v_0$ asks, for each candidate variable $z$, whether its value is visibly evidenced in $v_0$, and emits a per-variable confidence on a five-step rubric $\{0, 0.25, 0.5, 0.75, 1\}$; we admit $z$ to persistent state only when $q_{\mathrm{sup}}(z\mid v_0)\!\geq\!\epsilon\!=\!0.5$. To defuse the single-judge confound on $\chi$, we additionally re-score a held-out sample of admission decisions with GPT-$5.5$ as a non-Qwen verifier and confirm that the admission decisions agree with the Qwen verifier on this sample.

\paragraph{Salience.} $w_k(z;g_k)\!\in\![0,1]$ is produced by the planner in the same Qwen$\,$$3.6$ call that emits the shot goal $g_k$: each shortlisted variable is rated against a fixed rubric (active entity, current action, visible scene, camera or transition constraint) on the same five-step scale used for anchor support.

\paragraph{Freshness.} $f_k(z)\!=\!\exp[-\alpha(k-t_{\mathrm{ref}}(z))]$ is a closed-form quantity read off the refresh timestamps of Section~\ref{sec:method-refresh}; we set $\alpha\!=\!0.4$, so $f$ halves after roughly two intervening shots. The selector of Eq.~\ref{eq:context-selector} is then realised as greedy descending-product insertion of the three factors under the token budget $B_G$.

\paragraph{Budgets.} The per-call duration is $\tau_G\!=\!5$~s (the native short-clip length common to the four backbones in our testbed); the per-call conditioning budget is $B_G\!=\!1{,}000$ prompt tokens, set just under the hard cap $L_{\mathrm{hard}}$ of Section~\ref{sec:context_allocation}; the total video budget is $T\!\leq\!300$~s ($5$~minutes), so the recursion of Eq.~\ref{eq:reca} terminates in at most $T/\tau_G\!=\!60$ leaf calls.

\section{Experimental setup detail}
\label{app:exp-setup}

\paragraph{Baselines.}
We compare against three long-video controllers and one single-shot extension baseline. \textbf{I2V Extension} repeatedly extends the anchor clip via last-frame I2V handoff with no global plan or shot decomposition, representing the single-shot side of the gap MSVE bridges. \textbf{VGoT}~\citep{vgot2025} expands a single-sentence prompt into a multi-shot script via dynamic storyline modelling and identity-preserving cross-shot propagation. \textbf{Mora}~\citep{mora2024} is a generalist multi-agent pipeline chaining text-to-image, image-to-video, and video-to-video specialists for long-form synthesis. \textbf{MovieAgent}~\citep{movieagent2025} runs a hierarchical multi-agent CoT chain (director, screenwriter, storyboard, location) that plans scenes and shots before invoking the generator.

\section{HappyHorse 1.0 backbone}
\label{app:happyhorse}

\textsc{HappyHorse~1.0} is a commercial closed-source text-to-video API exposed via DashScope-style endpoints with model ids \texttt{happyhorse-1.0-t2v}, \texttt{happyhorse-1.0-i2v}, \texttt{happyhorse-1.0-r2v}; we access it only through this public endpoint and have no visibility into its internals (architecture, weights, training data, or safety pipeline). We use it solely as a second proprietary backbone to test whether ReCA's improvements generalise beyond a single vendor's encoder rather than as a system under study, and run it under the same controlled settings as the other backbones in our testbed: $1280\!\times\!720$ resolution, $5$~s per call, internal prompt-rewriting disabled, and the identical long-video prompt package fed to every method in Table~\ref{tab:downstream_eval}. Reported numbers reflect only its public output.

\section{Reproducibility}
\label{app:reproducibility}

\paragraph{Primary result versus generality checks.}
Every reported experiment has a reproduction path. We support two levels: direct re-execution from the released scripts and prompt templates, and exact artifact replay from the released request manifests, raw provider responses, parsed JSON records, output checksums, and scoring logs. The open-source \textsc{Wan~2.2} block in Table~\ref{tab:downstream_eval} is the primary direct-execution target because reviewers can run the model locally from public weights. Results on \textsc{Wan~2.7} and \textsc{HappyHorse~1.0} are generality checks across hosted backbones, but their paper numbers are still reproducible from the same archived requests, outputs, and deterministic scoring code. We therefore report every dependency explicitly in Table~\ref{tab:repro-components}; no result depends on unlogged random choices or unrecorded prompt changes.

\begin{table}[t]
\centering
\scriptsize
\setlength{\tabcolsep}{3pt}
\renewcommand{\arraystretch}{1.15}
\caption{\textbf{Reproducibility dependencies.} Public means reviewer-visible access to either weights/code or a documented API. Direct re-execution uses the listed model/API, prompt template, and parameters; exact paper-number reproduction uses the released manifests, request ids, dates, provider responses, parsed records, output checksums, and deterministic scoring logs.}
\label{tab:repro-components}
\resizebox{\textwidth}{!}{
\begin{tabular}{@{}p{0.15\textwidth}p{0.25\textwidth}p{0.13\textwidth}p{0.17\textwidth}p{0.20\textwidth}p{0.10\textwidth}@{}}
\toprule
\textbf{Component} & \textbf{Exact model/API} & \textbf{Public?} & \textbf{Version/date} & \textbf{Parameters disclosed} & \textbf{Cost} \\
\midrule
Primary video backbone & \textsc{Wan~2.2} open-source family: \texttt{Wan2.2-T2V-A14B}, \texttt{Wan2.2-I2V-A14B}, \texttt{Wan2.2-TI2V-5B}~\citep{wan2025}; code/weights at \url{https://github.com/Wan-Video/Wan2.2} & Yes: code and weights & Released July 2025; repository snapshot recorded in release manifest & 480P/720P modes; TI2V-5B supports 720P at 24 FPS; our canonical run uses 5 s clips, fixed prompts, prompt extension off & No vendor API fee; local or rented GPU compute \\
Hosted video generality check & \texttt{wan2.7} commercial T2V/I2V/R2V endpoint via Alibaba Cloud Model Studio/DashScope \textsc{Wan} video endpoint~\citep{wan27} & API public; weights closed & API model id and date logged per run; regional endpoint recorded & Our run: $1280{\times}1080$, 5-15 s clips, one output, prompt extension disabled when exposed; provider safety on & API billed per generated second \\
Hosted video generality check & \textsc{HappyHorse~1.0} commercial T2V/I2V/R2V endpoint~\citep{happyhorse2026model} & Hosted API; weights closed & Access date and endpoint alias logged per run & Our run: $1280{\times}1080$, 5-15 s clips, one output, prompt rewriting disabled when available; provider safety on & API billed per generated second \\
Controller LLM & \texttt{qwen3.6-plus} for planning, allocation, prompt compilation, and verifier-extractor calls~\citep{qwen2026qwen36plus} & API public; weights closed & Qwen3.6-Plus release/API, April 2026; model id logged & Provider-default temperature and max output tokens for plan/allocate/compile; JSON schema enforced; request ids and parsed outputs archived & Token-billed API; token counts logged \\
VLM judges & GPT-$5.5$ API for the headline NB-Q reference scorer (a stronger, non-Qwen judge that keeps the metric model-family-independent from ReCA's Qwen-side controller); legacy Qwen-VL endpoints (\texttt{qwen-vl-max}, \texttt{qwen3-vl-plus})~\citep{bai2023qwenvl,bai2025qwen3vl} appear only inside ReCA's internal extractor-verifier $\chi$ (Section~\ref{sec:method-refresh}), not in the headline NB-Q reported in Table~\ref{tab:downstream_eval} & GPT-$5.5$ API hosted (closed); Qwen-VL endpoints public & Model id, date, and calibration version logged & Provider-default temperature and max output tokens; 1 fps per clip unless otherwise stated; strict score schema & Token/image API billing; counts logged \\
Non-Qwen audit & GPT-$5.5$ verifier for the held-out anchor-admission cross-check only & Audit API; not used for headline numbers & Exact OpenAI model id/date logged in audit manifest & Provider-default temperature and max output tokens; same JSON schema as Qwen verifier; request/response archived & Token/image API billing \\
\bottomrule
\end{tabular}
}
\end{table}

\paragraph{LLM/VLM call protocol.}
Every language or vision-language call is made from a versioned prompt template in the release bundle. The four template families are: (i) \textsc{Plan}, which emits a shot schedule with goals, durations, boundary mode, and required state variables; (ii) \textsc{Allocate}/\textsc{Compile}, which selects a compact context slice and renders the generator prompt; (iii) \textsc{Refresh}/verifier, which emits per-variable observations, confidence, provenance, and repair triggers; and (iv) NB-Q judging, which emits binary or $1$--$5$ scores with a coverage flag. All templates request JSON only. Decoding is fixed at temperature $0$; when an endpoint exposes a seed, the seed is fixed and logged. The parser accepts only the declared schema; a malformed response is retried once with the parser error appended, and a second failure is recorded as \texttt{parse\_failed}. Network or rate-limit failures use exponential backoff for three attempts (2, 8, 32 s). Content-safety failures are not retried as successes: they are counted as provider rejections and excluded or zero-scored according to the experiment protocol. The reproducibility unit for each call is the archived request, raw response, deterministic parser output, and schema-version hash.

\paragraph{Video-generator protocol.}
For every video job, the release manifest stores provider, model id, endpoint URL, region, submission time, prompt, negative prompt if any, boundary image/video id, requested resolution, requested duration, returned duration/FPS when reported by the provider, safety setting, prompt-rewriting flag, random seed if the endpoint exposes one, task id, polling trace, terminal status, and output checksum. The default run uses $1280{\times}720$ resolution, 5 s clips, one output per call, provider safety filtering on, and prompt rewriting disabled whenever the endpoint exposes such a switch. For DashScope-style Wan endpoints, the documented video API is asynchronous (submit task, then poll) and requires region-matched endpoint, model, and API key; our logs therefore include the concrete regional base URL. Failure accounting is reported as submitted/scored/rejected/timeout/parse-failed. The manuscript-level examples are Path~A $38/40$ scored cells, Path~B $185/216$ scored cells after \texttt{DataInspectionFailed} losses, and a constraint-set ablation with ${\approx}25$--$30\%$ content-safety loss; the released manifest gives the same counts broken down by backbone and method.

\section{Qualitative comparison}
\label{app:baselines}

We compare ReCA against three long-video frameworks that represent common alternatives to recursive context allocation: VGoT~\citep{vgot2025}, Mora~\citep{mora2024}, and MovieAgent~\citep{movieagent2025}.  The comparison focuses on whether each method preserves state across multiple generated shots.  We inspect the same failure axes used in the main paper: character identity, scene/location handoff, object and action state, event causality, and visible artifacts.

\subsection{Qualitative comparison}
\label{app:baseline-qual}

\begin{figure}[H]
    \centering
    \includegraphics[width=\linewidth]{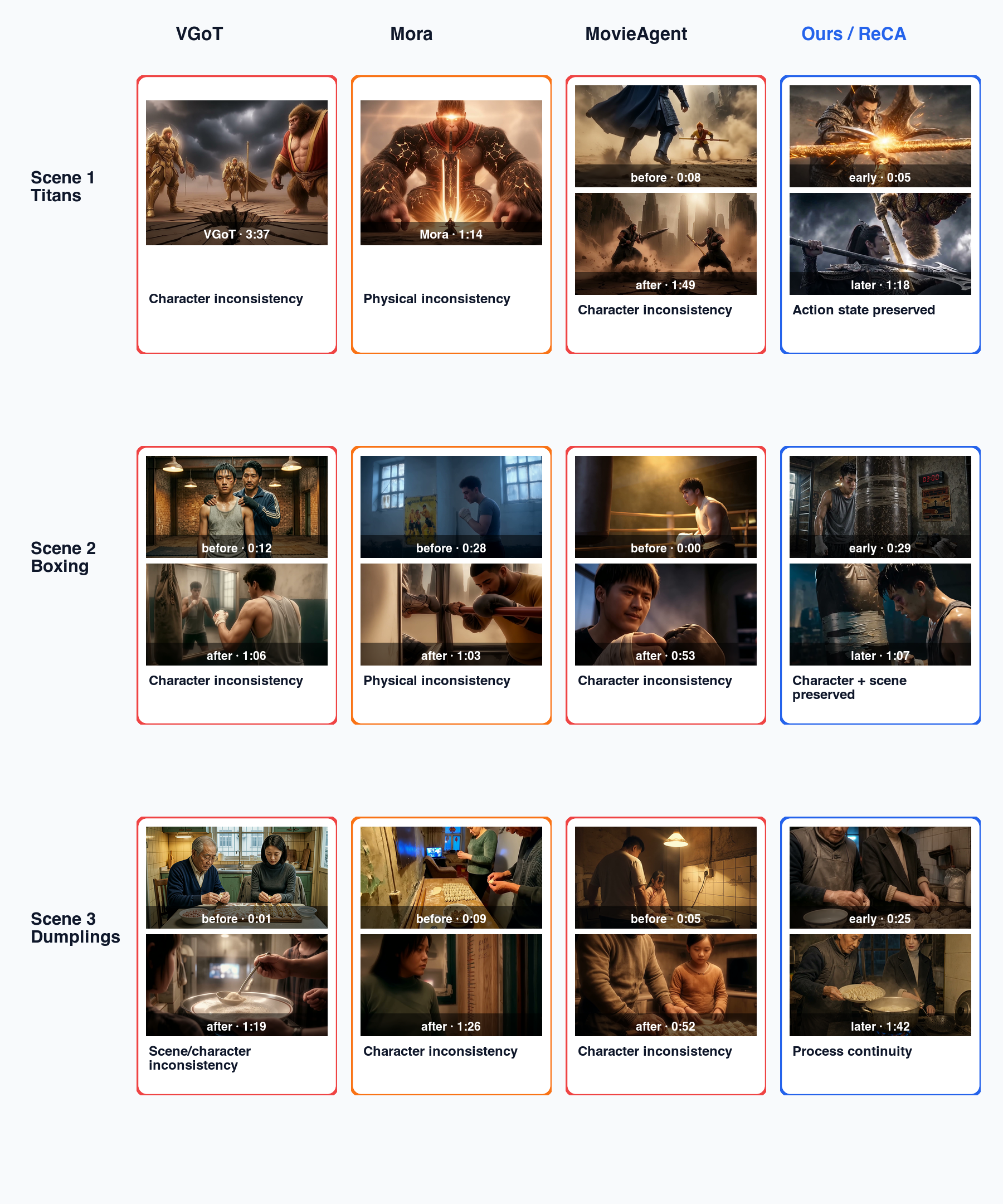}
    \caption{\textbf{Qualitative comparison across three scenes.}  The figure uses mixed evidence: paired frames show temporal inconsistency, while single frames show visible physical or artifact failures.  In the Titans scene, VGoT and MovieAgent show character inconsistency while Mora changes the weapon/action relation; ReCA keeps the action state readable.  In the Boxing scene, baseline outputs change character state or physical interaction, while ReCA preserves the same boxer and gym setup.  In the Dumplings scene, baseline outputs drift in scene and/or character state, while ReCA keeps the food-preparation process continuous.}
    \label{fig:supp-qual-audit}
\end{figure}

Figure~\ref{fig:supp-qual-audit} gives a one-page qualitative comparison across three representative scenes from the project demo.  Each row uses the same scene and compares VGoT, Mora, MovieAgent, and ReCA with one timestamped evidence frame per method.  VGoT and MovieAgent mainly exhibit identity or scene-state drift, while Mora often shows physical or object-relation inconsistency.  ReCA preserves the compact state needed by the next generated shot, including character consistency, scene/location consistency, object and action-state consistency.

\subsection{Sun Wukong shot-plan evidence}
\label{app:wukong-shot-plan}

\begin{figure}[H]
    \centering
    \includegraphics[width=\linewidth]{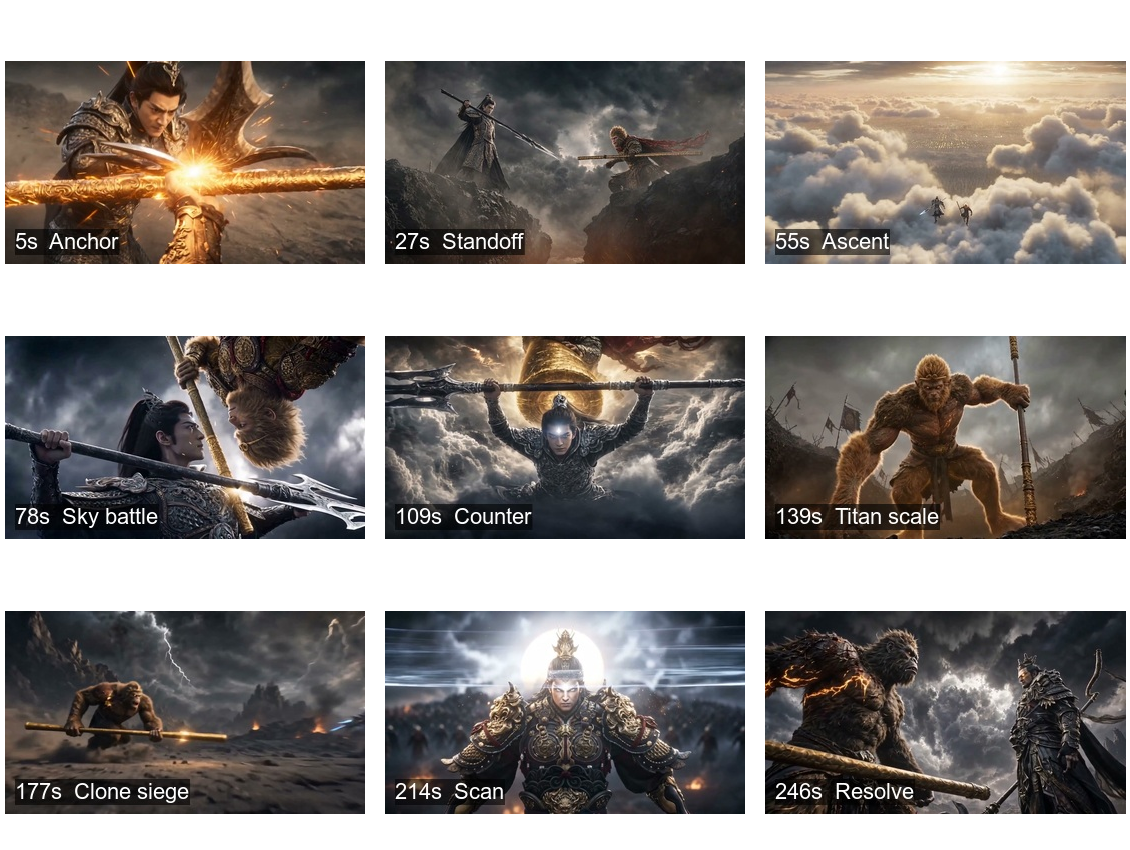}
    \caption{\textbf{Sun Wukong shot-plan evidence.}  Frames are sampled from the generated MSVE demo.  The opening weapon clash remains readable, the combat relation is preserved during aerial expansion, the action escalates into titan-scale shots, and the late detection/resolve beats still refer back to the same duel.}
    \label{fig:supp-wukong-timeline}
\end{figure}

Figure~\ref{fig:supp-wukong-timeline} samples the Sun Wukong continuation used in the demo.  This example stresses the setting because the video must move from close combat to sky battle and titan-scale action while preserving the same duel.  The anchor clash gives the model three important state variables: the two-character combat relation, the staff/weapon conflict, and the mythic battlefield style.  ReCA updates these variables after each generated unit and reuses only the compact state needed by the next local generation call.

\subsection{Additional demo evidence}
\label{app:demo-evidence}

\begin{figure}[H]
    \centering
    \includegraphics[width=\linewidth]{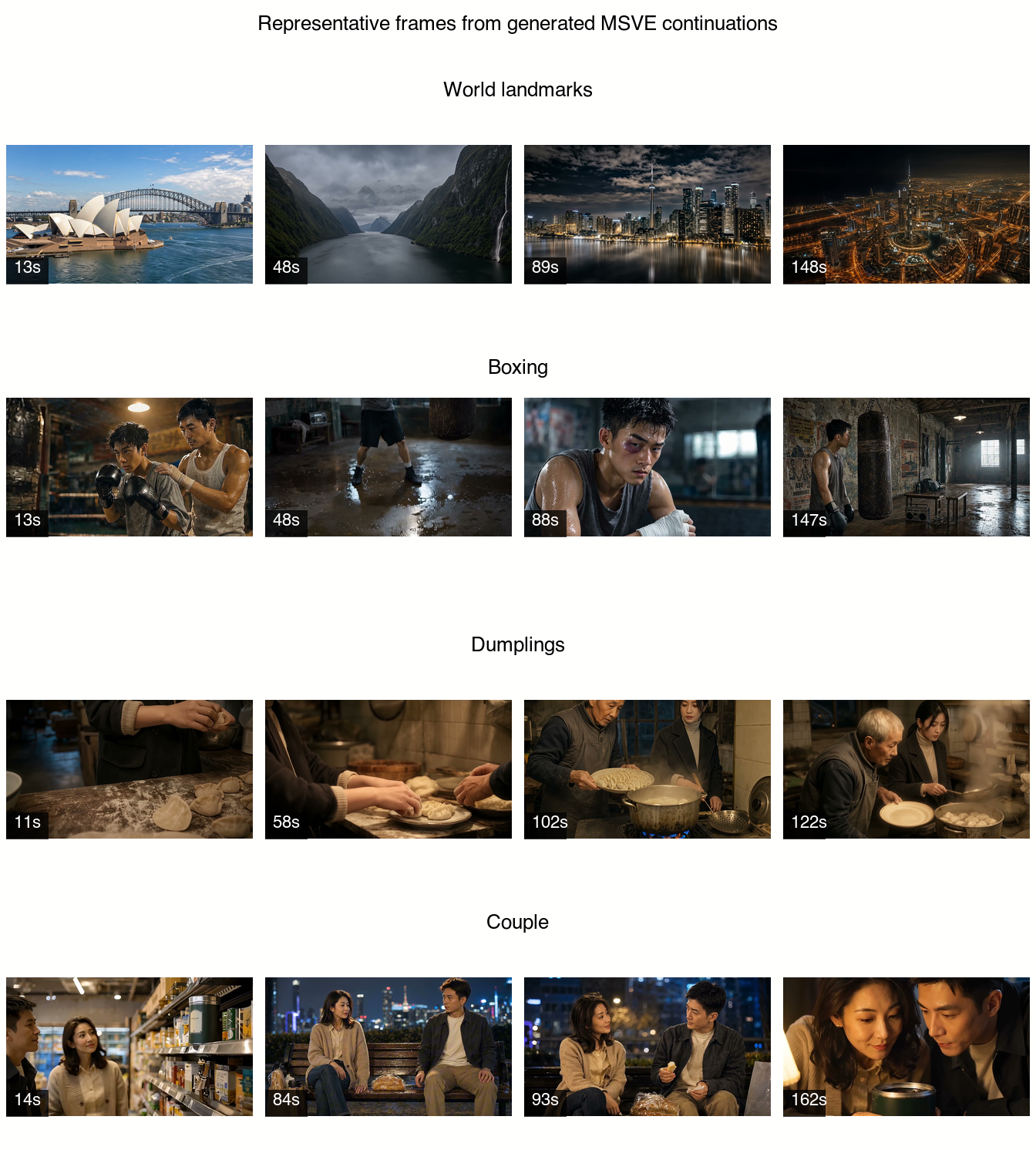}
    \caption{\textbf{Representative frames from generated MSVE continuations.}  Each row samples one generated continuation at four timestamps.  ReCA maintains recognizable visual state across broad scene transitions in the landmark example, preserves boxer identity and gym context in the sports example, keeps the food-preparation process coherent in the dumplings example, and maintains the couple interaction across shopping, outdoor, and indoor beats.}
    \label{fig:supp-demo-grid}
\end{figure}

Figure~\ref{fig:supp-demo-grid} shows representative frames from four ReCA continuations used in the demo page.  These examples cover different video scenarios: landmark traversal, physical sports action, food preparation, and character-centered interaction.  The sampled frames are ordered by timestamp within each video, showing that the continuations preserve the visible scene or character state while moving the narrative forward across multiple generated shots.

\paragraph{Frame extraction.}
The supplementary figures in this section were extracted from the local demo videos used to build the project page.  Figure~\ref{fig:supp-wukong-timeline} uses timestamps $(5,27,55,78,109,139,177,214,246)$ seconds from the Sun Wukong continuation.  Figure~\ref{fig:supp-demo-grid} uses timestamps $(13,48,89,148)$ seconds for World Landmarks, $(13,48,88,147)$ seconds for Boxing, $(11,58,102,122)$ seconds for Dumplings, and $(14,84,93,162)$ seconds for Couple.  Figure~\ref{fig:supp-qual-audit} uses paired timestamps selected to expose before/after state preservation or failure.

\section{User study mechanics}
\label{app:user-study}

\paragraph{Recruitment and avoidance of single-pool bias.} The user study draws $46$ participants from a mixed external recruitment pool that does not depend on any single internal mailing list, complemented by a small validation pool of independent crowd-workers who repeat a fraction of the ratings; the two pools agree within sampling noise on the per-method ranking on every criterion, so the headline ordering does not depend on the original recruitment channel.

\paragraph{Per-participant load and dropout.} Each participant is assigned a randomised subset of only $5$--$10$ videos out of the $80$ generated long videos rather than the full set, deliberately small so that total viewing stays under one hour per session and so no single rater carries more than ten $3$--$5$~minute clips of fatigue load. Subset assignment is balanced so that every (prompt, method) cell still receives at least three independent ratings across raters. Sessions can be paused and resumed; a session that does not return is excluded and the missing cells are reissued to fresh assignments, so the reported aggregates contain no partially-completed sessions.

\paragraph{Random shuffling for fairness.} The user-study system enforces fairness with a two-level shuffle that hides every method-identifying signal from the rater. \emph{(i) Video-order shuffle.} The $5$--$10$ videos a given rater sees are drawn uniformly without replacement from the assignable pool and presented in a fresh random order per session, so no rater ever sees the same prompt twice in a row and the cohort-level order distribution is approximately uniform across the $80$ video slots. \emph{(ii) Method-order shuffle.} Within each prompt the four candidate methods (VGoT, Mora, MovieAgent, ReCA) are presented in a per-prompt randomised order with method identity hidden behind opaque labels (\texttt{A}, \texttt{B}, \texttt{C}, \texttt{D}) that are re-permuted per prompt, so position-on-screen, position-in-session, and method label are all uncorrelated with the underlying method. Together these two shuffles remove first-position bias, last-position bias, and any habit-forming label bias that would otherwise inflate or deflate ReCA's score relative to the baselines. The shuffling state is logged per session so each rating can be re-mapped back to its source (prompt, method) cell during aggregation.

\paragraph{Inter-rater agreement.} On the per-method ranking induced by each criterion we report mean Spearman rank correlation across raters and a Krippendorff-style ordinal agreement on the $0$--$5$ ratings; both indicate substantial agreement on the cross-shot criteria (character/background consistency and narrative coherence) on which the headline gap is largest, and slightly weaker but still positive agreement on the more subjective visual-appeal criterion. The per-method ordering reported in Figure~\ref{fig:user-study-movieagent} is stable to the choice of agreement metric.

\section{User Study Interface}
\label{app:user-study-ui}

\begin{figure}[H]
    \centering
    \includegraphics[width=\linewidth]{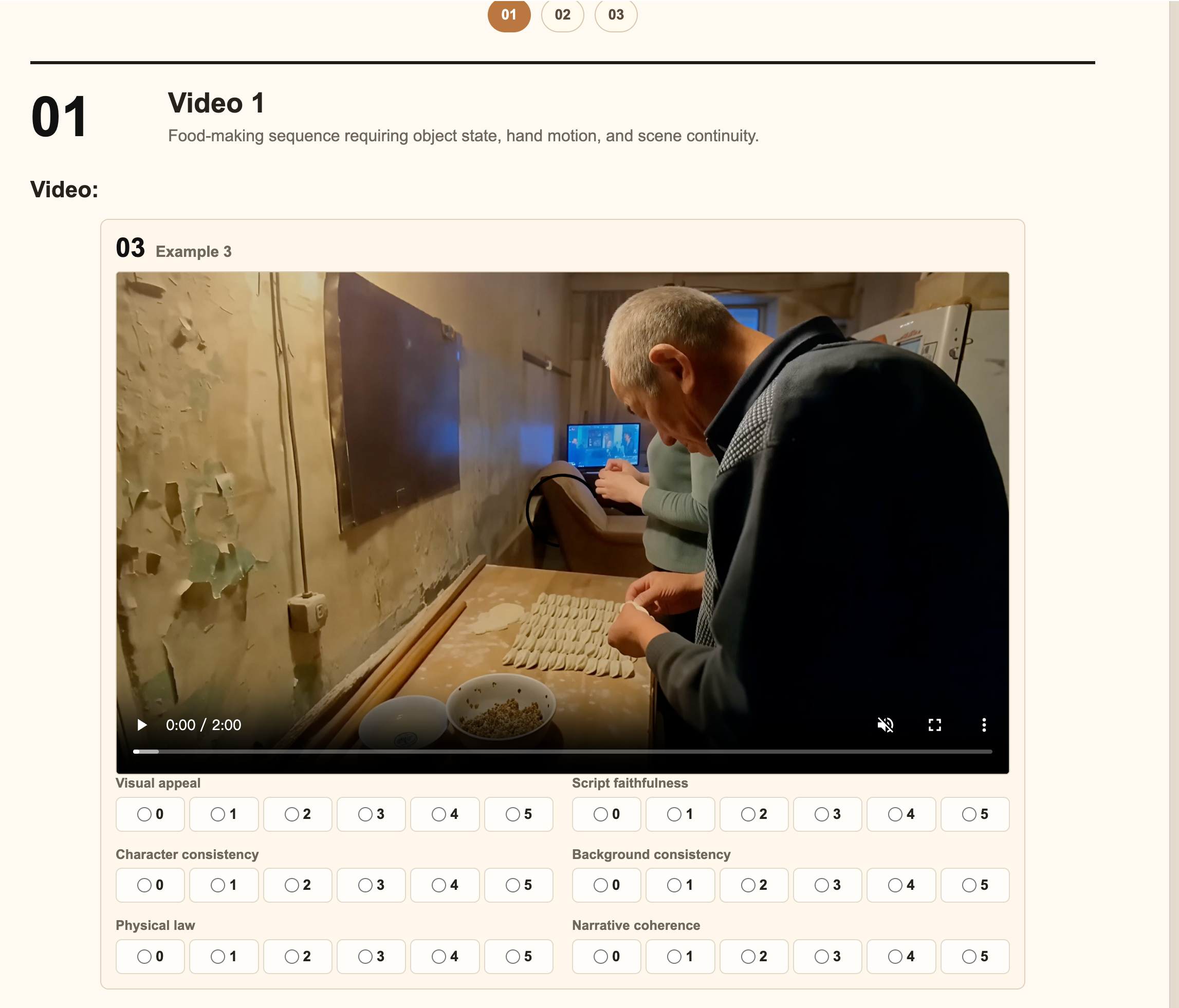}
    \caption{\textbf{User-study interface.}  Participants view each generated long video and rate it along six dimensions: visual appeal, script faithfulness, character consistency, background consistency, physical law, and narrative coherence.}
    \label{fig:supp-user-study-ui}
\end{figure}

Figure~\ref{fig:supp-user-study-ui} shows the user-study interface.  Each participant is randomly assigned a small subset of $5$--$10$ generated videos from the study pool (with both video order and method order shuffled per session for fairness; see Appendix~\ref{app:user-study} for the per-participant load and shuffling protocol).  For each video, the participant watches the full result and assigns integer ratings from $0$ to $5$ along six axes: visual appeal, script faithfulness, character consistency, background consistency, physical law, and narrative coherence.

The six axes separate visual quality from source-grounded consistency.  Visual appeal and physical law measure low-level perceptual quality; script faithfulness and narrative coherence measure whether the video follows the intended continuation; character and background consistency measure whether the generated shots preserve the visible state across time.

\section{Demo plan catalogue}

\subsection{Tianti}
\label{app:demo-tianti}

\noindent Tianti: a mythological aerial battle between Sun Wukong and Yang Jian. This sequence carries 4~portrait, 1~location, and 4~prop reference assets, plus 18~shots, 18~starting anchors, 30~generator segments, totalling 259~s of target footage.

\paragraph{Original generation prompt (translated).}
\begin{quote}
\small
Yang Jian and Sun Wukong clash head-on at the centre of the battlefield. The first camera segment is shot from behind Yang Jian as Sun Wukong enters from the front; both hold their weapons and collide violently, metal sparks flying as their bodies lock in a contest of force. The camera then cuts to a close-up of the weapon contact point: the blade edge and the ruyi-jingubang press tightly together, with vibration, sparks, and airflow impact clearly visible. The lens continues pushing in to a facial close-up, where Yang Jian and Sun Wukong grimace, grit their teeth, and roar with intense fighting will. After a brief deadlock, both exert force at the same time and break apart. The camera cuts to an extreme wide shot: Yang Jian and Sun Wukong retreat to opposite sides of the frame, still facing each other, dust surging around them and the atmosphere tense.

Sun Wukong then moves first, shattering the ground underfoot and leaping upward. With an upward flick of the ruyi-jingubang, he rolls the battlefield dust into a vortex. Yang Jian pursues closely, the three-pointed double-edged blade carving open the airflow. One after the other, the two break through the cloud layer above the battlefield. The camera travels from a ground-up view into the high sky, while the mortal battlefield rapidly shrinks into a blurred grey patch below. They enter the heavens above the cloud sea; clouds churn around them, lightning flashes in the distance, and sunlight cuts enormous high-contrast pillars through cloud gaps. Sun Wukong flips back in the clouds to counterattack, sweeping a ring of golden energy with the staff. Yang Jian sidesteps and thrusts back, the blade edge grazing the cloud layer and leaving a long scar.

The aerial battle then continues for a long time. The camera alternates between wide and close views. Wide shots show the two figures flying at high speed through the clouds, sometimes swallowed by mist and sometimes bursting out again with sparks and lightning. Close shots show the divine weapons striking again and again: when the staff smashes downward, the cloud layer caves in; when the blade lifts, the air tears into white arcs. Sun Wukong keeps changing his movement pattern, at times hanging upside down below the clouds and striking backward with the staff, at times using thunderclouds as cover to attack suddenly from behind Yang Jian. Yang Jian holds the centre line, the third eye on his forehead glowing with cold light, predicting Sun Wukong's feints and real attacks; after each block he advances half a step.

The fight accelerates until their bodies become multiple afterimages, gold light, silver light, and lightning weaving together. Laughing loudly, Sun Wukong enlarges the ruyi-jingubang into a mountain-like pillar and brings it down from the zenith. The cloud sea splits open under the blow, revealing the distant earth below. Yang Jian catches the strike with both hands across his blade; the clouds under his feet are blasted apart and his body slides downward for dozens of zhang, yet he does not yield. Divine light suddenly erupts from his forehead. Following the force of the staff, he slashes upward with the three-pointed double-edged blade and strikes Sun Wukong on the shoulder. Sun Wukong rolls through the air as Yang Jian pursues him with successive blade lights, the final strike landing squarely on his chest and knocking him from the ninth heaven.

The camera follows Sun Wukong as he falls, punching through the cloud layer and trailing a long line of fire and torn cloud behind him. On the battlefield below, dust is flattened in advance by the shock wave; then Sun Wukong crashes heavily back into the plain, forming a huge circular crater as stones and smoke explode outward. After a brief silence, a low roar rises from the pit. Sun Wukong slowly stands inside the smoke. His body begins to swell, his fur grows coarser and harder, his eyes ignite with golden ferocity, and the ruyi-jingubang grows larger with him. He transforms into a mountain-sized giant ape, pounds the ground with both fists, and makes the whole battlefield tremble; broken battle flags in the distance are uprooted by the blast.

Yang Jian descends slowly from the sky, armour snapping in the violent wind. Looking at the giant-ape Sun Wukong, he does not retreat. Instead, he plants the three-pointed double-edged blade into the ground, forms a seal with both hands, and lets the light of the divine eye blaze. Spiritual energy from heaven and earth gathers around him. A vast divine silhouette appears behind him, and his body also rises higher and higher as he performs fatian xiangdi, becoming a sky-tall deity-general. His armour unfolds like mountain ranges, the blade becomes a sky-piercing giant weapon, and the ground beneath him cracks under the pressure of divine force.

The giant-ape Sun Wukong charges out of the crater first, gripping the enormous staff with both hands and sweeping it across the battlefield; wherever the staff passes, rocks shatter and dust waves rise like tides. Deity-form Yang Jian raises the giant blade to block. The staff and blade collide again, this time with an impact like heaven and earth breaking apart; the whole battlefield splits into ring-shaped cracks. The camera pulls into an extreme long shot, showing two enormous figures fighting between sky and earth, every strike rolling up clouds and dust storms. The lens then returns to close range: giant-ape Sun Wukong roars and presses into close combat with fists, staff, and fang-like aggression, while Yang Jian holds his ground in divine form, the divine eye cutting through smoke and dust as the giant blade blocks blow after blow and searches for a counterattack. The image pauses on the instant when their giant weapons press together again, thunderclouds churning above, ground cracks spreading below, and neither side giving way as the fight enters a still grander gods-and-demons confrontation.

During the stalemate, giant-ape Sun Wukong suddenly grins coldly. Golden hairs are lifted from his body by the storm and turn into streaks of light falling around the battlefield. The camera sweeps from behind Yang Jian's shoulder across the shattered plain. Among smoke, rubble, broken banners, and cloud shadows, countless Sun Wukong figures appear at once: some remain giant apes, some return to human form with staff in hand, some hide behind rocks, and some hang upside down from mid-air to attack. All illusions roar together, their voices overlapping into a sky-covering echo, and true-or-false staff blows crash toward Yang Jian's cosmic body from every direction. Yang Jian does not swing wildly. He holds the giant blade across his front, plants his feet on the broken earth, and lets dozens of phantoms graze past blade and armour, scattering sparks from attacks whose truth is difficult to tell.

The image then cuts to a close-up of Yang Jian's celestial eye. The eye appears alone: first as a thin golden line opening at the centre of his brow, then expanding rapidly into a cold-white divine sun. The light does not explode outward; it scans the battlefield layer by layer. Illusions touched by the light first slow down, then their edges crack and dissolve into golden hairs and smoke. Fake bodies hidden in cloud shadows, fake bodies concealed in rubble, and fake bodies mixed into staff afterimages are exposed one after another. The camera tracks across the wide battlefield as hundreds and thousands of Sun Wukong phantoms are erased by the divine light, leaving only the real giant-ape Sun Wukong bursting from Yang Jian's rear flank, the staff scraping along the ground and raising a ridge-like wave of earth.

Yang Jian's celestial-eye light suddenly tightens and locks onto the real Sun Wukong. He does not turn away to dodge. Instead, he lifts the three-pointed double-edged blade behind him and rotates his enormous divine body into a returning slash. Staff and giant blade collide again at his rear flank, and the shock wave shatters the remaining illusions, sending golden hairs down like rain. Giant-ape Sun Wukong is forced back several steps, his feet ploughing deep trenches through the ground, but he soon steadies himself with a low growl. Yang Jian's celestial eye still shines on his brow, dimmer than before but fixed on Sun Wukong's true body, making it difficult for the transformation spell to unfold completely again.

After this exchange of divine techniques, the battle pauses for the first time. The camera pulls to an extreme wide shot: between heaven and earth, only two giant figures stand facing each other across the shattered battlefield. Dust slowly settles around them, and the thunderclouds above have been torn into several layers by the previous impact. Giant-ape Sun Wukong's chest heaves. The shoulder wound from Yang Jian's heavenly strike still glows with golden blood, each drop falling along the fur and punching a small pit in the ground. His fingers tremble slightly around the staff, yet he still raises his head through clenched teeth, fighting will undiminished. Yang Jian's cosmic armour is also cracked, the giant blade has visible notches, and a thin golden trail runs from his celestial eye like a wound left by exhausted divine power. He does not pursue; he only adjusts his breath slowly and raises the blade back to his side.

In the brief silence, broken flags settle in the distance and low thunder rolls inside the clouds. The camera cuts back and forth between the two. Sun Wukong wipes blood from the corner of his mouth; Yang Jian closes his celestial eye for a moment, then opens it again. Both know that the next clash will be heavier than all those before. There is no narration and no declaration of victory or defeat: only two exhausted gods-and-demons figures, unwilling to step back, lowering their centres of gravity in the broken world and preparing to charge each other again.
\end{quote}

\subsubsection*{Cast and setting}

\paragraph{Portraits.} Per-character T2I reference frames produced before generation; segments later reference these via the \texttt{reference\_inputs.portrait} field.

\begin{figure}[H]
  \centering
  \begin{subfigure}[t]{0.24\textwidth}
    \centering
    \includegraphics[width=\linewidth]{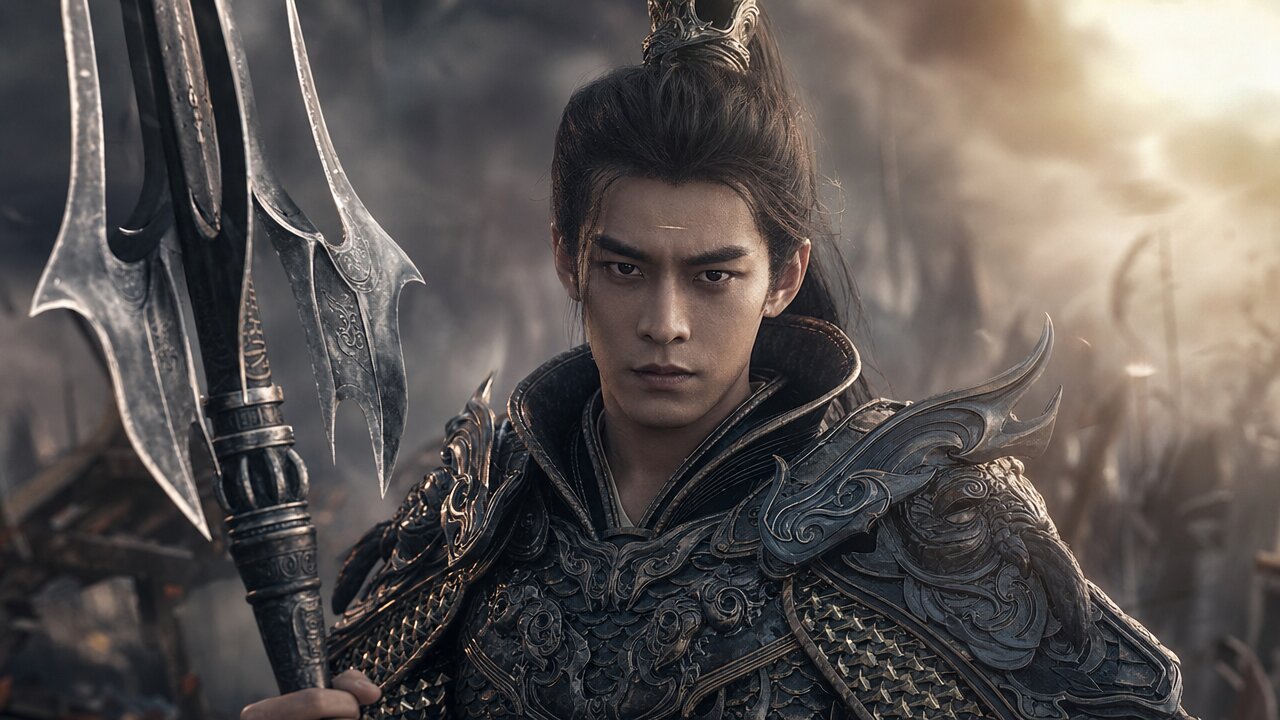}
    \caption*{\texttt{yang\_jian}}
  \end{subfigure}
  \hfill
  \begin{subfigure}[t]{0.24\textwidth}
    \centering
    \includegraphics[width=\linewidth]{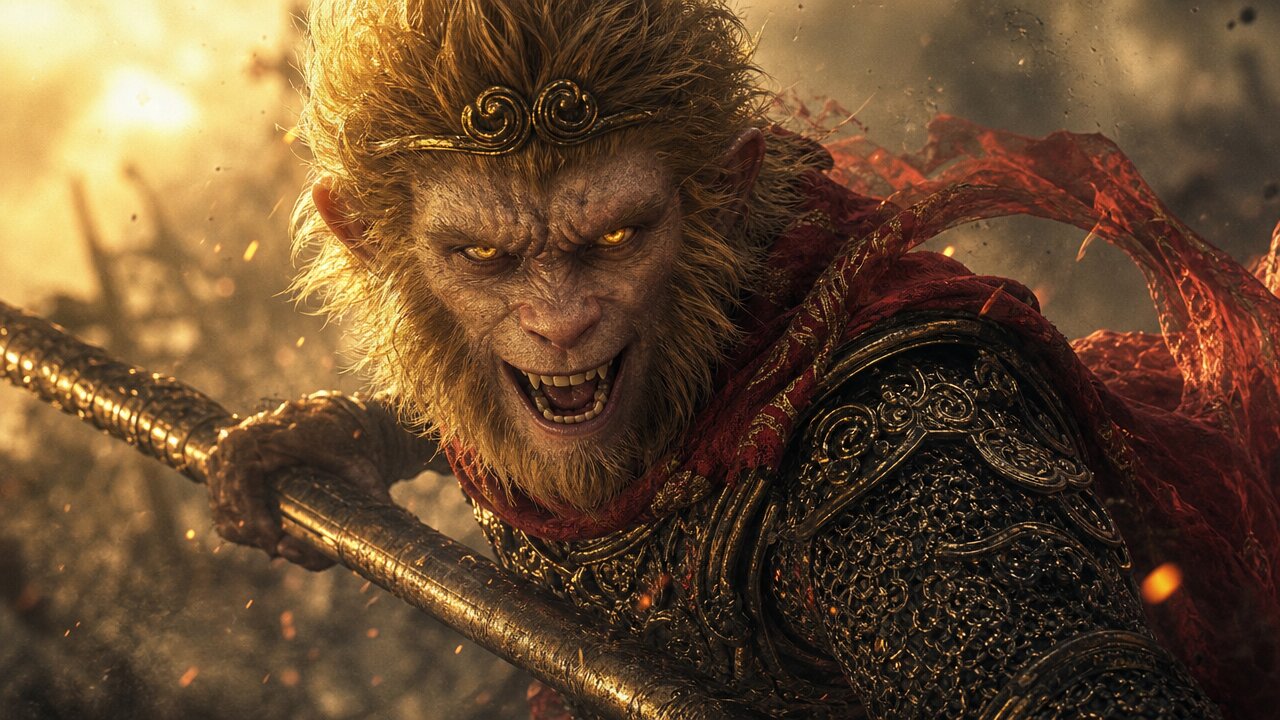}
    \caption*{\texttt{sun\_wukong}}
  \end{subfigure}
  \hfill
  \begin{subfigure}[t]{0.24\textwidth}
    \centering
    \includegraphics[width=\linewidth]{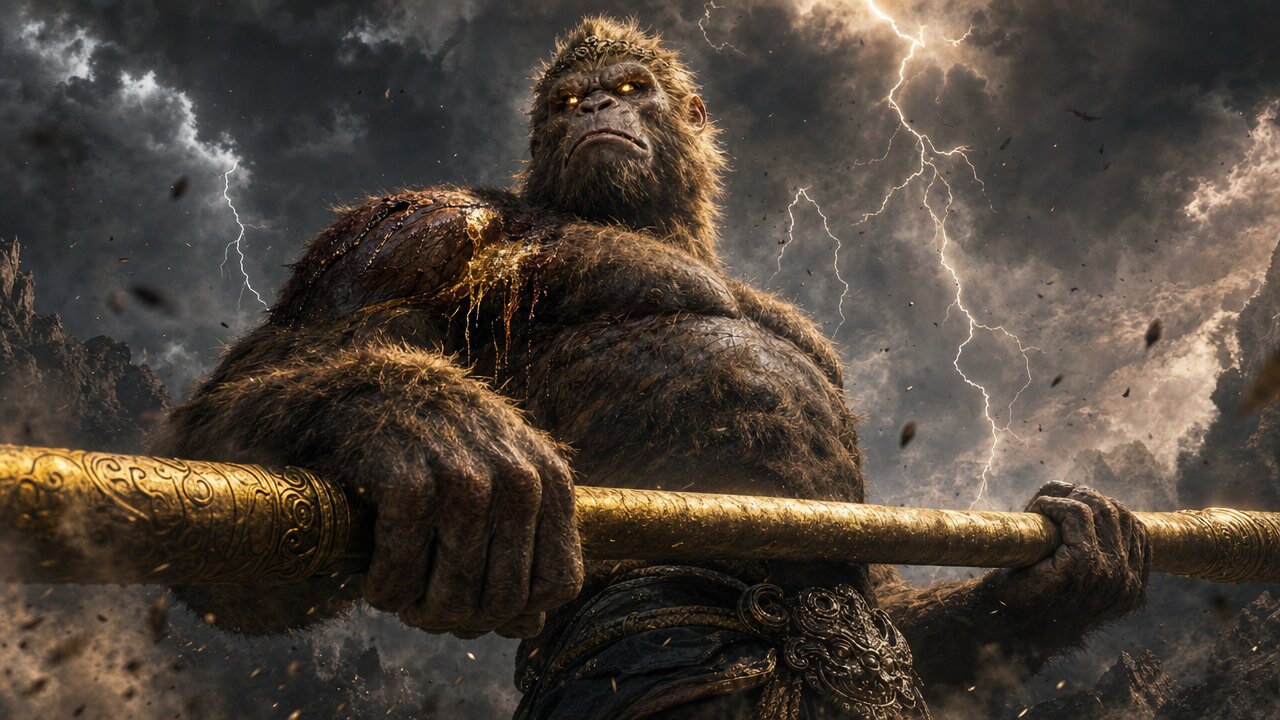}
    \caption*{\texttt{giant\_ape\_wukong}}
  \end{subfigure}
  \hfill
  \begin{subfigure}[t]{0.24\textwidth}
    \centering
    \includegraphics[width=\linewidth]{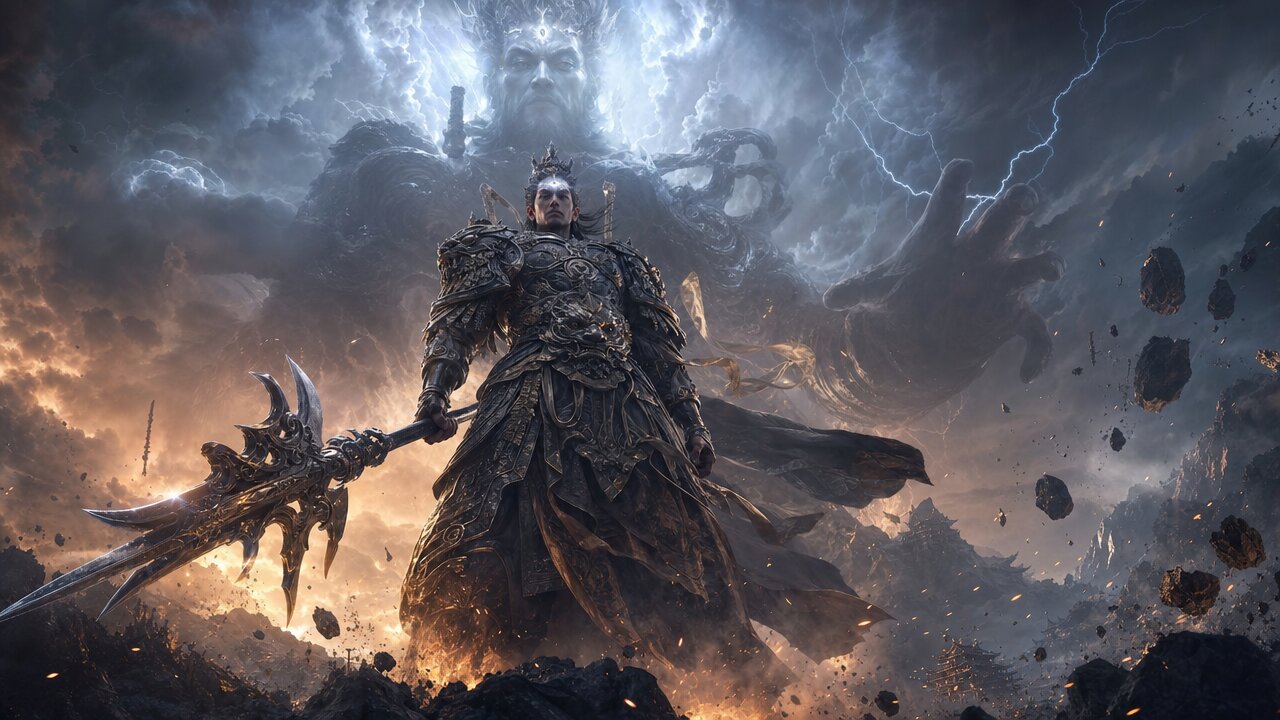}
    \caption*{\texttt{faxiang\_yang\_jian}}
  \end{subfigure}
  \caption{Tianti portrait references.}
  \label{fig:demo-tianti-cast-portraits}
\end{figure}

\begin{itemize}[leftmargin=1.2em,topsep=2pt,itemsep=2pt,parsep=0pt]
  \item \texttt{yang\_jian}: Cinematic realistic character portrait. Yang Jian, an Eastern divine general: stern, sharp features, the third eye between his brows closed into a fine gold line; silver-black armour scaled like fish plates, sharp pauldrons, black hair bound under a crown; the three-pointed double-edged blade in his hand; an upright, restrained pose, gaze firm and lethal. Battlefield dust haze and cold gold backlight surround him. Ultra-high detail, 85\,mm shallow depth-of-field half-portrait.
  \item \texttt{sun\_wukong}: Cinematic realistic character portrait. Sun Wukong, the warrior monkey-king: ape skull structure with a human-form expression, gold-brown fur, eyes burning gold; dark gold chainmail, torn red cape lifted in motion; the ruyi-jingubang in hand; a fierce, broad grin. Dust and fireflies cling to him; the background is the smoke of an ancient battlefield with strong warm-gold side-back light. 85\,mm shallow depth-of-field close-up.
  \item \texttt{giant\_ape\_wukong}: Cinematic realistic giant-creature portrait. Mountain-scale giant-ape Sun Wukong: coarse gold-brown fur standing up like steel needles, fangs exposed, eyes burning with gold ferocity, a sword-cut and gold blood-light on the shoulder, both hands gripping the giant ruyi-jingubang, chest broad and heavy; behind him, dust storm and thunderclouds churn. Low-angle look-up, epic proportions, strong chiaroscuro.
  \item \texttt{faxiang\_yang\_jian}: Cinematic realistic giant-deity portrait. The fatian (cosmic-form) Yang Jian: a sky-tall Eastern general, silver-black mountain-scale plate armour spreading open like ridges; the third eye burns a steady cold-white divine glow; he holds the sky-piercing three-pointed double-edged giant blade. His expression is calm and stern; the earth cracks beneath his feet, with a giant divine silhouette and thunderclouds behind him. 24\,mm low-angle look-up, vast epic feel.
\end{itemize}

\paragraph{Locations.} Per-location reference frames; segments later reference these via \texttt{reference\_inputs.place}.

\begin{figure}[H]
  \centering
  \begin{subfigure}[t]{0.323\textwidth}
    \centering
    \includegraphics[width=\linewidth]{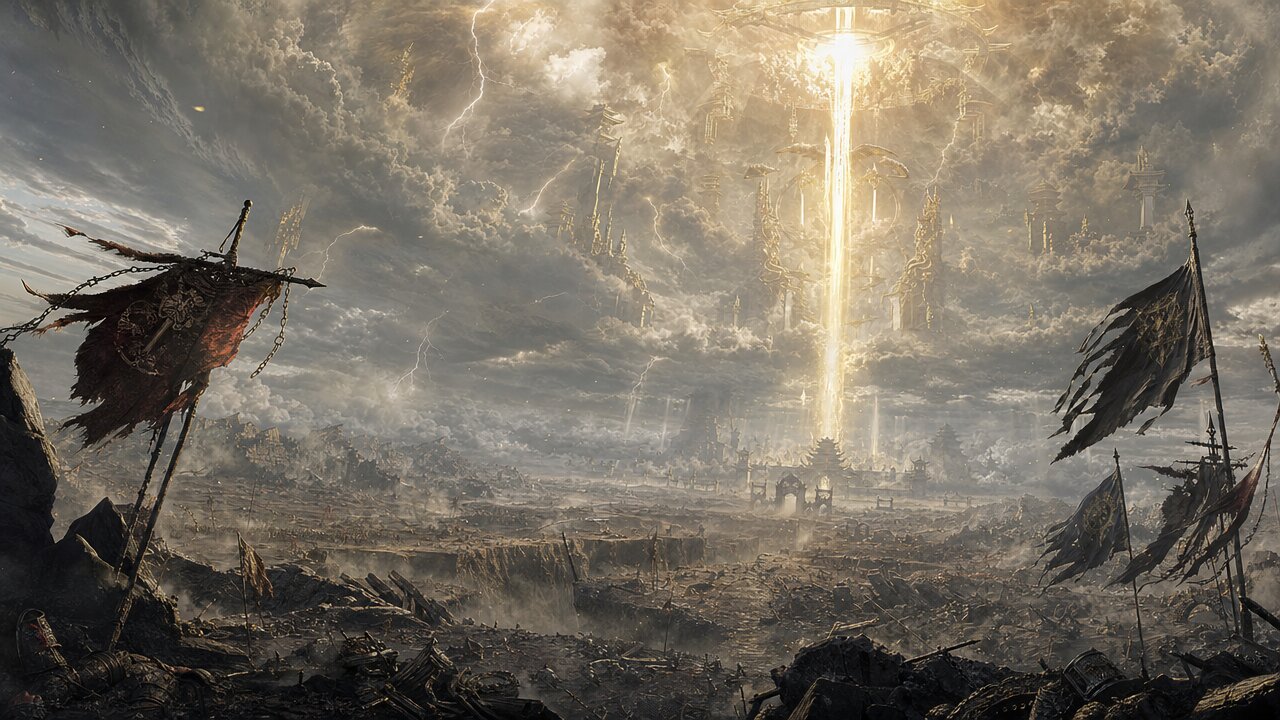}
    \caption*{\texttt{ancient\_battlefield}}
  \end{subfigure}
  \caption{Tianti location reference.}
  \label{fig:demo-tianti-cast-locations}
\end{figure}

\begin{itemize}[leftmargin=1.2em,topsep=2pt,itemsep=2pt,parsep=0pt]
  \item \texttt{ancient\_battlefield}: Whole-location reference: an ancient gods-and-demons battlefield connected with the high heavens. The ground is a broken plain: deep pits, torn banners, rubble, dust storms; high above, churning seas of cloud, lightning and giant light pillars. Colours dominated by cold grey, dust-brown, and gold-white divine light. Cinematic realistic epic-war atmosphere.
\end{itemize}

\paragraph{Props.} Per-prop reference frames; segments later reference these via \texttt{reference\_inputs.prop}.

\begin{figure}[H]
  \centering
  \begin{subfigure}[t]{0.24\textwidth}
    \centering
    \includegraphics[width=\linewidth]{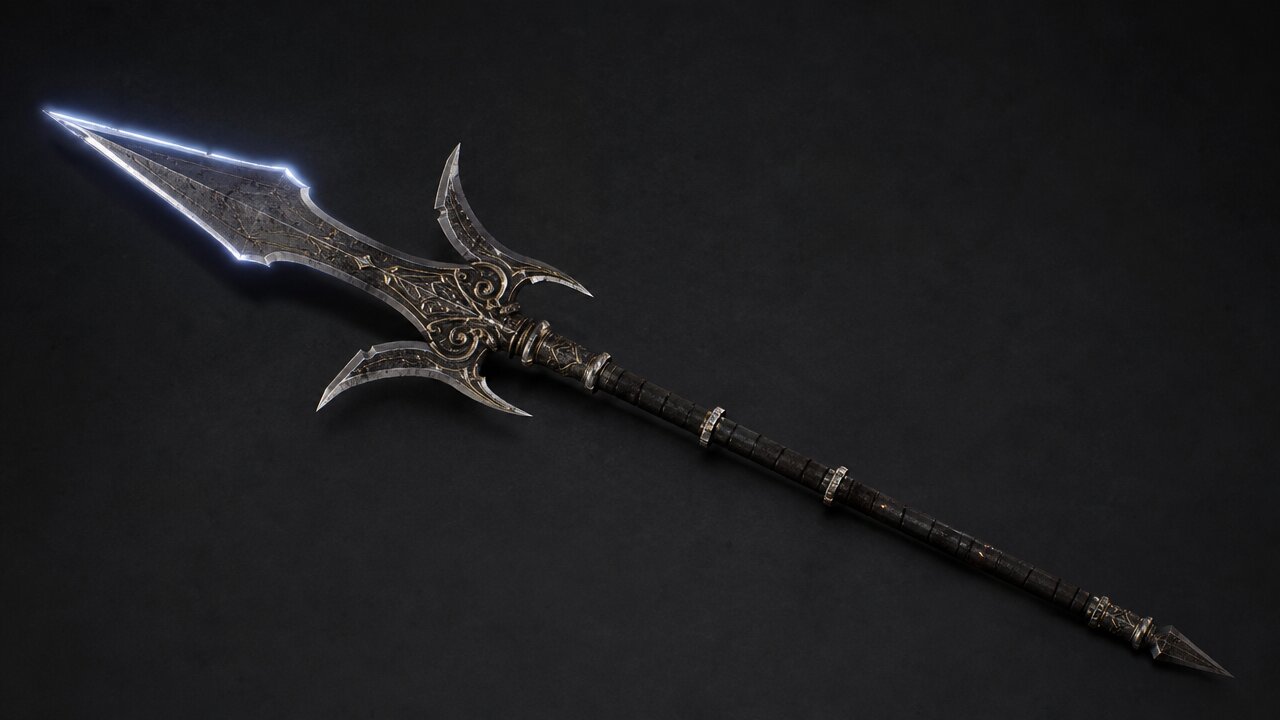}
    \caption*{\texttt{sanjian\_liangren\_blade}}
  \end{subfigure}
  \hfill
  \begin{subfigure}[t]{0.24\textwidth}
    \centering
    \includegraphics[width=\linewidth]{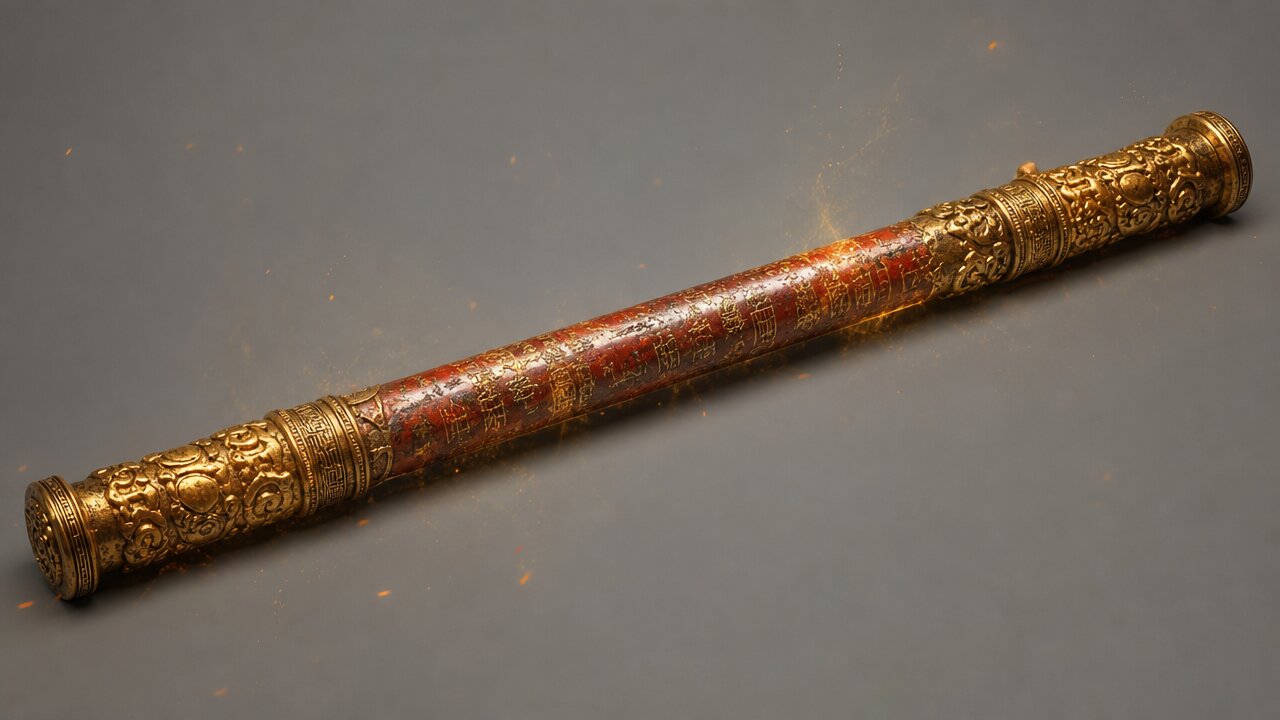}
    \caption*{\texttt{ruyi\_jingubang}}
  \end{subfigure}
  \hfill
  \begin{subfigure}[t]{0.24\textwidth}
    \centering
    \includegraphics[width=\linewidth]{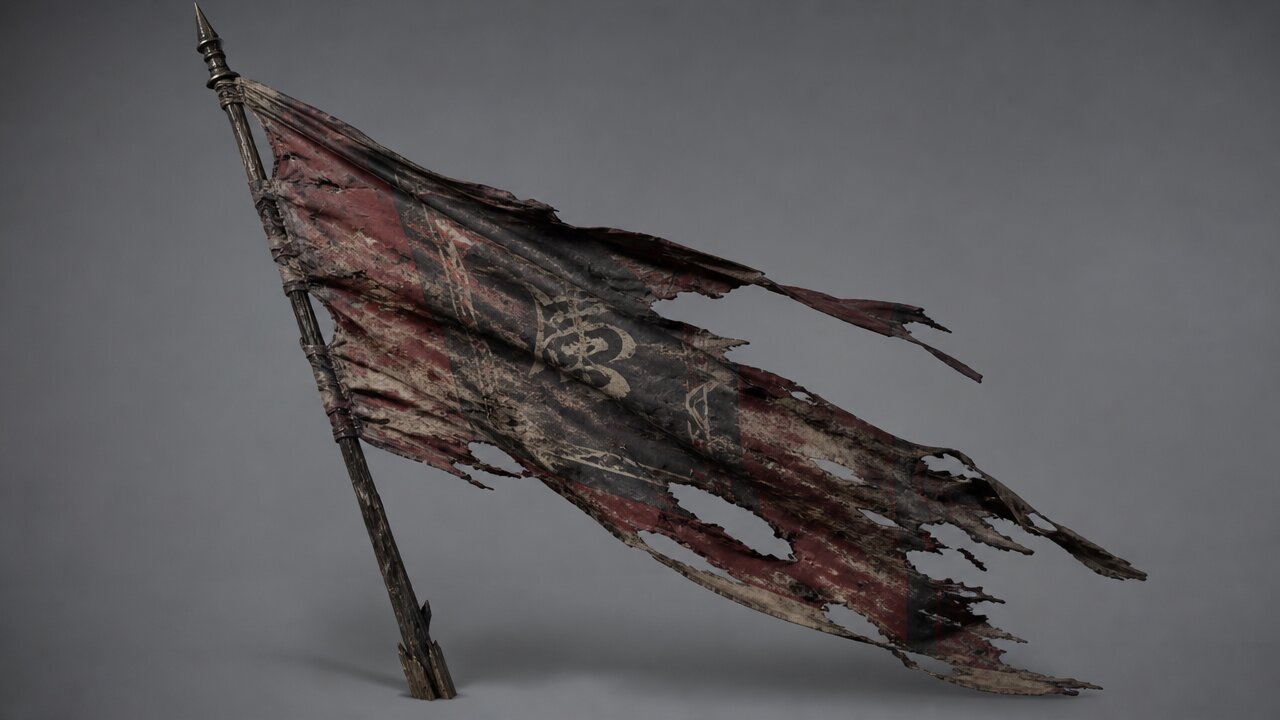}
    \caption*{\texttt{broken\_battle\_flags}}
  \end{subfigure}
  \hfill
  \begin{subfigure}[t]{0.24\textwidth}
    \centering
    \includegraphics[width=\linewidth]{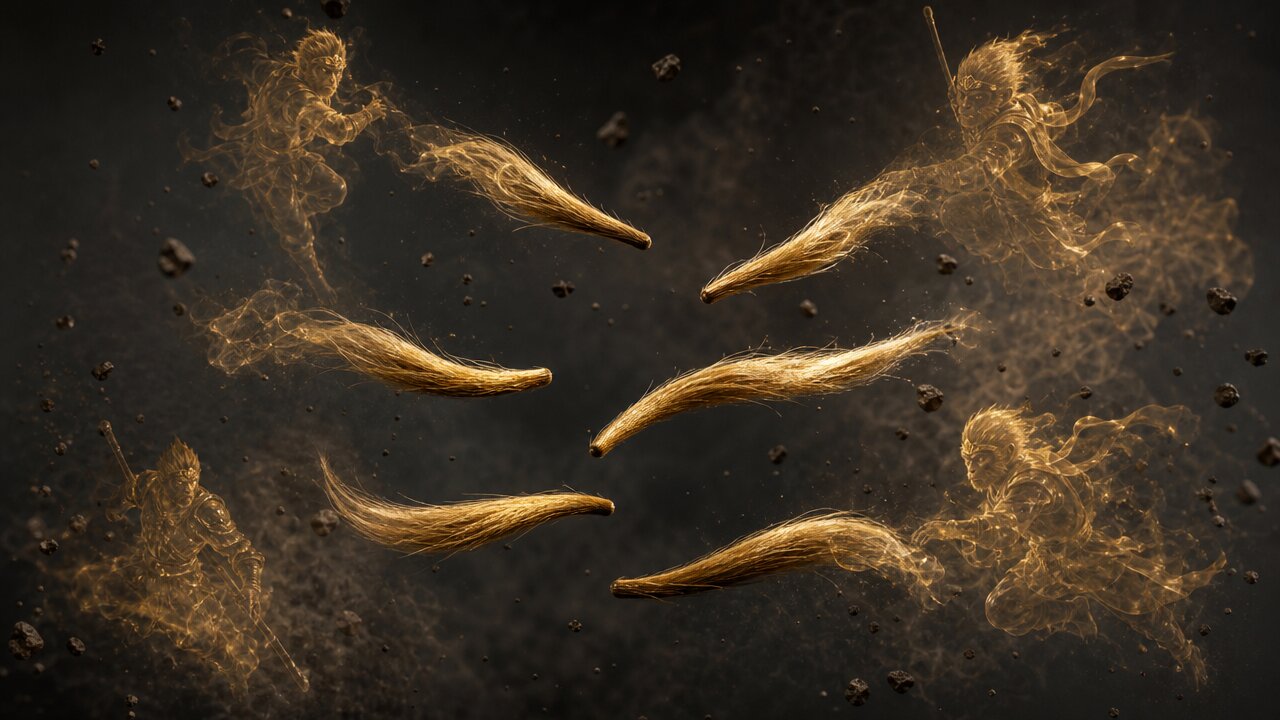}
    \caption*{\texttt{golden\_hair\_clones}}
  \end{subfigure}
  \caption{Tianti prop references.}
  \label{fig:demo-tianti-cast-props}
\end{figure}

\begin{itemize}[leftmargin=1.2em,topsep=2pt,itemsep=2pt,parsep=0pt]
  \item \texttt{sanjian\_liangren\_blade}: Static prop reference: Yang Jian's three-pointed double-edged blade: a long-shafted silver-black metal weapon with a wide, sharp main edge and two smaller side edges symmetrically extended; a cold-white divine glow along the edges with fine battle nicks; the haft wrapped in black leather with metal hoops, dusted with fireflies and earth. Cinematic realistic prop feel.
  \item \texttt{ruyi\_jingubang}: Static prop reference: Sun Wukong's ruyi-jingubang: a heavy gold-red metal staff, with clear gold-band ornaments at both ends, ancient sigils and battle scratches along the body, sparks, dust and a gold aura on the surface; heavy in proportion; cinematic realistic prop sheen.
  \item \texttt{broken\_battle\_flags}: Static prop reference: torn battle banners on an ancient battlefield: dark red and grey-black cloth ripped, poles snapped or tilted, edges burnt, covered with earth, suitable for being uprooted by a shock wave. Realistic war-prop feel.
  \item \texttt{golden\_hair\_clones}: Static prop reference: Sun Wukong's supernatural hairs: coarse gold strands floating in the air, each emitting a faint gold glow, some turning into the edges of translucent monkey-king phantoms; smoke and rubble in the background; realistic fantasy-spell feel.
\end{itemize}

\subsubsection*{Shots}

\paragraph{Shot~1: \texttt{shot01\_back\_clash} (10\,s).} \emph{Story goal:} Establish the power level of the first head-on collision between Yang Jian and Sun Wukong, and the combative tone in which neither side is willing to yield.

\begin{figure}[H]
  \centering
  \includegraphics[width=\linewidth]{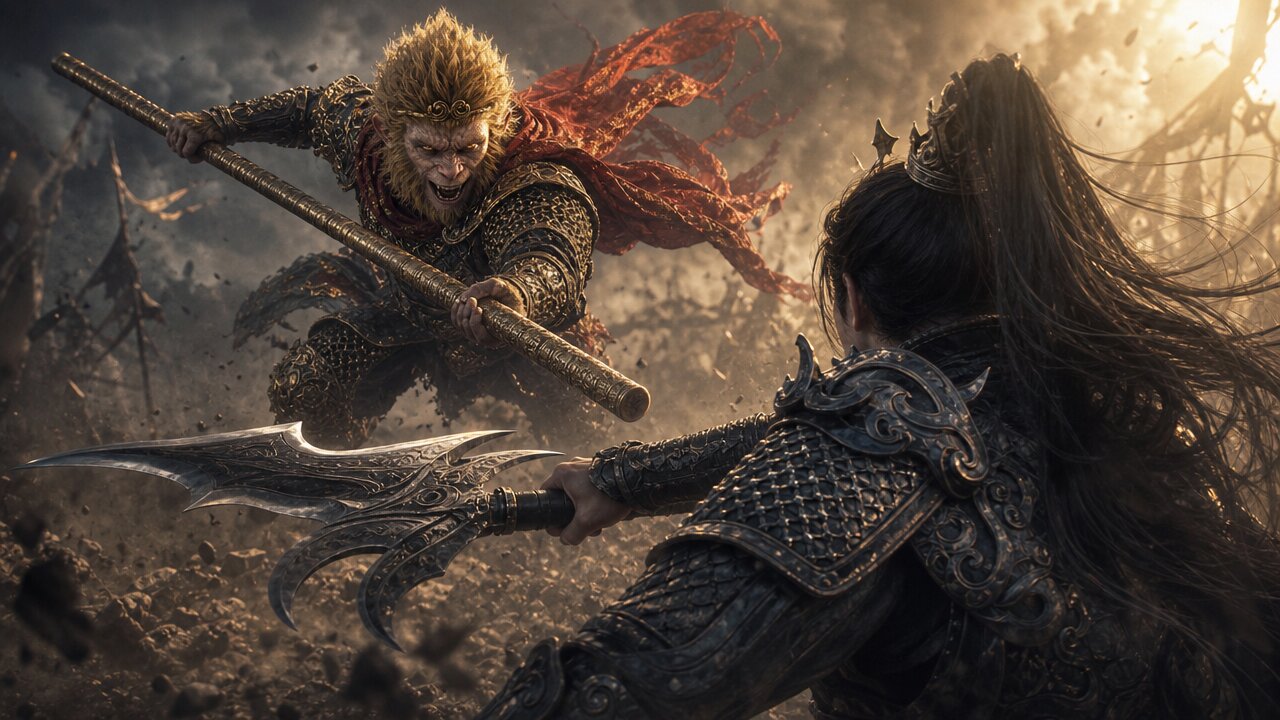}
  \caption{Starting anchor frame \texttt{a01\_back\_clash\_start} for Tianti shot~1.}
  \label{fig:demo-tianti-a01_back_clash_start}
\end{figure}

\noindent\emph{Segments.}
\begin{itemize}[leftmargin=1.2em,topsep=2pt,itemsep=2pt,parsep=0pt]
  \item \texttt{seg\_shot01\_back\_clash\_00} (10\,s): From a low angle behind Yang Jian's silver-black armour, dolly forward toward him. He grips his three-pointed double-edged blade with both hands, lowering his stance as cold dust rolls along the plates of his armour. Sun Wukong leaps in head-on, his dark gold chainmail and torn red cape lifted by the gale, the ruyi-jingubang bearing down on the blade with heavy gold light. The two divine weapons collide hard, dense orange-gold sparks burst at the contact, fireflies explode along the edges of staff and blade, and the airflow shocks the surrounding dust into a ring-shaped wave. The shot pushes between the two: Yang Jian's back leans forward without yielding, Sun Wukong gritting his teeth as he presses down, the two locked at the centre of the battlefield.
\end{itemize}

\paragraph{Shot~2: \texttt{shot02\_weapon\_lock\_close} (5\,s).} \emph{Story goal:} Magnify the pressure at the weapon contact point so the audience feels the vibration and shock of two divine weapons meeting head-on.

\begin{figure}[H]
  \centering
  \includegraphics[width=\linewidth]{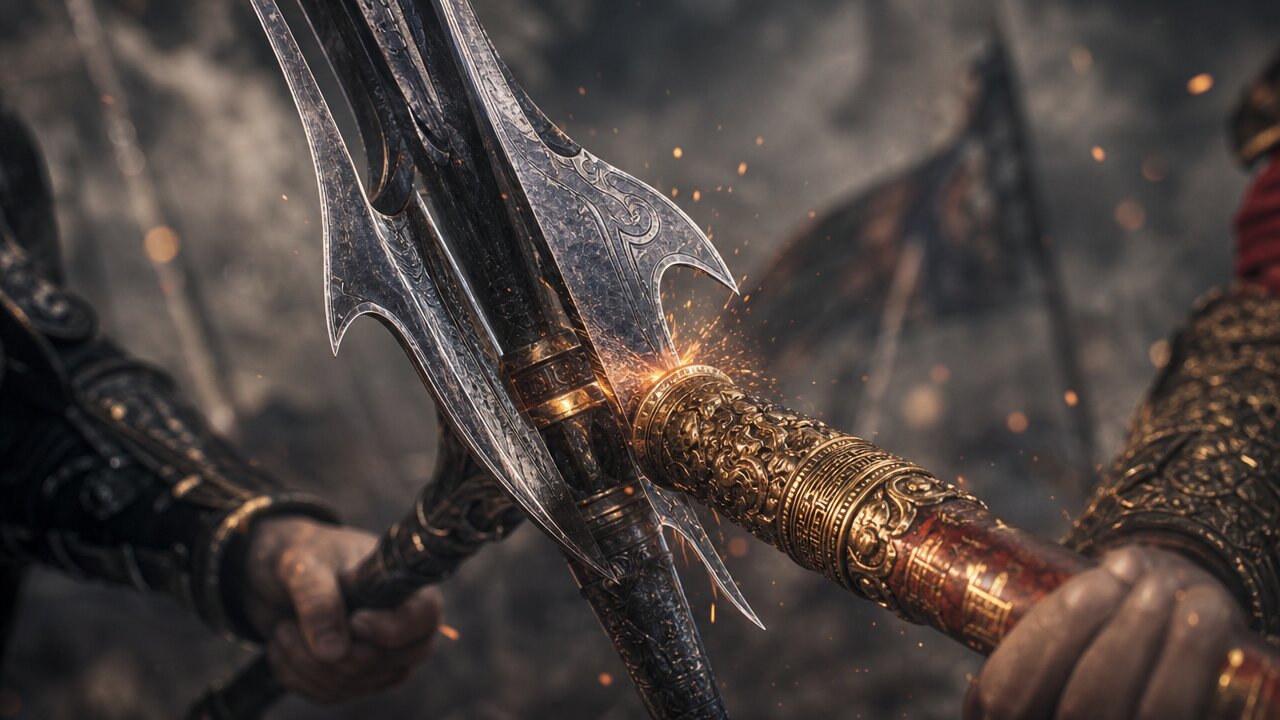}
  \caption{Starting anchor frame \texttt{a02\_weapon\_lock\_close\_start} for Tianti shot~2.}
  \label{fig:demo-tianti-a02_weapon_lock_close_start}
\end{figure}

\noindent\emph{Segments.}
\begin{itemize}[leftmargin=1.2em,topsep=2pt,itemsep=2pt,parsep=0pt]
  \item \texttt{seg\_shot02\_weapon\_lock\_close\_00} (5\,s): From an 85mm macro flat-on view of the weapon contact, continue dollying in. The edge of the three-pointed double-edged blade and the ruyi-jingubang press together harder in the centre of the frame, with both arms still applying force in the shallow depth-of-field foreground. The orange-gold sparks at the contact escalate from a few stray flares to a dense spray; cold-white blade light flickers along the metal scratches, and fine micro-tremors and cracking light appear on staff and blade. The airflow is squeezed into a ring-shaped shock wave that pushes grey-brown dust away. End-frame: blade and staff biting deep into one another, sparks and air-rings expanding outward in a high-contrast instant.
\end{itemize}

\paragraph{Shot~3: \texttt{shot03\_faces\_roar} (6\,s).} \emph{Story goal:} Reveal the two characters' fighting will, anger, and defiance, escalating the conflict from physical contest to a contest of will.

\begin{figure}[H]
  \centering
  \includegraphics[width=\linewidth]{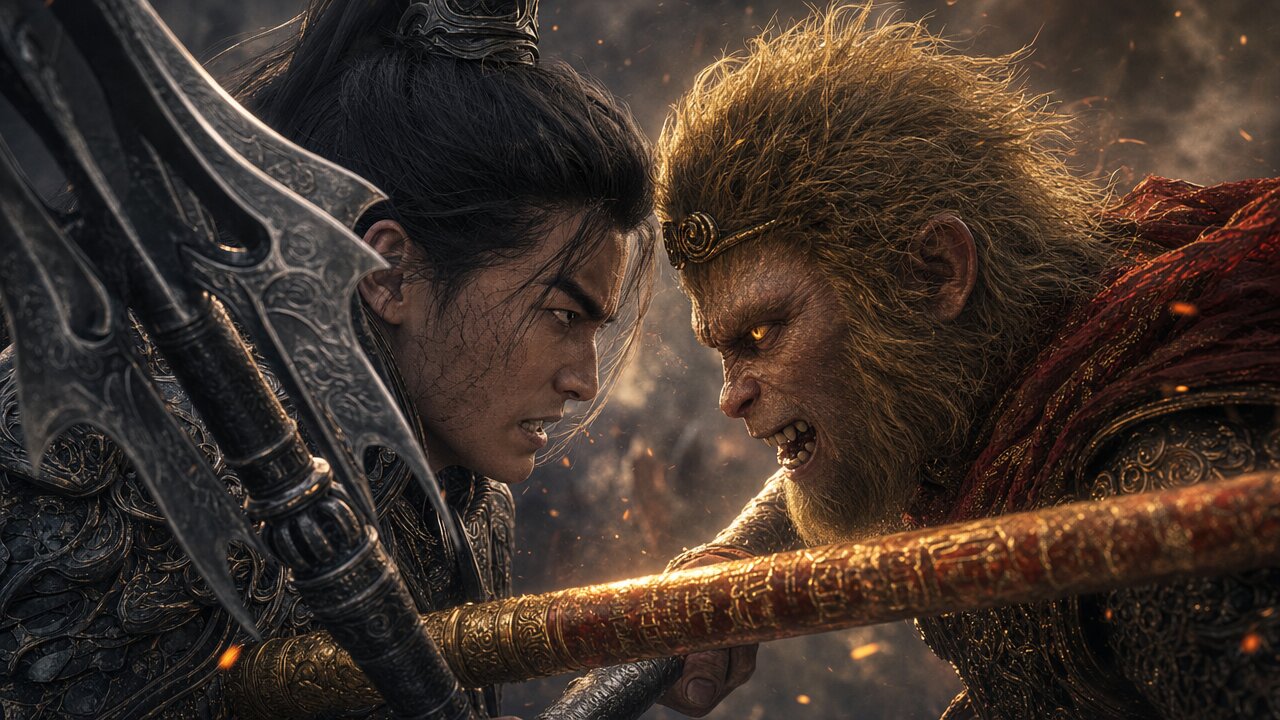}
  \caption{Starting anchor frame \texttt{a03\_faces\_roar\_start} for Tianti shot~3.}
  \label{fig:demo-tianti-a03_faces_roar_start}
\end{figure}

\noindent\emph{Segments.}
\begin{itemize}[leftmargin=1.2em,topsep=2pt,itemsep=2pt,parsep=0pt]
  \item \texttt{seg\_shot03\_faces\_roar\_00} (6\,s): From a tight foreground framed by the weapon shafts and crossing shadows, continue dollying in until Yang Jian's and Sun Wukong's faces are nose to nose, foreheads veined, teeth clenched, eyes glaring. Handheld micro-shake; breath and the locked force are pressed between the two faces while orange-gold sparks streak past their cheeks and a cold grey sky darkens the armour edges. The lens crowds in to a face close-up; their fury fully ignites, both mouths open in a roar, the fine line of Yang Jian's third eye glows faintly, and Sun Wukong's golden irises burn even brighter.
\end{itemize}

\paragraph{Shot~4: \texttt{shot04\_separated\_standoff} (6\,s).} \emph{Story goal:} End the first round of close stalemate, restore distance between the two combatants, and set up Sun Wukong's first ascent for the next wave of offence.

\begin{figure}[H]
  \centering
  \includegraphics[width=\linewidth]{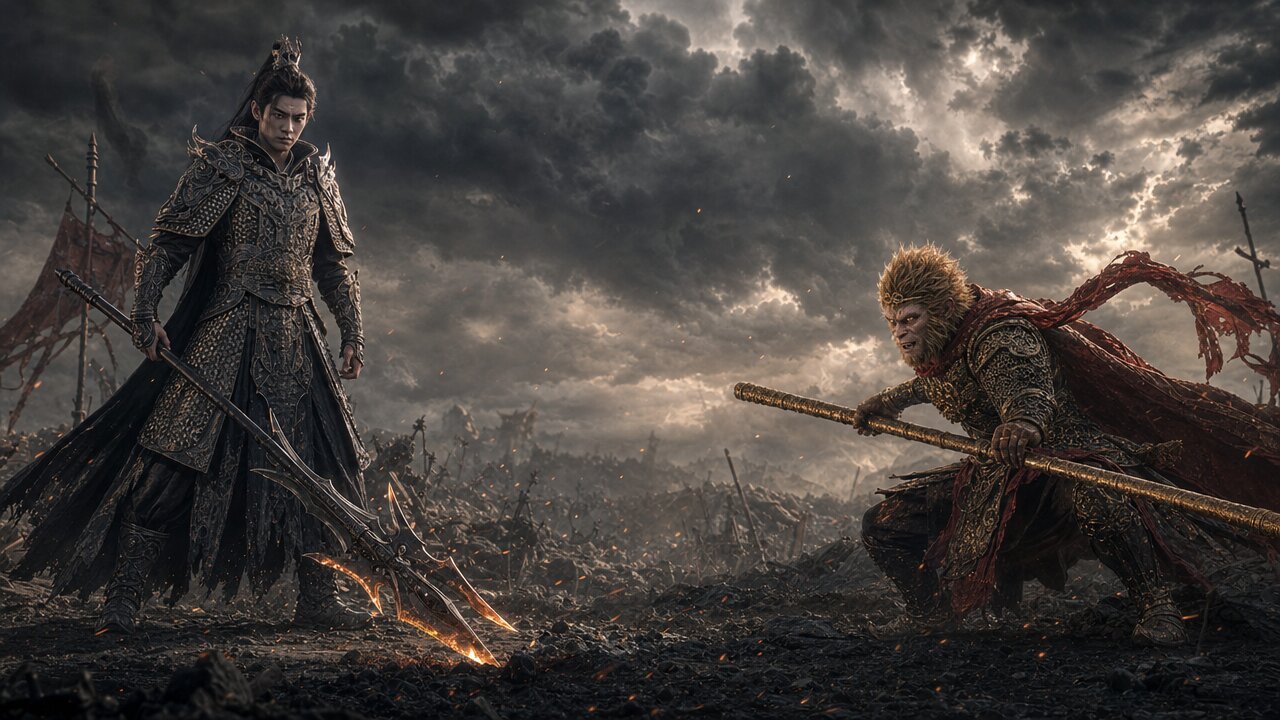}
  \caption{Starting anchor frame \texttt{a04\_separated\_standoff\_start} for Tianti shot~4.}
  \label{fig:demo-tianti-a04_separated_standoff_start}
\end{figure}

\noindent\emph{Segments.}
\begin{itemize}[leftmargin=1.2em,topsep=2pt,itemsep=2pt,parsep=0pt]
  \item \texttt{seg\_shot04\_separated\_standoff\_00} (6\,s): From a wide shot, catch the moment they break apart. Yang Jian on the left steadies his footing, his silver-black armour dusted, his three-pointed double-edged blade rising slowly from a tilted ground-pointing position; Sun Wukong on the right squats low to drop his centre of gravity, the ruyi-jingubang held across his body, the torn red cape lifted by the residual wind. The lens dollies out slightly; the dust at the battlefield centre is pushed aside by both auras into a clear grey-brown rift, with scattered sparks faintly glowing in the cold grey haze. End-frame: the two flanking the frame, weapons pointed at each other in a frozen standoff.
\end{itemize}

\paragraph{Shot~5: \texttt{shot05\_ascent\_through\_clouds} (28\,s).} \emph{Story goal:} Push the fight from the mortal battlefield up into the high heavens, expanding the scale of the duel and establishing the epic feel of vertical space.

\begin{figure}[H]
  \centering
  \includegraphics[width=\linewidth]{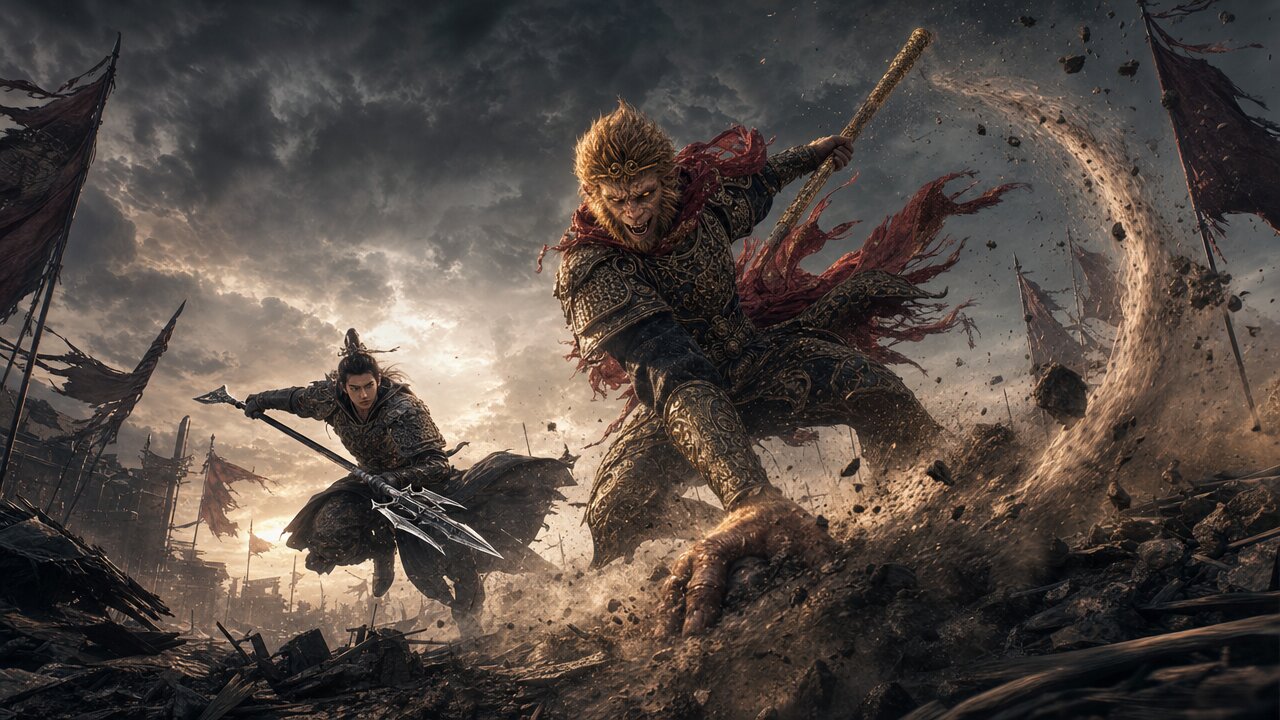}
  \caption{Starting anchor frame \texttt{a05\_ascent\_through\_clouds\_start} for Tianti shot~5.}
  \label{fig:demo-tianti-a05_ascent_through_clouds_start}
\end{figure}

\noindent\emph{Segments.}
\begin{itemize}[leftmargin=1.2em,topsep=2pt,itemsep=2pt,parsep=0pt]
  \item \texttt{seg\_shot05\_ascent\_through\_clouds\_00} (10\,s): From a low-angle look-up, catch the instant Sun Wukong stomps and flicks his staff upward. Cracks burst outward beneath his feet; the dust whipped up by the ruyi-jingubang quickly grows into a heavy column. His golden eyes light up, the torn red cape stretches taut in the rising air, and he rockets skyward riding the staff's momentum; Yang Jian's silver-black armour follows close behind, the three-pointed double-edged blade carving a cold-white arc along the dust wave. The camera lifts off the ground with the column, the broken banners and rubble of the battlefield swept into the lower frame. End-frame: the two leaving the ground, with the dust column running through the centre of the picture in a high-altitude take-off pose.
  \item \texttt{seg\_shot05\_ascent\_through\_clouds\_01} (9\,s): Continuing the rising posture inside the dust column, the camera keeps streaking up with the two combatants. The cracked ground, broken banners and rubble shrink rapidly below. Sun Wukong is still half a body length ahead, the ruyi-jingubang pulling the dust wave upward and the gold light along the staff turning the column edge into a burning ring; Yang Jian holds the pursuit line, the three-pointed double-edged blade splitting the oncoming air, and the white wake behind the blade is clearly drawn out. Heavy low cloud presses against the top of the frame; the dust-brown battlefield tone is swallowed by cold grey cloud-shadow. End-frame: the two reaching the underside of the cloud cover, the cloud floor pushed into a concave dent by the rising airflow.
  \item \texttt{seg\_shot05\_ascent\_through\_clouds\_02} (9\,s): From the underside of the low cloud, the climb continues. Sun Wukong's ruyi-jingubang pierces the dense cloud first; the cloud cover rolls open as if drilled by a giant blade, and white-grey vapour rushes past either side of the lens. Yang Jian follows close, the third eye on his forehead still a fine closed line; the cold white edge of the three-pointed double-edged blade carves a clear corridor through the mist. Both figures are momentarily swallowed by the cloud sea, then burst out again in fragments of cloud and gold-white light pillars. The colour shifts entirely from ground dust-brown to a high-altitude gold-white and cold blue. End-frame: the two stand one ahead of the other above the cloud sea, with the mortal battlefield shrunk to a blurred grey patch below.
\end{itemize}

\paragraph{Shot~6: \texttt{shot06\_cloud\_counter} (14\,s).} \emph{Story goal:} Stage the first attack-defence reversal in the clouds, showing that Sun Wukong's footwork is agile and Yang Jian's reaction is precise.

\begin{figure}[H]
  \centering
  \includegraphics[width=\linewidth]{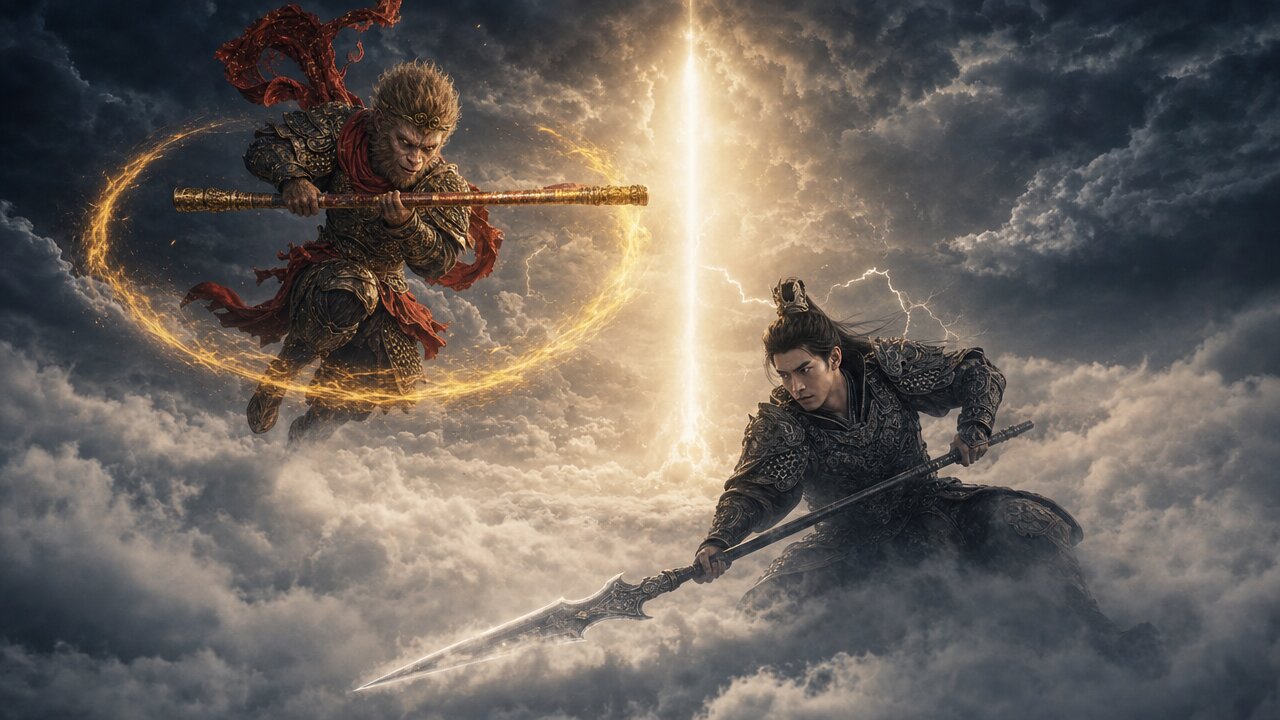}
  \caption{Starting anchor frame \texttt{a06\_cloud\_counter\_start} for Tianti shot~6.}
  \label{fig:demo-tianti-a06_cloud_counter_start}
\end{figure}

\noindent\emph{Segments.}
\begin{itemize}[leftmargin=1.2em,topsep=2pt,itemsep=2pt,parsep=0pt]
  \item \texttt{seg\_shot06\_cloud\_counter\_00} (7\,s): Cut in above the white-grey vapour of the cloud surface. Sun Wukong twists half his body around, golden eyes laughing, and sweeps the ruyi-jingubang held across his body. The camera arcs around him as he turns; a half-circle of golden energy expands into a bright arc, and the heavy clouds are flicked into rolling layers by the staff-wind. Yang Jian's silver-black armour ducks low to one side, the three-pointed double-edged blade skimming the cloud surface and dodging the leading edge of the energy wave. Gold-white light pillars and dark-blue thunderclouds intersect. End-frame: the golden arc sweeps past Yang Jian's flank as he is already coiled to thrust back.
  \item \texttt{seg\_shot06\_cloud\_counter\_01} (7\,s): From the frame in which Yang Jian has crouched and gathered force, the lens pulls off Sun Wukong's golden energy wave and dollies fast along the cloud surface toward the blade tip. Yang Jian raises his cold gaze, his silver-black armour shaking in the wind, and the three-pointed double-edged blade strikes through the gap left by the dodge, and the cold-white blade light cuts through the cloud layer like a long thin lightning bolt; behind him, the residual ring of Sun Wukong's golden sweep diffuses. The cloud surface is opened by the blade into a long deep gash. End-frame: the blade tip pierces the cloud sea and the rift extends into the distance.
\end{itemize}

\paragraph{Shot~7: \texttt{shot07\_high\_sky\_battle} (24\,s).} \emph{Story goal:} Show the high-altitude battle entering a sustained tangle, foregrounding Sun Wukong's unpredictable transformations and Yang Jian's celestial-eye prediction logic.

\begin{figure}[H]
  \centering
  \includegraphics[width=\linewidth]{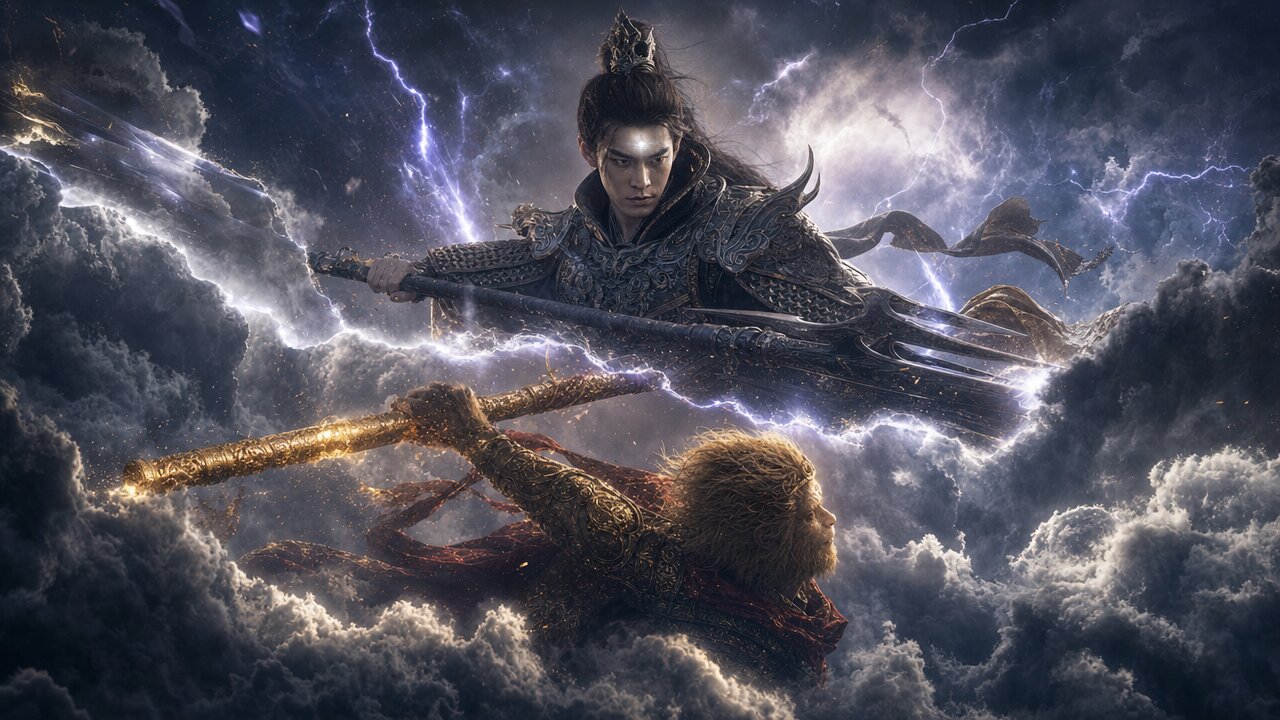}
  \caption{Starting anchor frame \texttt{a07\_high\_sky\_battle\_start} for Tianti shot~7.}
  \label{fig:demo-tianti-a07_high_sky_battle_start}
\end{figure}

\noindent\emph{Segments.}
\begin{itemize}[leftmargin=1.2em,topsep=2pt,itemsep=2pt,parsep=0pt]
  \item \texttt{seg\_shot07\_high\_sky\_battle\_00} (8\,s): From an oblique downward angle of Yang Jian holding the centre line, continue dollying in. Sun Wukong hangs upside-down beneath the cloud and reverses his staff: half his silhouette is swallowed by the thunderhead and then bursts out again as he tears through the cloud. The lens snaps closer to the strike point of staff and blade, where the gold and silver-white edge light burst into dense small sparks while blue-violet lightning flickers behind the mist. Yang Jian's third eye gleams faintly with cold light for the first time, suppressing the staff path in advance. End-frame: Yang Jian blocks the ruyi-jingubang horizontally with his blade, with Sun Wukong inverted half a step above him.
  \item \texttt{seg\_shot07\_high\_sky\_battle\_01} (8\,s): Continuing from the mid-air freeze of locked weapons, Sun Wukong rebounds on the recoil and disappears into the side cloud bank. His outline blurs in and out of the vapour, and the after-image of the ruyi-jingubang slashes diagonally from behind Yang Jian's flank. The lens whips through the cloud and lands beside Yang Jian's face; the cold light of the third eye is even cleaner, like a fine line cutting through the smoke, and he turns just in time to seal the centre line before the ambush lands. End-frame: Yang Jian steps sideways and raises the blade to block a blow from behind, with Sun Wukong emerging from the edge of the thunderhead, the staff tip stopped just outside the blade.
  \item \texttt{seg\_shot07\_high\_sky\_battle\_02} (8\,s): From the over-shoulder block, continue. Sun Wukong shifts his angle continuously, skimming the cloud spine one moment and being swallowed by mist the next, leaving behind only the gold arcs of the ruyi-jingubang and a monkey-king silhouette in the lightning. The lens snap-cuts between long shots and weapon close-ups; each lift of the three-pointed double-edged blade tears out a silver-white air-arc, while Yang Jian's cold third-eye light tracks the real among the feints and presses closer block by block. End-frame: Yang Jian closes in still further with the blade, while several after-images of Sun Wukong's strike stop half a step away.
\end{itemize}

\paragraph{Shot~8: \texttt{shot08\_mountain\_staff\_press} (12\,s).} \emph{Story goal:} Let Sun Wukong manifest the giant-form ruyi-jingubang as an overwhelming threat and push the fight to a higher-intensity flashpoint.

\begin{figure}[H]
  \centering
  \includegraphics[width=\linewidth]{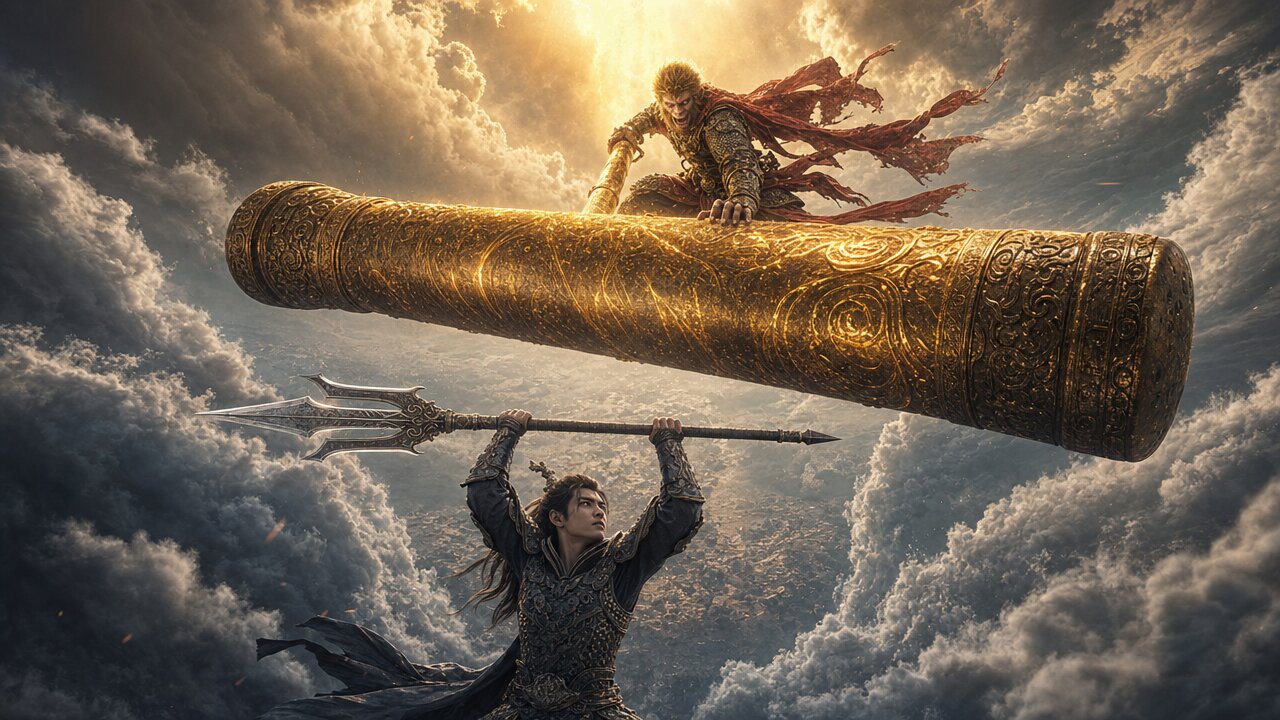}
  \caption{Starting anchor frame \texttt{a08\_mountain\_staff\_press\_start} for Tianti shot~8.}
  \label{fig:demo-tianti-a08_mountain_staff_press_start}
\end{figure}

\noindent\emph{Segments.}
\begin{itemize}[leftmargin=1.2em,topsep=2pt,itemsep=2pt,parsep=0pt]
  \item \texttt{seg\_shot08\_mountain\_staff\_press\_00} (12\,s): From a low-angle look-up at the edge of a rift in the cloud sea, Yang Jian holds the blade across his body and refuses to fall back, the silver-black plates glinting cold under the giant pressure. Above him at the zenith, Sun Wukong laughs wildly and lowers a mountain-sized ruyi-jingubang slowly down; the base of the staff blocks out half the sky, with gold-red metal patterns, glyph etchings and a heavy shadow pressing toward the blade. At the moment of contact this is not a glancing touch but a mountain-pillar weight bearing down on the three-pointed double-edged blade; orange-gold sparks spray densely along the edge, and the cloud sea is torn open on both sides to reveal the distant ground far below. The lens then follows Yang Jian as he is driven down, the cloud cover ringing outward beneath his feet. End-frame: he has slid a long way down yet is still bracing with the blade across his body.
\end{itemize}

\paragraph{Shot~9: \texttt{shot09\_counter\_knockdown} (9\,s).} \emph{Story goal:} Complete Yang Jian's pivot from absorbing to counter-attacking, with the first clear wounding and suppression of Sun Wukong.

\begin{figure}[H]
  \centering
  \includegraphics[width=\linewidth]{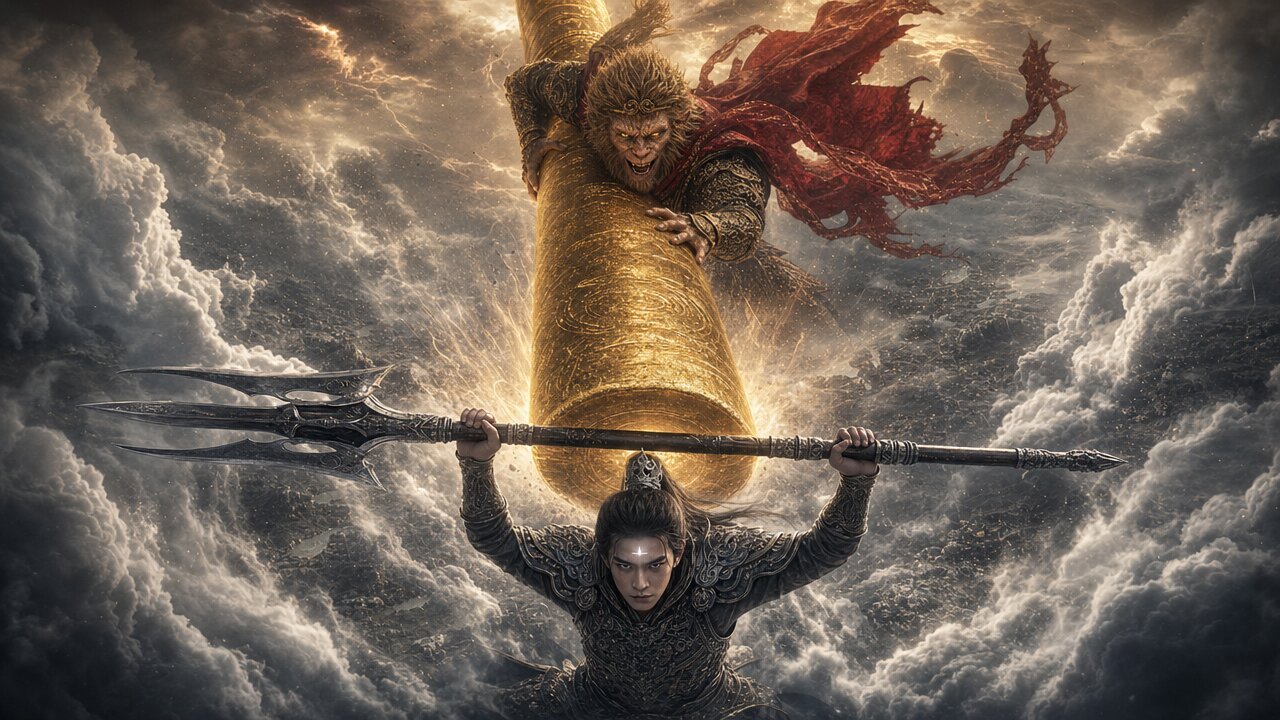}
  \caption{Starting anchor frame \texttt{a09\_counter\_knockdown\_start} for Tianti shot~9.}
  \label{fig:demo-tianti-a09_counter_knockdown_start}
\end{figure}

\noindent\emph{Segments.}
\begin{itemize}[leftmargin=1.2em,topsep=2pt,itemsep=2pt,parsep=0pt]
  \item \texttt{seg\_shot09\_counter\_knockdown\_00} (9\,s): Begin from the side low angle of Yang Jian pinned by the giant ruyi-jingubang, with the cloud beneath his feet scattered into a circular hole. The cold-white aura at his forehead intensifies fast. The lens whips diagonally upward to follow the back-cut of the three-pointed double-edged blade: silver-white blade light tears through the mist along the pressure of the ruyi-jingubang and pushes Sun Wukong back in a succession of strikes; the cold-white aura briefly drowns out the warm gold staff shadow, and sparks and cloud fragments are flung aside by the impact. The final blade stroke lands square on Sun Wukong's chest; he tilts back and falls toward the rift in the cloud, while Yang Jian's pursuing pose is held high above with the blade.
\end{itemize}

\paragraph{Shot~10: \texttt{shot10\_wukong\_fall\_impact} (14\,s).} \emph{Story goal:} Convert the mid-air defeat into a massive ground-level disaster, accumulating Sun Wukong's rage and destructive energy ahead of the giant-ape transformation.

\begin{figure}[H]
  \centering
  \includegraphics[width=\linewidth]{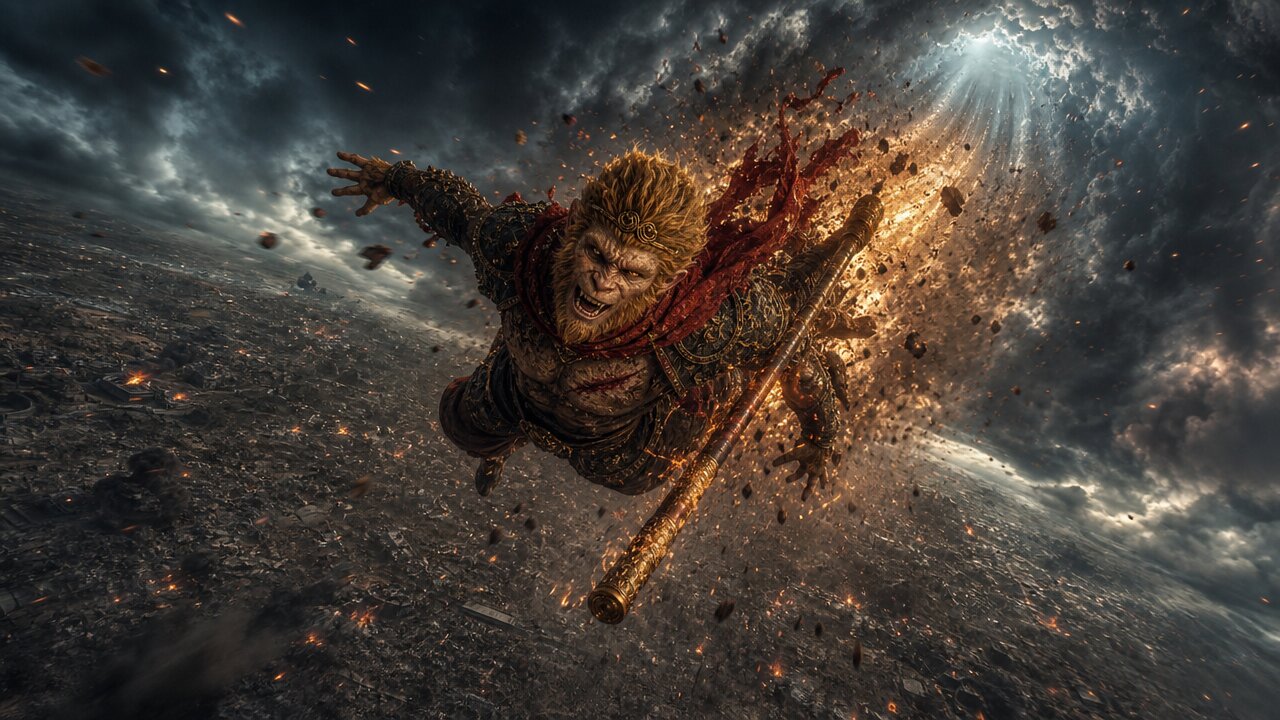}
  \caption{Starting anchor frame \texttt{a10\_wukong\_fall\_impact\_start} for Tianti shot~10.}
  \label{fig:demo-tianti-a10_wukong_fall_impact_start}
\end{figure}

\noindent\emph{Segments.}
\begin{itemize}[leftmargin=1.2em,topsep=2pt,itemsep=2pt,parsep=0pt]
  \item \texttt{seg\_shot10\_wukong\_fall\_impact\_00} (8\,s): Continuing the over-the-shoulder follow, Sun Wukong has a fresh blade-cut on his chest, his body tilted back in a high-speed fall, the ruyi-jingubang still glued to his side. The lens bites onto his trajectory as he punches through the heavy cloud cover; the broken cloud is dragged into a long white-grey tunnel, warm orange flame burns ever brighter behind him, and the cold blue high-altitude shadow is pulled away fast. Beneath, the mortal battlefield rapidly enlarges from a blurred grey-brown circle; outlines of dust and broken banners surface. End-frame: Sun Wukong wreathed in flame hangs above the battlefield, the dust ring of the impact already pre-flattened on the ground.
  \item \texttt{seg\_shot10\_wukong\_fall\_impact\_01} (6\,s): From the end-frame of Sun Wukong wreathed in fire above the field, the lens follows his last vertical drop into the grey-brown plain. At impact, the frame briefly shakes; warm orange flame slams into the cold grey dust ocean, and the ground caves in as if struck by a giant hammer. Circular cracks race outward; rubble, broken halberds and torn banners are flung up by the shock. The dust explodes into a heavy ring wall. End-frame: a giant circular crater dominates the centre of the field; debris and smoke roll outward and entirely obscure the bottom of the pit.
\end{itemize}

\paragraph{Shot~11: \texttt{shot11\_giant\_ape\_transformation} (18\,s).} \emph{Story goal:} Render Sun Wukong's wounded fury and complete the key state change from human-form monkey king to a mountain-sized giant ape.

\begin{figure}[H]
  \centering
  \includegraphics[width=\linewidth]{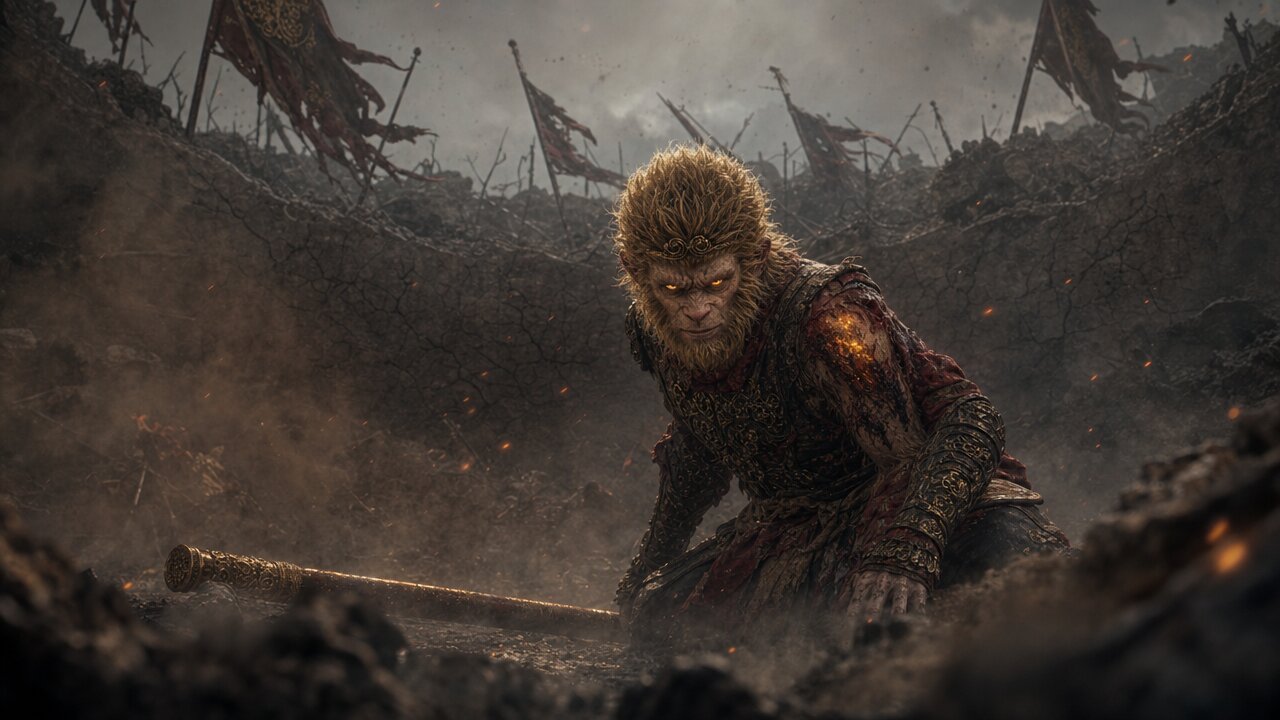}
  \caption{Starting anchor frame \texttt{a11\_giant\_ape\_transformation\_start} for Tianti shot~11.}
  \label{fig:demo-tianti-a11_giant_ape_transformation_start}
\end{figure}

\noindent\emph{Segments.}
\begin{itemize}[leftmargin=1.2em,topsep=2pt,itemsep=2pt,parsep=0pt]
  \item \texttt{seg\_shot11\_giant\_ape\_transformation\_00} (10\,s): Sun Wukong, still in human size, half-kneels at the bottom of the deep pit, gold light glowing from a wound on his shoulder, fierce light pressed into the smoke from his eyes. The lens fixes a low angle and pulls back slowly. He raises his head with a low growl; bone presses outward beneath the armour, the gold-brown fur thickens and hardens layer by layer, his shoulders and back swell, and broken stone falls from the pit wall in the wave of his breathing; the ruyi-jingubang at his side grows longer and heavier with him, gold-red veins lighting up in the smoke. End-frame: he has already grown into a half-giant-ape posture, one knee braced down, one hand gripping the enlarged ruyi-jingubang.
  \item \texttt{seg\_shot11\_giant\_ape\_transformation\_01} (8\,s): From the half-giant-ape stance Sun Wukong continues to grow upward from the pit. The lens still pulls back at a low angle, forced by his bulk to reveal a deeper, wider crater. The coarse gold-brown fur stands up like steel needles, the gold wound at his shoulder drips light along the fur, and the gold of his eyes turns the smoke into a fierce gold haze; the ruyi-jingubang expands into a pillar in synchrony, gripped tight by both fists and pressed down. Cracks burst outward at the bottom of the pit. End-frame: a mountain-scale giant ape stands inside the deep pit, the giant ruyi-jingubang in his hand, both fists and the staff pressed against the ground.
\end{itemize}

\paragraph{Shot~12: \texttt{shot12\_yangjian\_fatian} (18\,s).} \emph{Story goal:} Respond to Sun Wukong's giant-ape transformation by showing Yang Jian invoking fatian (cosmic-body magnification) at a matching divine scale to keep the duel balanced.

\begin{figure}[H]
  \centering
  \includegraphics[width=\linewidth]{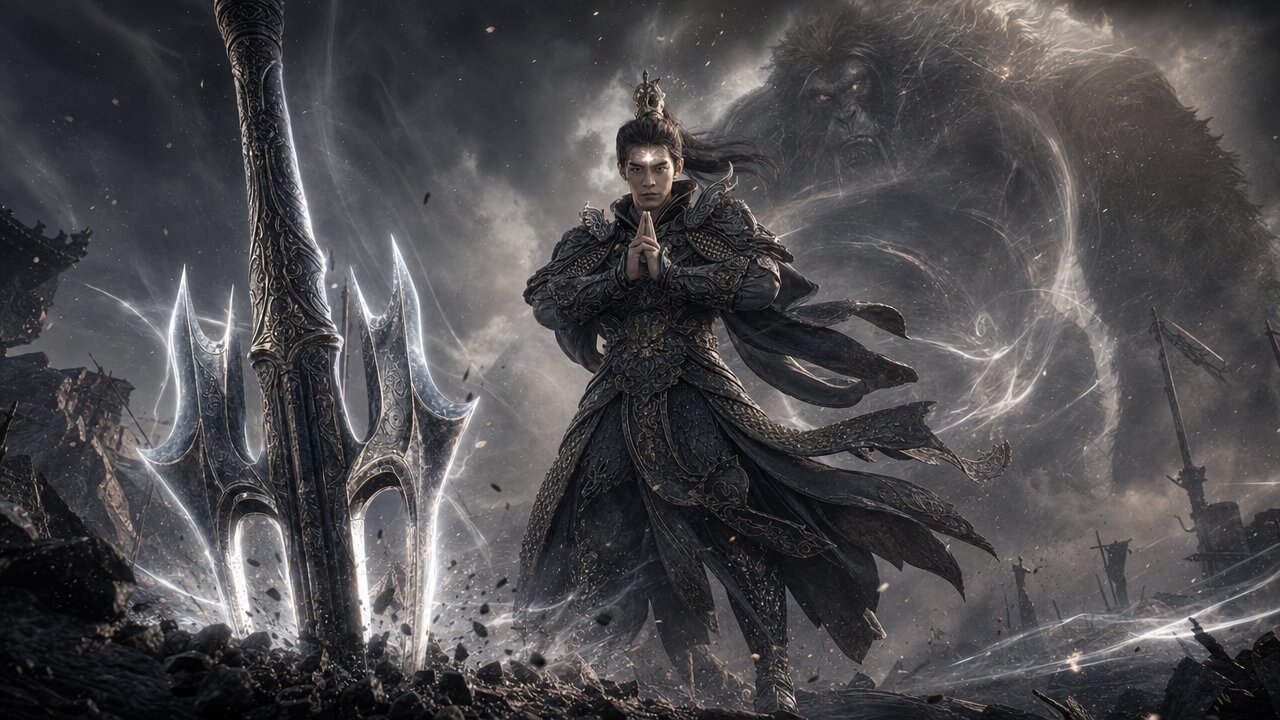}
  \caption{Starting anchor frame \texttt{a12\_yangjian\_fatian\_start} for Tianti shot~12.}
  \label{fig:demo-tianti-a12_yangjian_fatian_start}
\end{figure}

\noindent\emph{Segments.}
\begin{itemize}[leftmargin=1.2em,topsep=2pt,itemsep=2pt,parsep=0pt]
  \item \texttt{seg\_shot12\_yangjian\_fatian\_00} (9\,s): Yang Jian stands behind the three-pointed double-edged blade driven into the cracked earth, both hands forming a mudra, motionless; the cold-white aura at the third eye punches through the dust haze. The lens slowly rises around him and pulls back; the ambient qi of heaven and earth turns into rings of silver-white streamlines that converge on his armour, the hilt, and his brow, while a giant divine silhouette behind him solidifies from a thin outline. In the distance the giant-ape silhouette still presses through the dust; Yang Jian's body begins to grow many times taller, cracks fanning out beneath his feet. End-frame: a solemn picture of the divine silhouette hanging high and the giant blade beginning to extend.
  \item \texttt{seg\_shot12\_yangjian\_fatian\_01} (9\,s): From the end-frame of the suspended divine silhouette, the lens keeps pulling back and rising; the third eye between Yang Jian's brows burns steady as a cold-white divine sun, and the silhouette merges with his body. The silver-black armour expands layer by layer into mountain-scale plate; pauldrons, breastplate and tasset open heavily under the divine light; the three-pointed double-edged blade swells in sync as if pulled out of the ground, its edges becoming a sky-piercing giant blade. The cold-white divine glow overwhelms the dust-brown battlefield. End-frame: the deity-form Yang Jian stands sky-tall, the giant blade upright at his side, with the earth cracking beneath his feet.
\end{itemize}

\paragraph{Shot~13: \texttt{shot13\_titan\_clash} (22\,s).} \emph{Story goal:} Open the titan-scale melee between the giant ape and the cosmic-form deity, confirming that the battle has scaled from personal martial skill to world-shaking forces.

\begin{figure}[H]
  \centering
  \includegraphics[width=\linewidth]{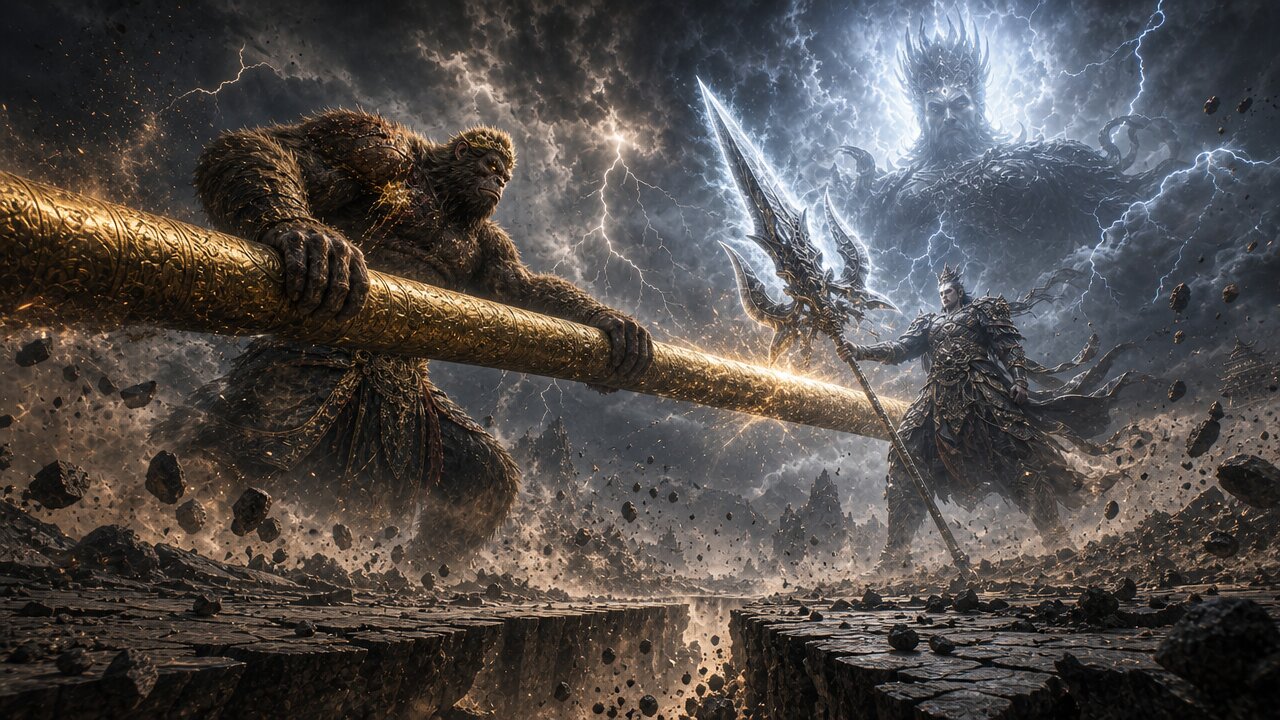}
  \caption{Starting anchor frame \texttt{a13\_titan\_clash\_start} for Tianti shot~13.}
  \label{fig:demo-tianti-a13_titan_clash_start}
\end{figure}

\noindent\emph{Segments.}
\begin{itemize}[leftmargin=1.2em,topsep=2pt,itemsep=2pt,parsep=0pt]
  \item \texttt{seg\_shot13\_titan\_clash\_00} (11\,s): From an extreme low-angle setup, the giant-ape Sun Wukong's eyes blaze gold; he swings the giant ruyi-jingubang horizontally into the raised sky-piercing blade of Yang Jian's deity form. At contact, gold staff-light and cold-white blade-light burst into a heavy shock ring, the cracks in the ground extend toward the foot of the lens, and rubble is thrown half-way up. The lens snaps to an extreme long shot; the two mountain-sized figures cross heavily through the dust storm, with thunderclouds overhead. End-frame: their giant weapons briefly part and they each lower their centre of gravity through rolling dust waves.
  \item \texttt{seg\_shot13\_titan\_clash\_01} (11\,s): From the end-frame of the two giant silhouettes facing off through dust, giant-ape Sun Wukong roars low and lunges in, with gold blood-light at the older shoulder wound, both hands swinging the giant ruyi-jingubang into a close-range body press. The lens hard-cuts from the extreme long shot to a near shot on the same line, where fist, fang-like fury and the staff's gold light close in on Yang Jian layer by layer; the deity-form Yang Jian nails his footing into the cracked earth, the cold-white third-eye light cuts through the smoke, and the sky-piercing blade absorbs the blow after blow. End-frame: the ruyi-jingubang and the three-pointed double-edged giant blade lock against each other again, while thunderclouds roll and cracks fan out from beneath their feet.
\end{itemize}

\paragraph{Shot~14: \texttt{shot14\_clone\_siege} (20\,s).} \emph{Story goal:} Introduce Sun Wukong's clone technique so Yang Jian is besieged by an indistinguishable swarm of true and false bodies, shifting the strategic pressure.

\begin{figure}[H]
  \centering
  \includegraphics[width=\linewidth]{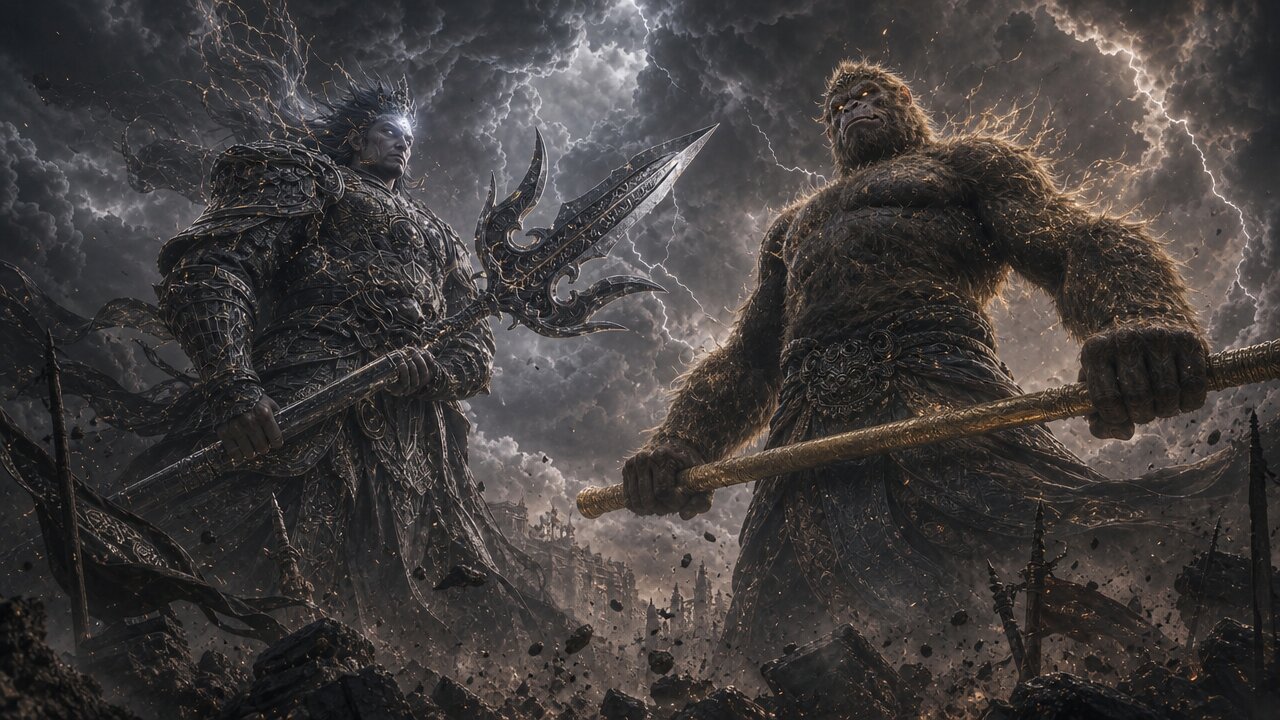}
  \caption{Starting anchor frame \texttt{a14\_clone\_siege\_start} for Tianti shot~14.}
  \label{fig:demo-tianti-a14_clone_siege_start}
\end{figure}

\noindent\emph{Segments.}
\begin{itemize}[leftmargin=1.2em,topsep=2pt,itemsep=2pt,parsep=0pt]
  \item \texttt{seg\_shot14\_clone\_siege\_00} (11\,s): Sweep horizontally from behind the cracked giant shoulder of the deity-form Yang Jian; the giant blade is held across the left of the frame, while the giant-ape Sun Wukong grins coldly, the gold blood-light still on his shoulder wound. The wind lifts the gold fine hairs on his body, and the hairs fall like glowing iron filings across the broken plain among torn banners, rubble and cloud shadow, and each landing lengthens into a translucent silhouette of Sun Wukong. The lens pans laterally and slowly; in the cold-grey dust the gold after-images stack layer by layer, and the true and false figures slowly fill the edges of the field. Yang Jian still holds his blade steady, the third eye dimmed yet fixed forward.
  \item \texttt{seg\_shot14\_clone\_siege\_01} (9\,s): From the end-frame of the field full of phantoms, the lens keeps sweeping over Yang Jian's shoulder onto a wider broken plain, the cold grey dust cut by countless gold after-images. The phantoms roar in unison; giant apes, human-form figures with the staff, decoys hidden behind boulders and inverted figures hanging in mid-air close in from every direction, and true and false ruyi-jingubangs form dense gold arcs. The focus is the moment when staff after-images skim Yang Jian's giant blade and armour: sparks burst in chains along the cracks, and the indistinguishable impact rattles a gold-white wave at the blade edge. The deity-form Yang Jian holds his footing in the cracked ground, the blade held in front, sparks freezing on the armour surface.
\end{itemize}

\paragraph{Shot~15: \texttt{shot15\_third\_eye\_scan} (16\,s).} \emph{Story goal:} Show Yang Jian's third eye breaking the true/false clones, turning Sun Wukong's supernatural advantage back against him.

\begin{figure}[H]
  \centering
  \includegraphics[width=\linewidth]{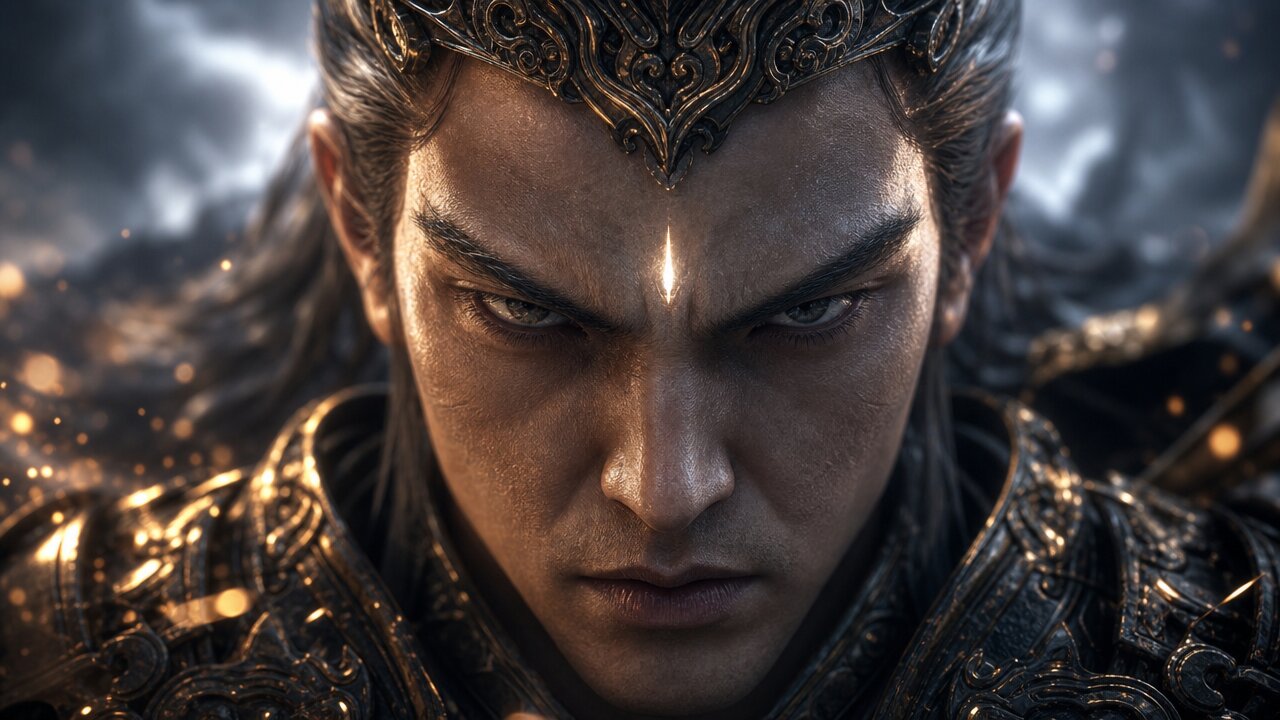}
  \caption{Starting anchor frame \texttt{a15\_third\_eye\_scan\_start} for Tianti shot~15.}
  \label{fig:demo-tianti-a15_third_eye_scan_start}
\end{figure}

\noindent\emph{Segments.}
\begin{itemize}[leftmargin=1.2em,topsep=2pt,itemsep=2pt,parsep=0pt]
  \item \texttt{seg\_shot15\_third\_eye\_scan\_00} (7\,s): Begin from an extreme close-up on the brow of Yang Jian's giant divine face. The third eye opens slowly along a fine gold line, and a cold-white divine glow expands from the inside out into a restrained divine sun. The lens pulls back slowly; the light is not an explosion but a thin scanning band, like the surface of water, passing across his cheek and crown. The gold after-image points around him are caught by the band and their edges shake and crumble into fine gold dust. End-frame: the third eye is fully open and the first layer of divine light spreads across the battlefield.
  \item \texttt{seg\_shot15\_third\_eye\_scan\_01} (9\,s): From the end-frame of the third eye's light fanning out, the lens keeps pulling out and pans across the broad broken plain. The cold-white scan light passes layer by layer over dust, rubble, banners and cloud shadow; hundreds of Sun Wukong phantoms slow inside the band, their edges flake away like paper ash and turn into gold fur and dust haze; the false bodies hidden among the staff after-images expose their flaws one by one and dissolve. End-frame: the wide battlefield in which the phantoms are nearly cleared, with the direction of the real giant ape exposed by the cold-white beam in the distance.
\end{itemize}

\paragraph{Shot~16: \texttt{shot16\_true\_body\_counter} (7\,s).} \emph{Story goal:} Let Yang Jian land a fast counter once the true body is locked, interrupting Sun Wukong's next sneak attack.

\begin{figure}[H]
  \centering
  \includegraphics[width=\linewidth]{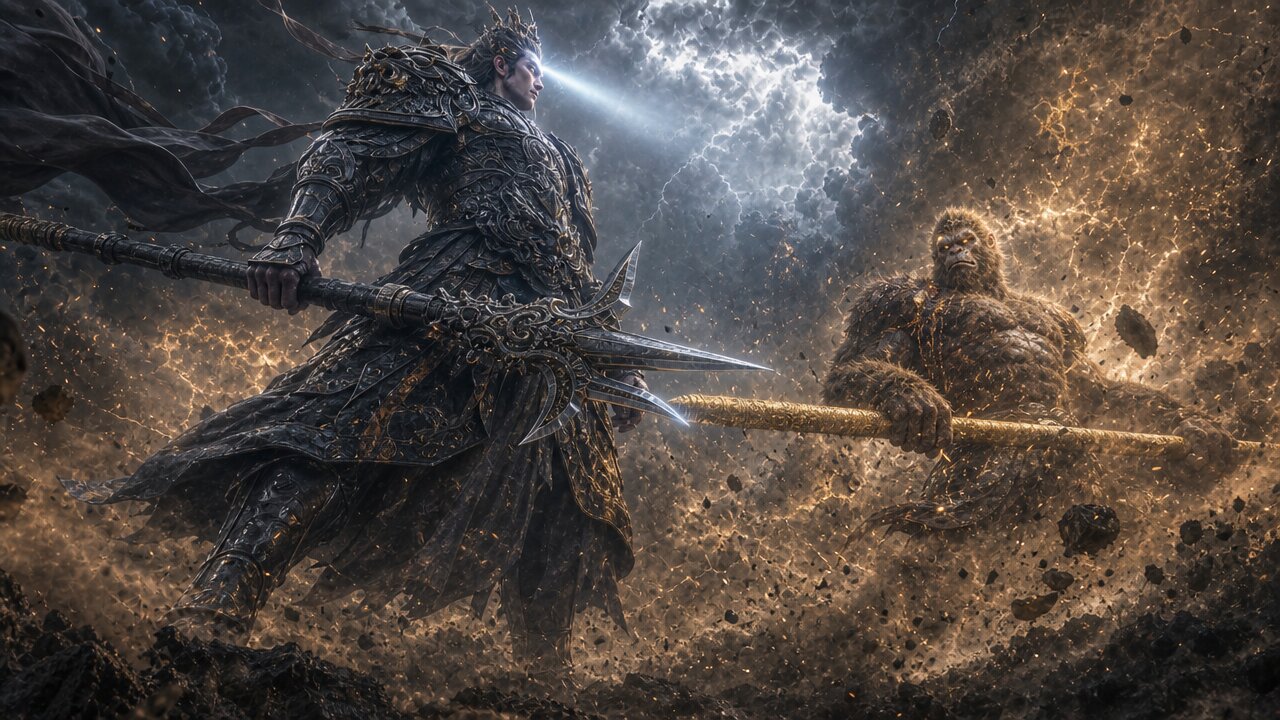}
  \caption{Starting anchor frame \texttt{a16\_true\_body\_counter\_start} for Tianti shot~16.}
  \label{fig:demo-tianti-a16_true_body_counter_start}
\end{figure}

\noindent\emph{Segments.}
\begin{itemize}[leftmargin=1.2em,topsep=2pt,itemsep=2pt,parsep=0pt]
  \item \texttt{seg\_shot16\_true\_body\_counter\_00} (7\,s): From the half-turn of the deity-form Yang Jian as the cold-white third-eye beam locks the true body to his rear, the lens whips in a fast arc along his giant shoulder. The three-pointed double-edged giant blade swings in a backhand cut, the cold-white edge slicing the dust-brown smoke; giant-ape Sun Wukong meets it with the ruyi-jingubang pressed flat to the ground, the staff lifting a ridge-like wave of earth. At impact, the staff and giant blade collide hard at the rear, and the screen-shake shock wave shatters the residual phantoms into a rain of gold fur. The giant ape is forced back several steps and stops, the soles of his feet ploughing deep trenches into the ground.
\end{itemize}

\paragraph{Shot~17: \texttt{shot17\_aftermath\_standoff} (18\,s).} \emph{Story goal:} Convey both sides' exhaustion, wounds, and undiminished will after the supernatural exchange, bringing the fight to its first real pause.

\begin{figure}[H]
  \centering
  \includegraphics[width=\linewidth]{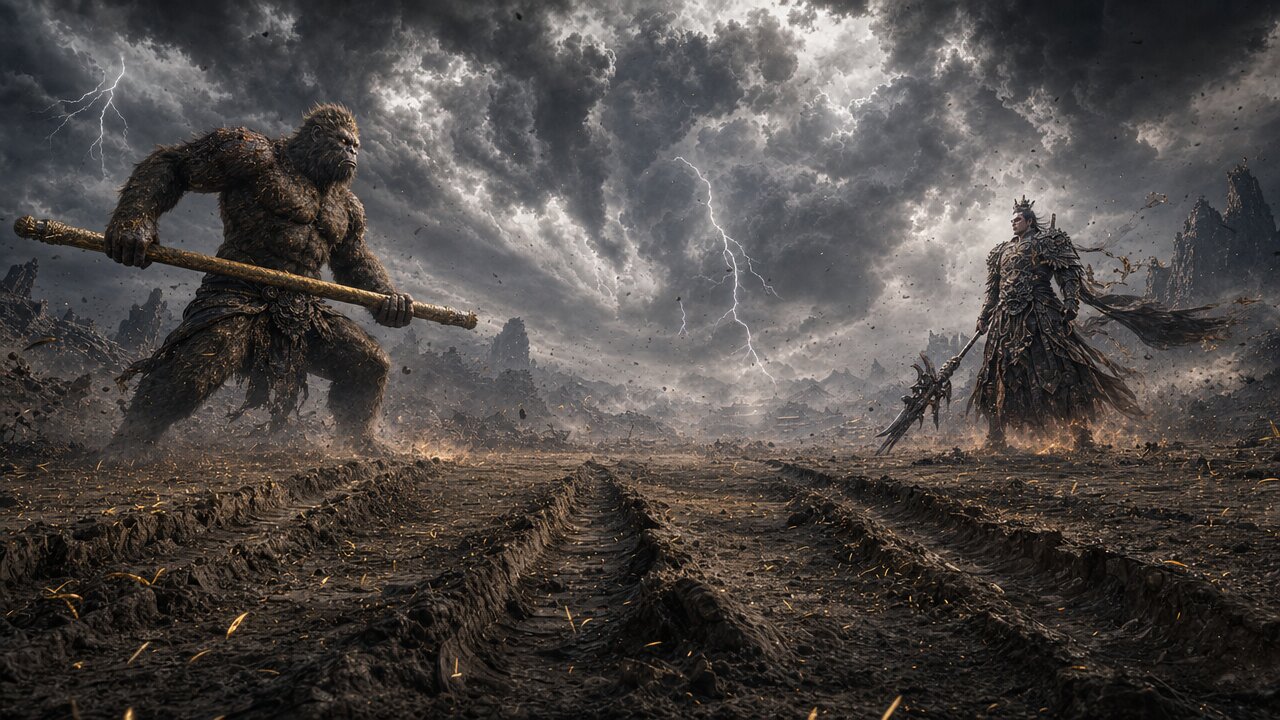}
  \caption{Starting anchor frame \texttt{a17\_aftermath\_standoff\_start} for Tianti shot~17.}
  \label{fig:demo-tianti-a17_aftermath_standoff_start}
\end{figure}

\noindent\emph{Segments.}
\begin{itemize}[leftmargin=1.2em,topsep=2pt,itemsep=2pt,parsep=0pt]
  \item \texttt{seg\_shot17\_aftermath\_standoff\_00} (10\,s): From an extreme wide shot, dolly in slowly. Gold fur falls like rain over the cracked ground, with deep trenches separating the giant-ape Sun Wukong from the deity-form Yang Jian. The dust waves settle layer by layer; the thunderclouds are torn into grey-black strata; the gold blood-light on Sun Wukong's older shoulder wound is faint, and the third eye on Yang Jian's brow is still lit but visibly dimmed. The lens lowers the pace under a cold-grey sky and finally lands on the two giant silhouettes facing off in silence across the broken battlefield.
  \item \texttt{seg\_shot17\_aftermath\_standoff\_01} (8\,s): The lens keeps moving in slowly; the framing remains very wide, but exhaustion now weighs more than action. The rise and fall of the giant ape's chest is sketched by cold-grey light, the gold blood-light on his shoulder dripping down the coarse fur to the ground; the deity-form Yang Jian does not pursue, the cracks in his silver-black plate growing visible, and a chip in the edge of the giant blade revealed in the faint light. End-frame: a still composition of wounded combatants holding a standoff across the field.
\end{itemize}

\paragraph{Shot~18: \texttt{shot18\_final\_ready} (12\,s).} \emph{Story goal:} Close this segment with the two combatants re-arming with no winner declared, emphasising that they are spent yet unwilling to retreat, presaging an even heavier next clash.

\begin{figure}[H]
  \centering
  \includegraphics[width=\linewidth]{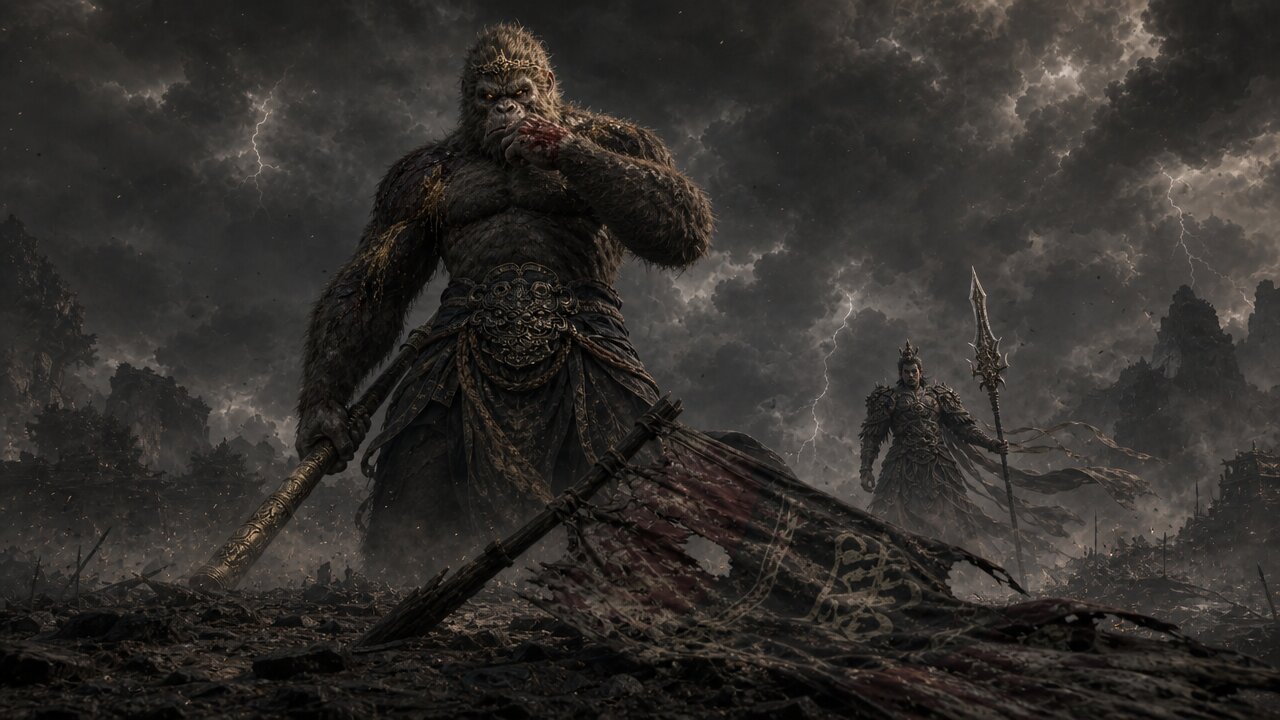}
  \caption{Starting anchor frame \texttt{a18\_final\_ready\_start} for Tianti shot~18.}
  \label{fig:demo-tianti-a18_final_ready_start}
\end{figure}

\noindent\emph{Segments.}
\begin{itemize}[leftmargin=1.2em,topsep=2pt,itemsep=2pt,parsep=0pt]
  \item \texttt{seg\_shot18\_final\_ready\_00} (12\,s): From a quiet image of broken banners settled on cracked ground, the lens cuts back and forth restrainedly between giant-ape Sun Wukong and deity-form Yang Jian. The gold blood mark on Sun Wukong's shoulder has not yet dried; he wipes the corner of his mouth and slowly tightens his grip on the giant ruyi-jingubang. Yang Jian's closed third eye pauses briefly and reopens in cold-grey lightning; the giant blade is brought upright. Low thunder and slowly settling dust hold the whole frame. End-frame: a standoff composition of the two, both lowering their centres of gravity again, ready to charge each other once more.
\end{itemize}

\subsection{Dumplings}
\label{app:demo-dumplings}

\noindent Dumplings: a Lunar New Year father--daughter reconciliation in the family kitchen. This sequence carries 3~portrait, 1~location, and 5~prop reference assets, plus 8~shots, 8~starting anchors, 15~generator segments, totalling 133~s of target footage.

\paragraph{Original generation prompt (translated).}
\begin{quote}
\small
In an old kitchen on Lunar New Year's Eve, an elderly father and his adult daughter stand side by side making dumplings. The sound of the television comes faintly from the living room, and the rest of the family has not returned yet. The daughter has not spent New Year at home for three years. She has come back not to stay, but to persuade her father to sell this apartment, which has become too old. The kitchen is narrow. Flour covers the cutting board; beside it are only a basin of filling, a bowl of water, and several rows of dumplings that have not yet gone into the pot. The father wraps slowly, pinching every pleat with excessive care. The daughter is skilled with her hands, but she never looks at him, as if one glance would make too many old memories return.

The father first passes a rolled dumpling wrapper to her side. He does not ask directly how she has been these years; he only reminds her not to put in too much filling. She takes the wrapper and moves quickly, but makes one dumpling crooked. She means to set it aside and ignore it. The father sees it, but unlike when she was a child, he does not correct her. He simply pushes the bowl of water a little closer to her hand. That motion makes her stop. She dips her fingers into the water and seals the edge again, but still does not pick up the conversation.

The kitchen contains many traces of time that have never been cleared away. Old photographs on the wall have faded. The clock is stopped at the afternoon when her mother died. On the windowsill still sits the enamel cup the daughter used when she was in school. Each time the father reaches for filling, he avoids the cup, as if afraid of knocking it over. The daughter notices that all these things are still here. Her tone slowly hardens as she says the house leaks, the stairs are too steep, and it is unsafe for one person to live here. She says these things without raising her head, placing each finished dumpling along the edge of the plate as if arranging evidence.

The father does not argue right away. He keeps making dumplings. He says the roof has already been repaired, the stair light has been replaced, and the kitchen window only lets in a little wind in winter. The daughter finally stops and looks toward the taped seam by the window, where the sound of firecrackers leaks in from outside. She says these small repairs cannot solve the problem. Only after speaking does she realize that her voice is colder than the countdown on the television. The wrapper in the father's hand has been pinched too tightly and has split at the edge. He lowers his head, spreads it flat again, and does not look at her.

The disagreement never becomes a loud quarrel. It turns into a longer silence. The father places the best dumpling he has made on her side, as if nothing has happened. The daughter looks down at it and suddenly sees that its pleats are exactly like the way her mother used to make them: three presses from the left, a closing fold from the right, and a gentle press in the middle. She remembers that when she was little, her mother always put the ugly dumplings into her bowl and said they were the luckiest ones. The daughter does not say this memory aloud. She only takes back the crooked dumpling she made earlier and re-pleats it by following her father's pattern.

The father sees her repairing the dumpling, and his hands slow down as well. He does not praise her. He simply moves the empty plate closer to her so she can place the repaired dumpling in the centre. From the television comes the rehearsal sound of the Spring Festival countdown; someone outside sets off firecrackers early, and the kitchen glass trembles softly. The daughter looks toward the pencil height marks on the door frame, left from when her height was measured every year as a child. Beside the highest line there is still her mother's handwriting. She reaches out and touches the mark, leaving a little flour on her finger.

When the water boils, the father slides rows of dumplings into the pot. At first the daughter only stands beside him and watches. Then she picks up the slotted spoon and gently moves the dumplings through the water so they will not stick to the bottom. Steam rises, hiding both of their faces and softening the old traces around the kitchen. The father turns the heat down; the daughter sets the lid askew so the steam can escape from one side. Their motions finally begin to cooperate, no longer separate as they were at the beginning.

After the dumplings are cooked, the father first ladles the re-pleated dumpling into her bowl. The daughter does not eat immediately. She lowers her head and watches it float slowly in the hot broth. She does not mention selling the apartment again, and she does not say she will stay. In the white steam, she only says very softly that next year they could repair the kitchen window first. The father hears her and does not press for more. He only turns the heat down a little more, as if afraid the sentence might be scattered by the steam. The ending rests beside the cutting board in the old kitchen: a little flour remains on the empty plate, the height marks on the door frame have not been erased, and father and daughter stand side by side at the stove as the steam from the pot rises quietly.
\end{quote}

\subsubsection*{Cast and setting}

\paragraph{Portraits.} Per-character T2I reference frames produced before generation; segments later reference these via the \texttt{reference\_inputs.portrait} field.

\begin{figure}[H]
  \centering
  \begin{subfigure}[t]{0.24\textwidth}
    \centering
    \includegraphics[width=\linewidth]{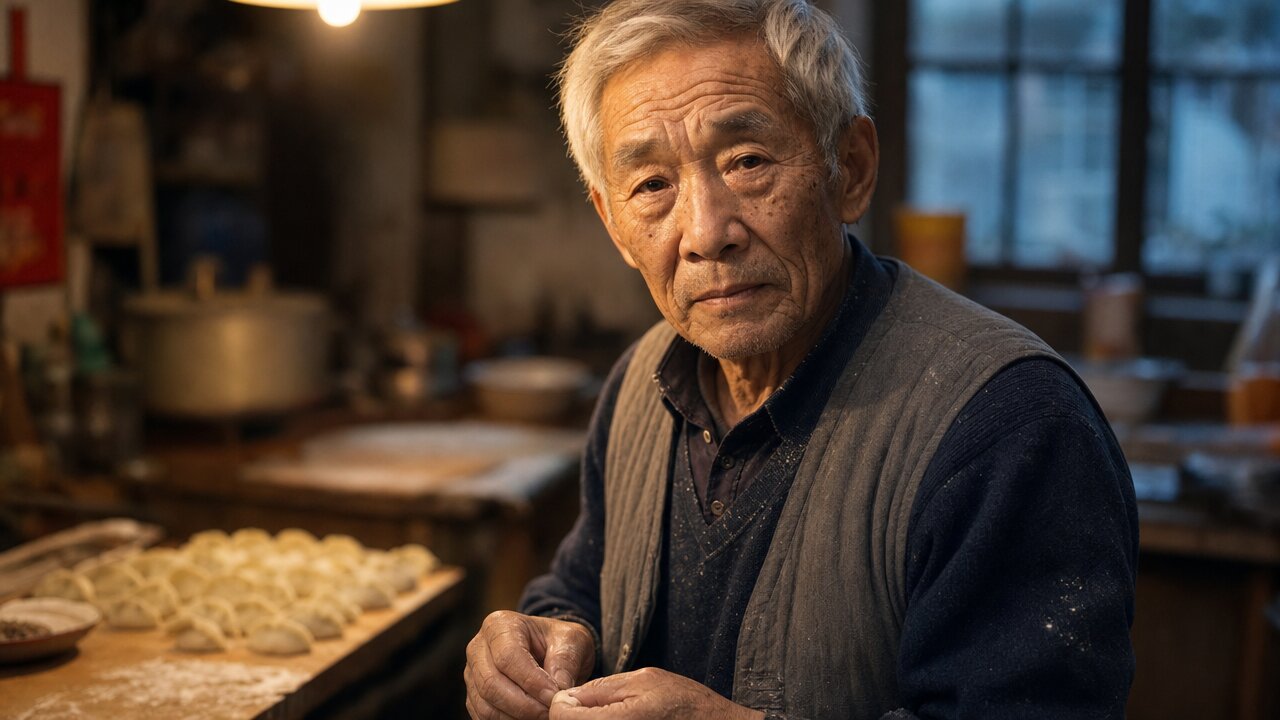}
    \caption*{\texttt{old\_father}}
  \end{subfigure}
  \hfill
  \begin{subfigure}[t]{0.24\textwidth}
    \centering
    \includegraphics[width=\linewidth]{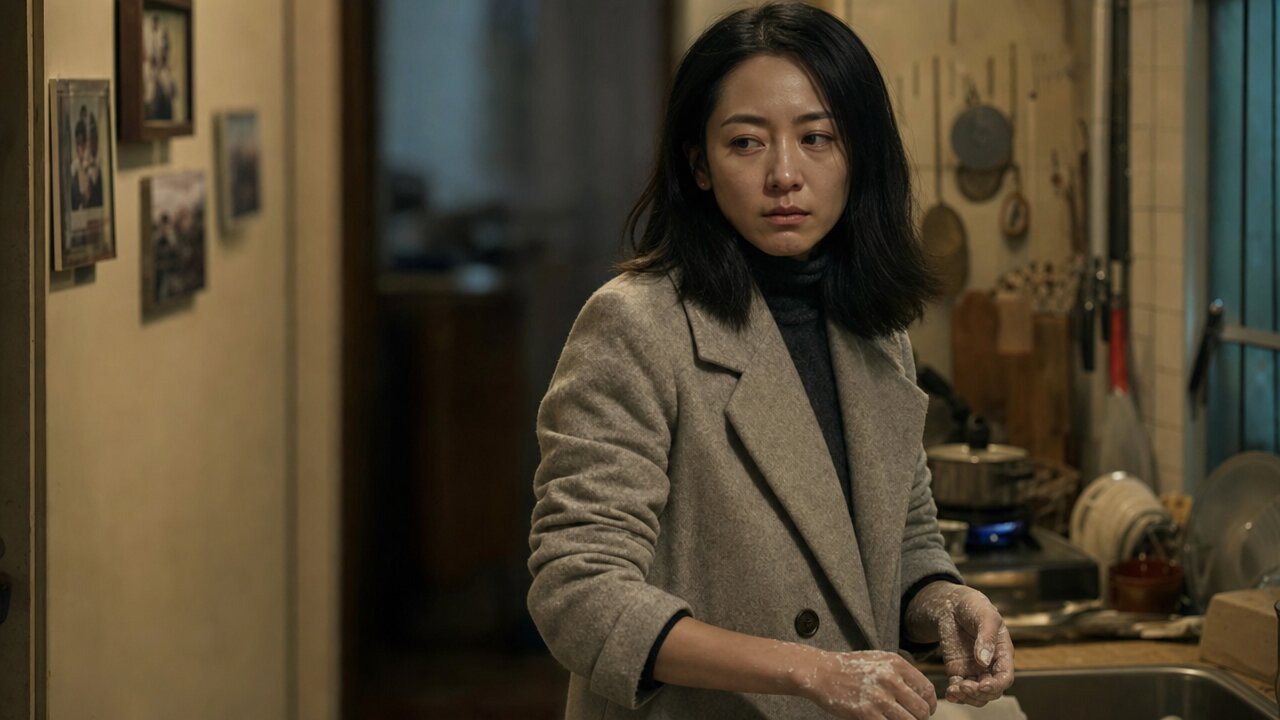}
    \caption*{\texttt{adult\_daughter}}
  \end{subfigure}
  \hfill
  \begin{subfigure}[t]{0.24\textwidth}
    \centering
    \includegraphics[width=\linewidth]{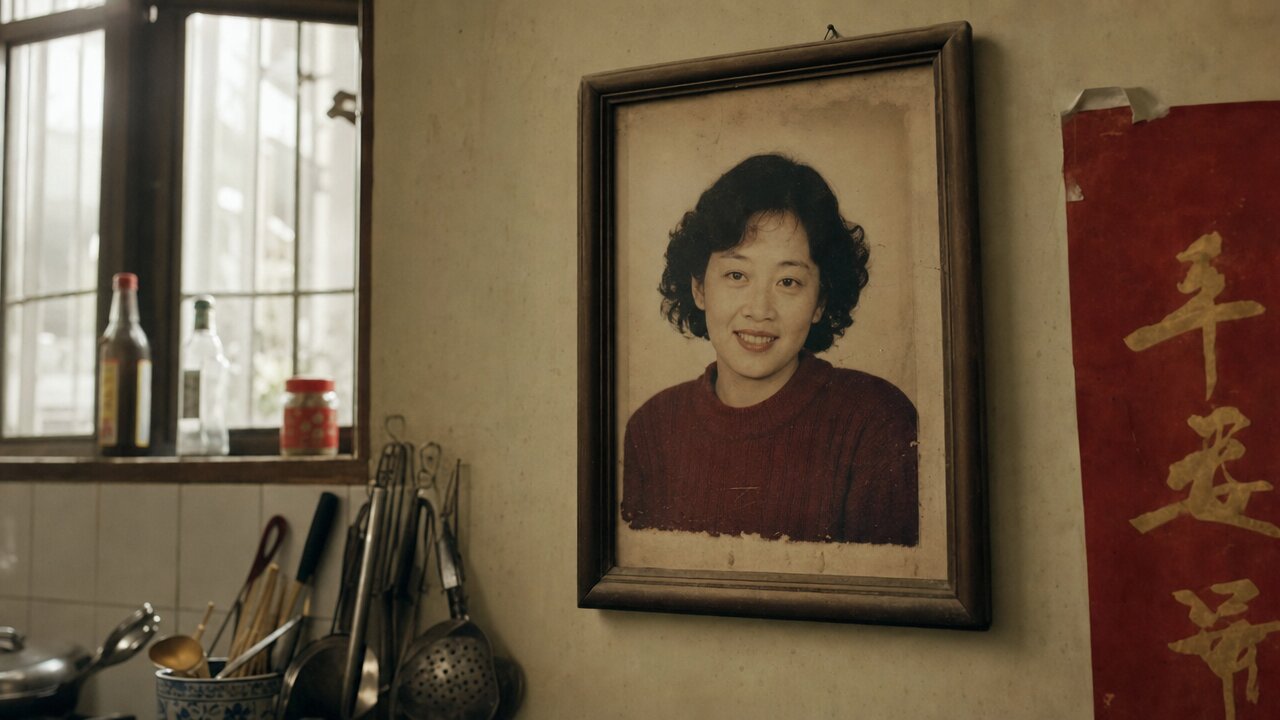}
    \caption*{\texttt{late\_mother\_photo}}
  \end{subfigure}
  \caption{Dumplings portrait references.}
  \label{fig:demo-dumplings-cast-portraits}
\end{figure}

\begin{itemize}[leftmargin=1.2em,topsep=2pt,itemsep=2pt,parsep=0pt]
  \item \texttt{old\_father}: Cinematic realistic character portrait. An elderly Chinese father in an old kitchen on Lunar New Year's Eve, around seventy, thin but with a still-upright posture; grey-white short hair; deep wrinkles and faint age spots on the face; an old dark blue sweater with a grey cotton vest, with a little flour on the cuffs; coarse hands with fine cracks; a restrained, gentle expression; a quiet exhaustion in the eyes from long-time solitary living. Warm yellow ceiling lamp blended with cool blue window light. 50\,mm shallow depth-of-field; the cutting board and dumplings blur in the background.
  \item \texttt{adult\_daughter}: Cinematic realistic character portrait. A Chinese adult daughter in her thirties who is back home for Lunar New Year's Eve after years of working away; black shoulder-length hair with a slightly tired sheen; a beige-grey wool overcoat over a dark turtleneck, cuffs rolled up, flour on her hands. Her face is held in check, her gaze flickering and her mouth tight, controlled. The urban look of her clothes contrasts with the old family kitchen; under warm yellow light there is a slight cool blue rim from the window. 85\,mm shallow depth-of-field; the narrow kitchen wall with old photos blurs behind.
  \item \texttt{late\_mother\_photo}: Cinematic realistic reference of an old photograph: the faded family photo of a deceased Chinese mother, in her forties, gentle and unadorned, short curly hair, in a 1990s dark red sweater, smiling at the camera. The photo paper is yellowed with fine scratches and faded edges, appearing inside an old frame on the kitchen wall, not as a living person. Soft natural light, nostalgic family-documentary feel.
\end{itemize}

\paragraph{Locations.} Per-location reference frames; segments later reference these via \texttt{reference\_inputs.place}.

\begin{figure}[H]
  \centering
  \begin{subfigure}[t]{0.323\textwidth}
    \centering
    \includegraphics[width=\linewidth]{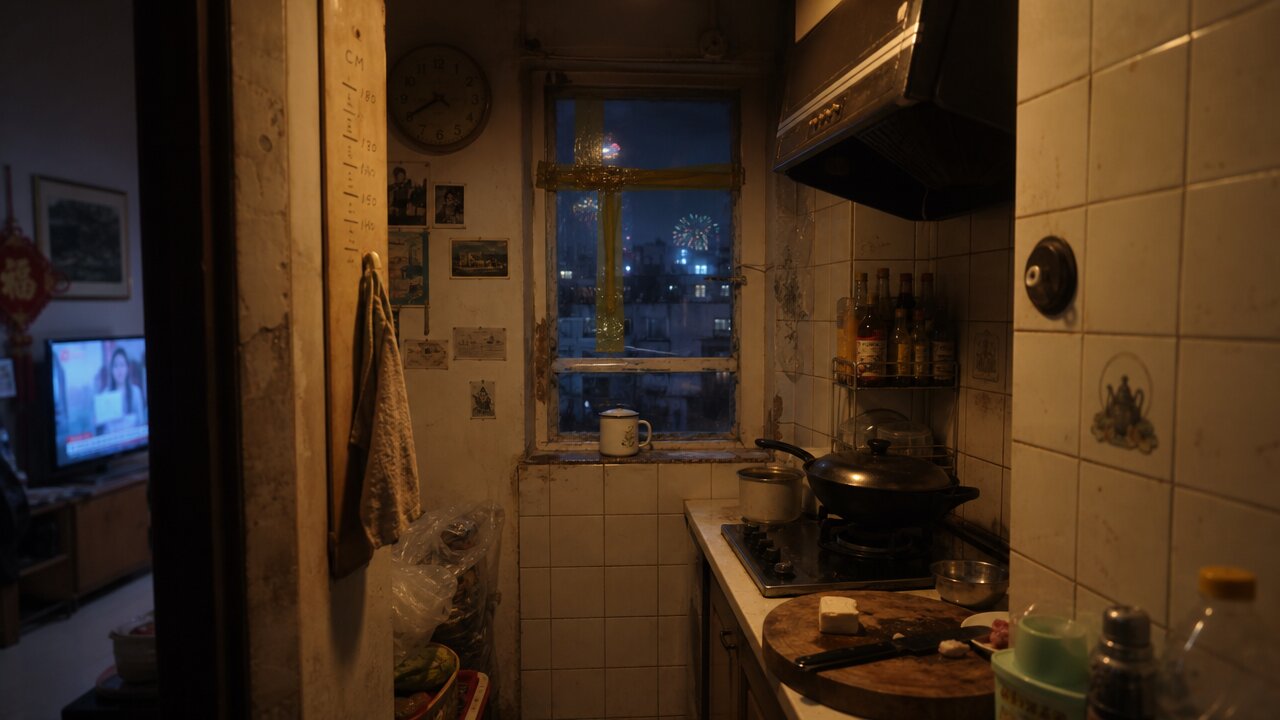}
    \caption*{\texttt{old\_kitchen}}
  \end{subfigure}
  \caption{Dumplings location reference.}
  \label{fig:demo-dumplings-cast-locations}
\end{figure}

\begin{itemize}[leftmargin=1.2em,topsep=2pt,itemsep=2pt,parsep=0pt]
  \item \texttt{old\_kitchen}: The narrow kitchen of an old apartment block on Lunar New Year's Eve: tight in space, with yellowed tiles, an old wooden cutting board, an old gas stove, taped-up window seams, a stopped wall clock, faded photos, an old enamel cup on the windowsill, and door-frame height marks: together the long-uncleared traces of family time. The warm yellow ceiling lamp is dim; the cool blue night and distant fireworks light occasionally come through the window; the living-room TV is far in the background. Overall: cinematic realistic, restrained, intimate, with a slight humid and oily kitchen feel.
\end{itemize}

\paragraph{Props.} Per-prop reference frames; segments later reference these via \texttt{reference\_inputs.prop}.

\begin{figure}[H]
  \centering
  \begin{subfigure}[t]{0.24\textwidth}
    \centering
    \includegraphics[width=\linewidth]{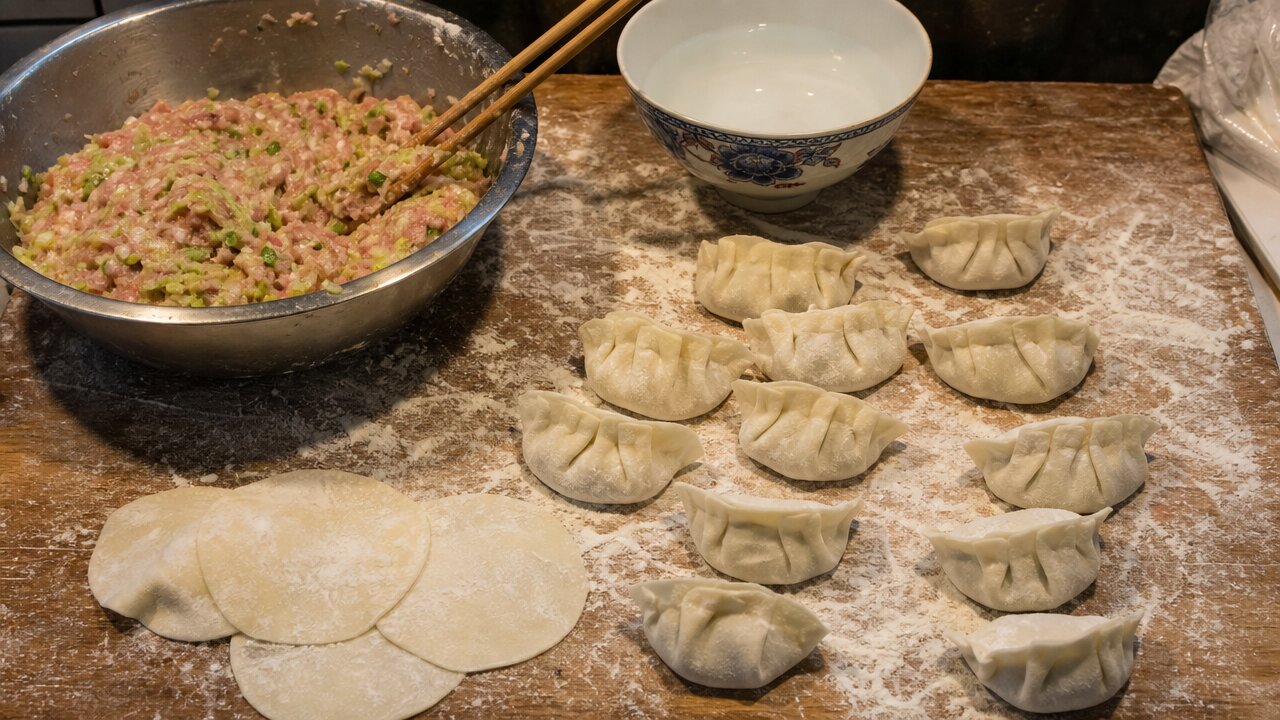}
    \caption*{\texttt{dumpling\_setup}}
  \end{subfigure}
  \hfill
  \begin{subfigure}[t]{0.24\textwidth}
    \centering
    \includegraphics[width=\linewidth]{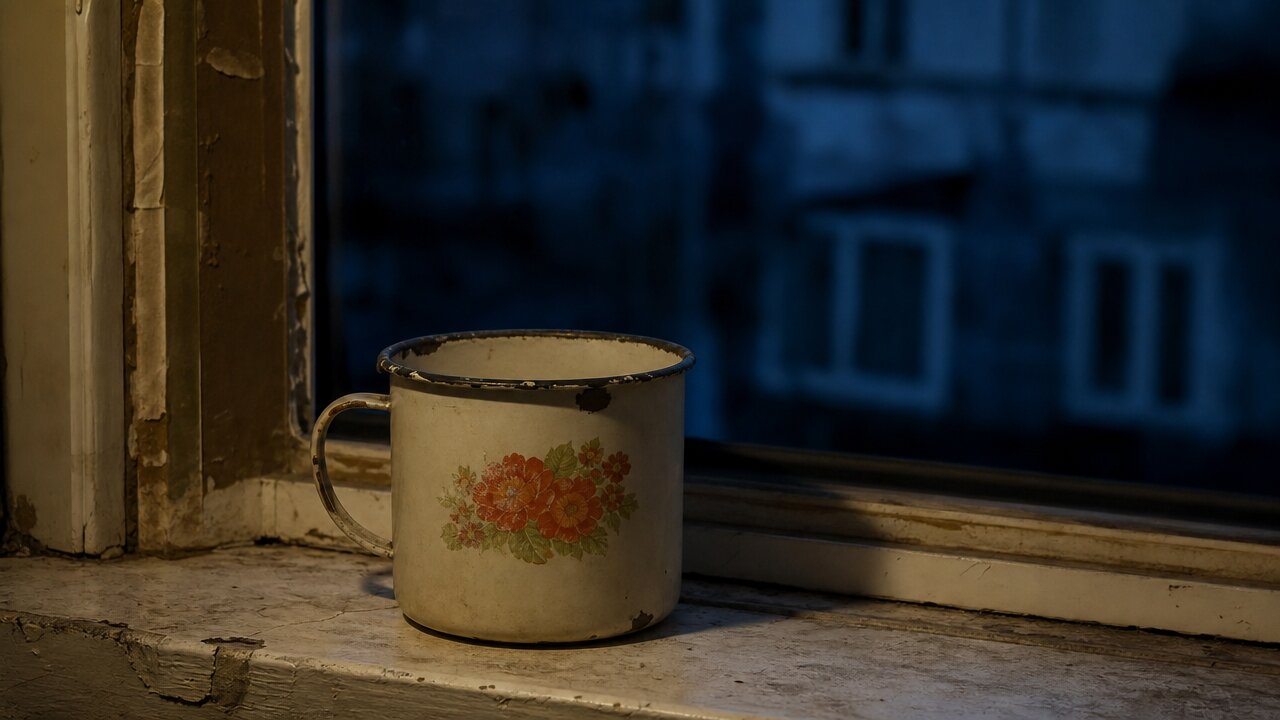}
    \caption*{\texttt{enamel\_cup}}
  \end{subfigure}
  \hfill
  \begin{subfigure}[t]{0.24\textwidth}
    \centering
    \includegraphics[width=\linewidth]{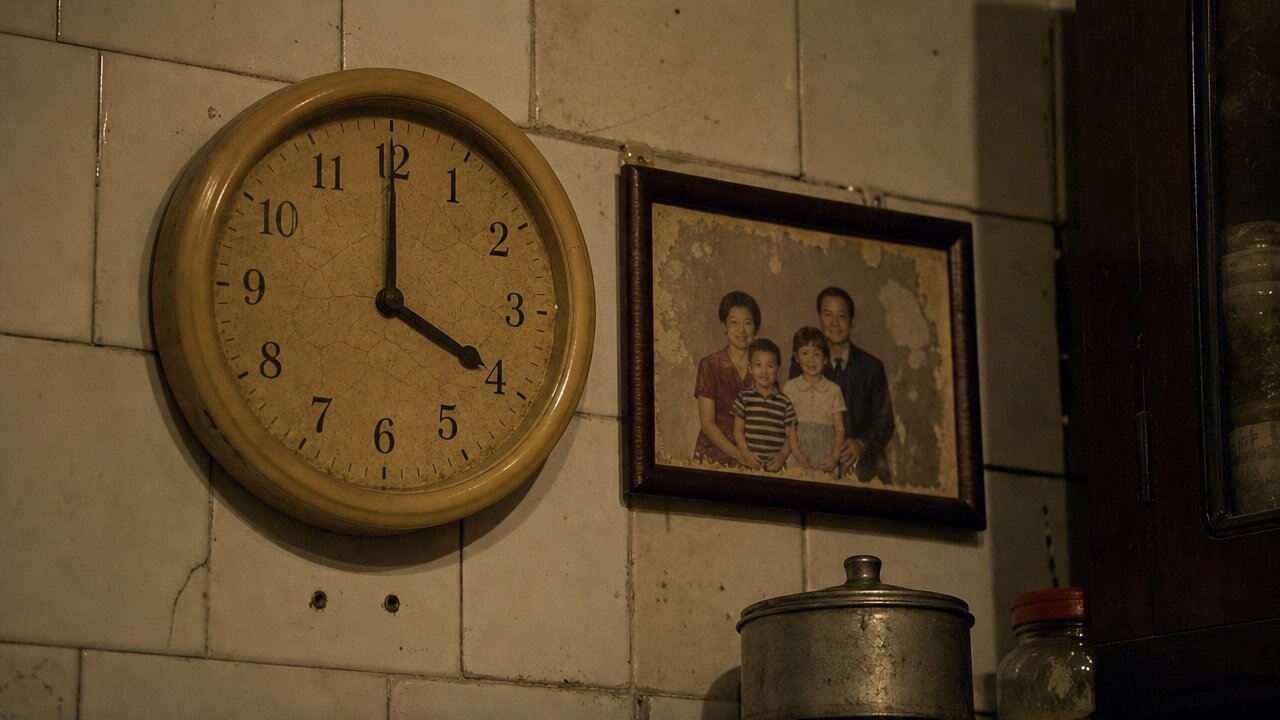}
    \caption*{\texttt{stopped\_clock}}
  \end{subfigure}
  \hfill
  \begin{subfigure}[t]{0.24\textwidth}
    \centering
    \includegraphics[width=\linewidth]{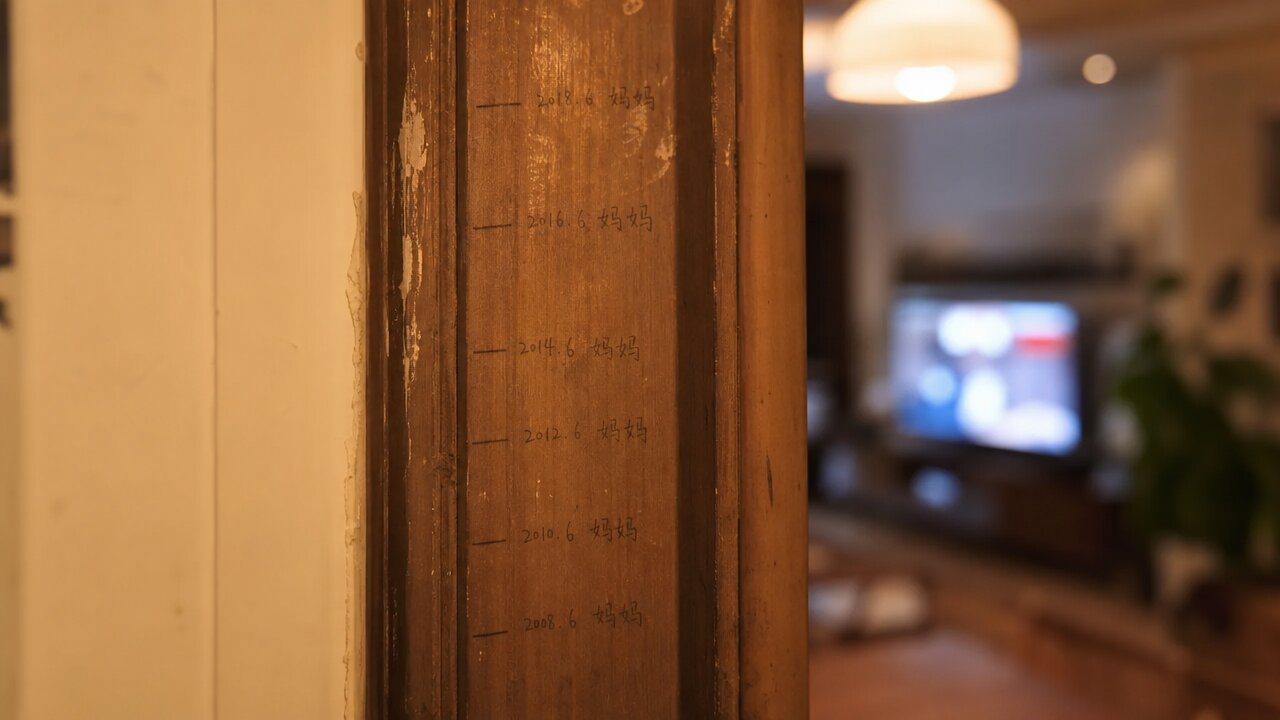}
    \caption*{\texttt{height\_marks}}
  \end{subfigure}
  \\[2pt]
  \begin{subfigure}[t]{0.24\textwidth}
    \centering
    \includegraphics[width=\linewidth]{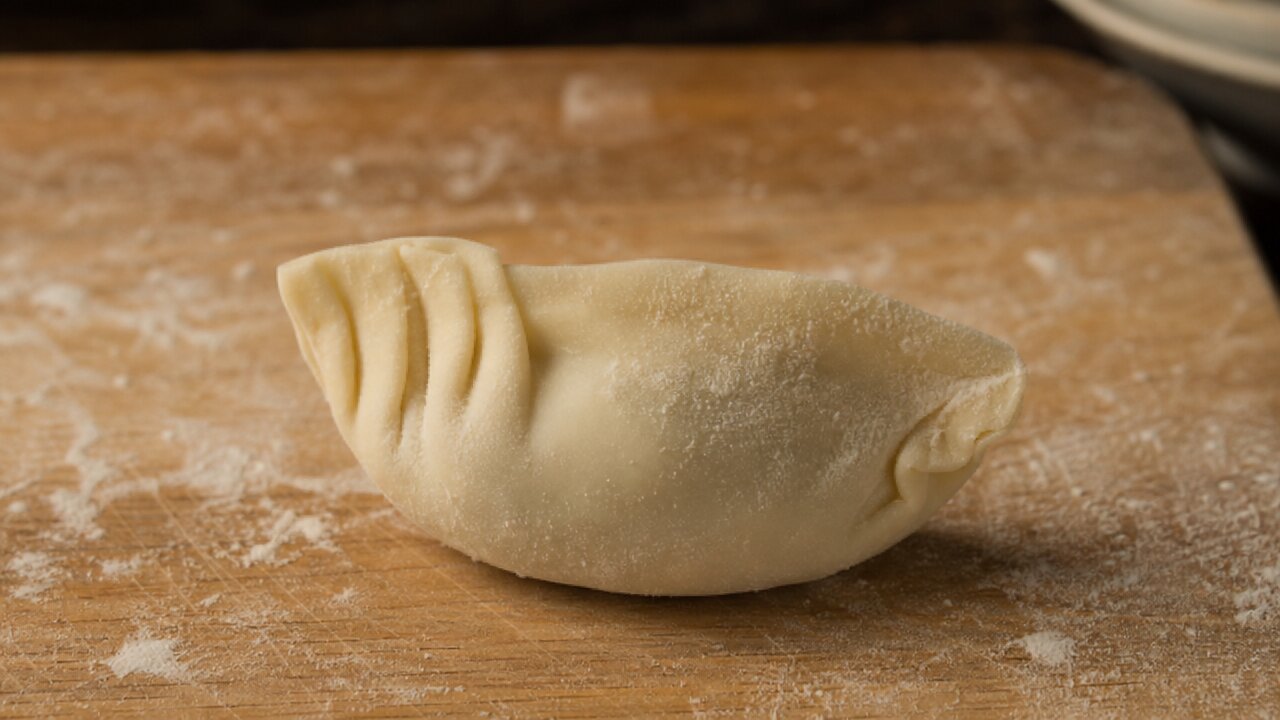}
    \caption*{\texttt{repaired\_dumpling}}
  \end{subfigure}
  \caption{Dumplings prop references.}
  \label{fig:demo-dumplings-cast-props}
\end{figure}

\begin{itemize}[leftmargin=1.2em,topsep=2pt,itemsep=2pt,parsep=0pt]
  \item \texttt{dumpling\_setup}: Static prop reference: an old wooden cutting board lightly dusted with flour, a bowl of pork-and-cabbage filling, a bowl of clean water, a few rolled dumpling skins, several rows of hand-folded dumplings (some neatly pleated, some slightly uneven), a ceramic plate with old-pattern edges; realistic family-documentary feel under warm yellow kitchen light.
  \item \texttt{enamel\_cup}: Static prop reference: a 1990s student-style white enamel cup, with chipped paint at the rim and a faded red flower pattern on the body; on the windowsill of an old kitchen, with thin dust on its surface; window-frame tape and cool blue night blur beside it, carrying the trace of an older time.
  \item \texttt{stopped\_clock}: Static prop reference: an old-style round wall clock, plastic case yellowed, dial slightly cracked, hands stopped at an afternoon hour; hung on the tiled wall of the old kitchen, with a faded family photo beside it; under low warm yellow light, a family-documentary feel of frozen time.
  \item \texttt{height\_marks}: Static prop reference: a row of pencil height lines on an old wooden door frame, with small handwritten years and the mother's writing beside them; the paint is worn and dark, with a few flour fingerprints near the highest mark; texture clear under warm yellow kitchen light, the living-room TV light blurred behind.
  \item \texttt{repaired\_dumpling}: Static prop reference: a hand-folded dumpling placed at the centre of an old ceramic bowl or cutting board; pleats are pressed three times on the left, closed on the right and lightly pressed flat in the middle; slight imperfection at the edge from having been bent and then mended; a little flour on the surface; plain home-cooking feel under warm yellow light.
\end{itemize}

\subsubsection*{Shots}

\paragraph{Shot~1: \texttt{shot01\_new\_year\_kitchen} (28\,s).} \emph{Story goal:} Establish the distance between long-separated father and adult daughter on Lunar New Year's Eve, with the old kitchen as the central container of family memory.

\begin{figure}[H]
  \centering
  \includegraphics[width=\linewidth]{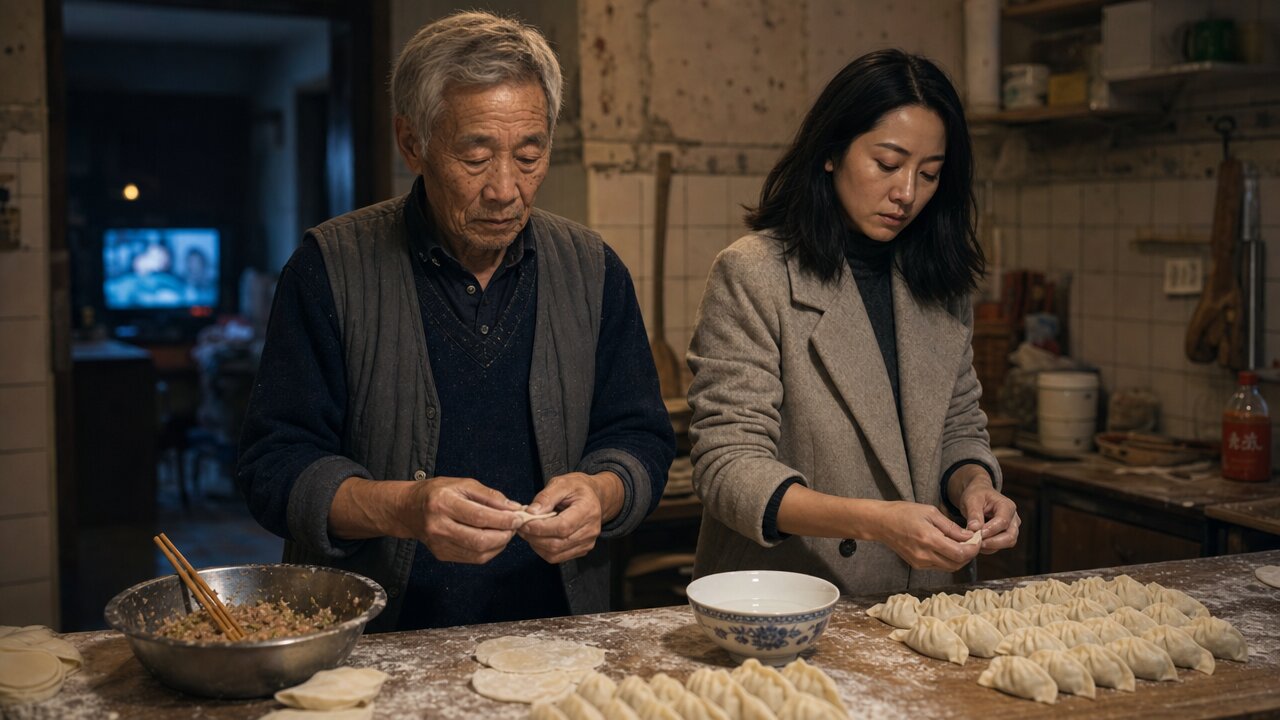}
  \caption{Starting anchor frame \texttt{a01\_new\_year\_kitchen\_start} for Dumplings shot~1.}
  \label{fig:demo-dumplings-a01_new_year_kitchen_start}
\end{figure}

\noindent\emph{Segments.}
\begin{itemize}[leftmargin=1.2em,topsep=2pt,itemsep=2pt,parsep=0pt]
  \item \texttt{seg\_shot01\_new\_year\_kitchen\_00} (10\,s): From a foreground of flour, a bowl of clean water and rows of raw dumplings on the cutting board, the lens pans slowly along the narrow kitchen counter and dollies in lightly. The elderly father, in an old dark blue sweater and a grey vest, holds a dumpling skin and looks down to add filling, pinching the pleats slowly. The adult daughter, in a beige overcoat with the cuffs rolled up, keeps wrapping with her head down, her gaze always avoiding her father. The warm yellow ceiling lamp presses on yellowed tiles, while the cool blue glow of the living-room TV sweeps lightly past the doorway. End-frame: at their hands, the father's first dumpling with fine pleats lands on the left of the board, the dumpling skin in the daughter's hand half-folded.
  \item \texttt{seg\_shot01\_new\_year\_kitchen\_01} (9\,s): Following on from the hand position of the previous frame, the lens keeps panning very slowly along the counter, with the weight on two pairs of flour-dusted hands. The father's coarse fingers under the dark blue cuff press each pleat too earnestly; every dumpling edge is clamped tight. The daughter's beige cuff is rolled up; her movements are faster and more practised, but her face stays down, gaze never crossing the board. The bowl of clean water sits between them like a quiet boundary; the filling bowl and the flour are aged-looking under the warm light. End-frame: in the middle of the board the father has added a few neatly pleated dumplings, while two or three slightly hurried-looking ones sit in front of the daughter.
  \item \texttt{seg\_shot01\_new\_year\_kitchen\_02} (9\,s): From the rows of dumplings in the middle of the board, the lens lifts slightly to include both the hands and the side-by-side bodies in the frame. The father, still in his dark blue sweater and grey vest, leans his back and shoulders restrainedly toward the counter and slowly finishes the dumpling in his hands; the daughter, her beige cuff dusted with flour, deftly arranges the new dumplings in front of herself but does not turn to look at him. The warm yellow lamp, the yellowed tiles and the cool blue TV light keep the quiet distance. End-frame: a medium shot of father and daughter side by side, a few finely pleated dumplings beside his hand and several in front of her, the two close in space yet their gazes still not meeting.
\end{itemize}

\paragraph{Shot~2: \texttt{shot02\_wrapped\_wrong} (12\,s).} \emph{Story goal:} Through the father's indirect remark about not over-stuffing the filling, and the daughter's misshapen dumpling, render the unspoken care and resistance neither can voice.

\begin{figure}[H]
  \centering
  \includegraphics[width=\linewidth]{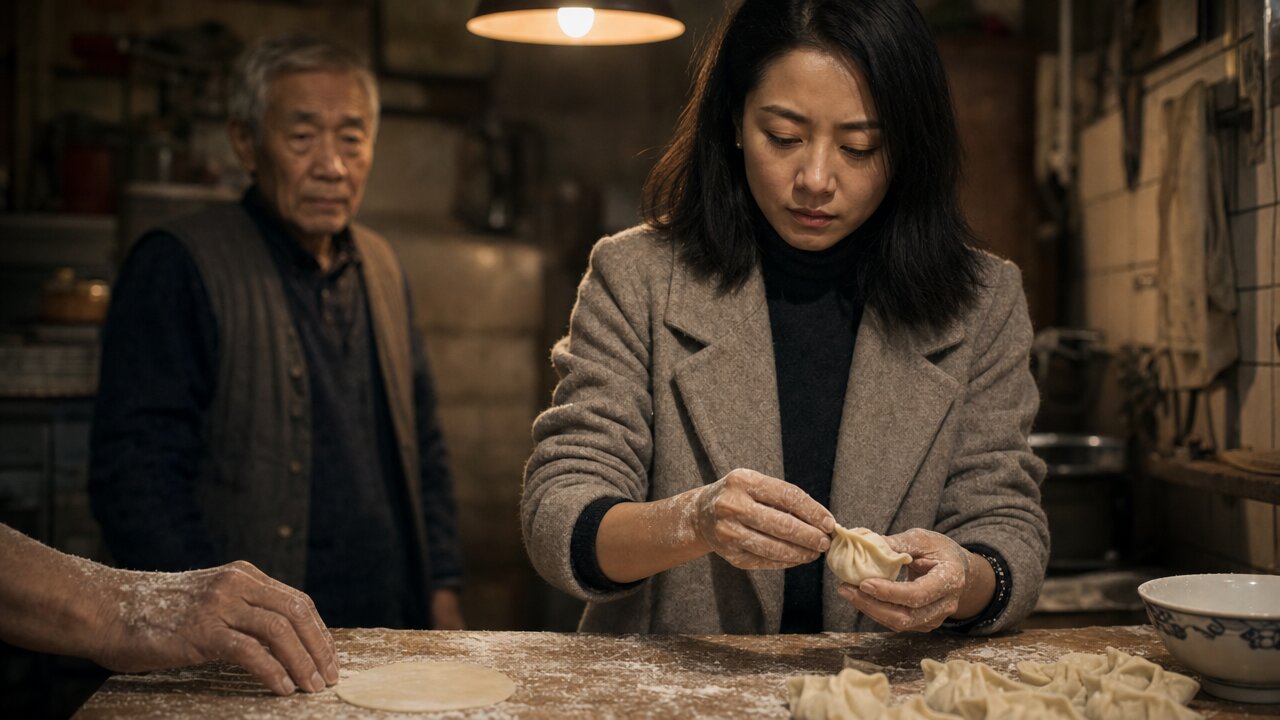}
  \caption{Starting anchor frame \texttt{a02\_wrapped\_wrong\_start} for Dumplings shot~2.}
  \label{fig:demo-dumplings-a02_wrapped_wrong_start}
\end{figure}

\noindent\emph{Segments.}
\begin{itemize}[leftmargin=1.2em,topsep=2pt,itemsep=2pt,parsep=0pt]
  \item \texttt{seg\_shot02\_wrapped\_wrong\_00} (6\,s): From a near shot, catch the misshapen dumpling: the daughter's face is down, her cuff dusted with flour, and the father's half body is still blurred in the background. The father does not look up to ask; he only softly notes that the filling should not be too much, his wrinkled hand reaching slowly from the left across the counter to the back right. The lens slightly racks focus to the bowl of clean water; the warm yellow lamp presses the awkward silence; the bowl is pushed to her side and stops. The daughter's fingers hover over the misshapen dumpling.
  \item \texttt{seg\_shot02\_wrapped\_wrong\_01} (6\,s): From the near shot of the bowl resting beside her hand, the daughter is silent for a beat, head still down, not picking up his words. Her fingertip dabs a little water from the rim of the bowl, then lightly returns to the open seam of the misshapen dumpling, the lens fixed between her hands and the flour, with the father's restrained outline in shallow focus and motionless. A hint of water glints along the dumpling skin's edge; she pinches the split edge tight, her finger paused beside the bowl.
\end{itemize}

\paragraph{Shot~3: \texttt{shot03\_time\_traces} (24\,s).} \emph{Story goal:} Reveal unsorted old objects and the absent mother in the kitchen, shifting the daughter's reason for selling the apartment from practical safety to discomfort with frozen memory.

\begin{figure}[H]
  \centering
  \includegraphics[width=\linewidth]{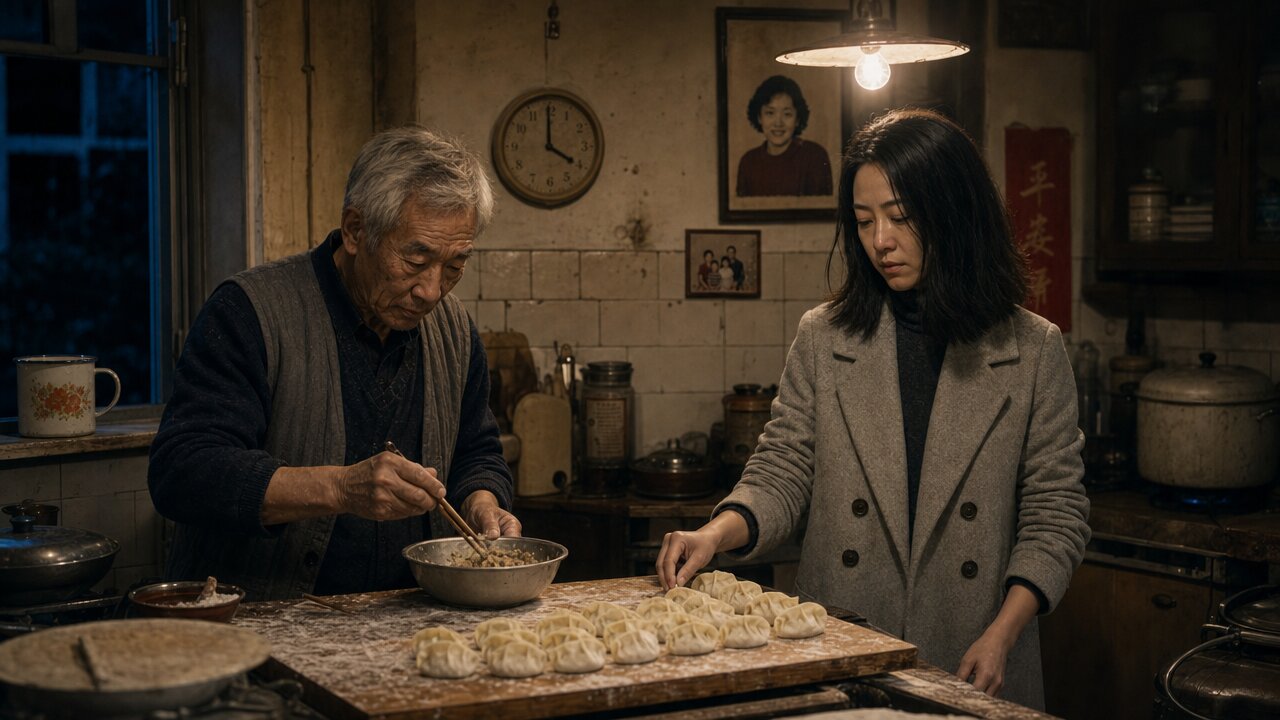}
  \caption{Starting anchor frame \texttt{a03\_time\_traces\_start} for Dumplings shot~3.}
  \label{fig:demo-dumplings-a03_time_traces_start}
\end{figure}

\noindent\emph{Segments.}
\begin{itemize}[leftmargin=1.2em,topsep=2pt,itemsep=2pt,parsep=0pt]
  \item \texttt{seg\_shot03\_time\_traces\_00} (13\,s): From the dumplings and scattered flour at the edge of the counter, the daughter still stands on the right with her head down, her mouth set tight, while the father on the left deliberately reaches around the old enamel cup on the windowsill when fetching filling. The lens pans slowly from their hands to the wall: a faded photo of the absent mother, a clock stopped in the afternoon, and the chipped enamel cup are picked out by the warm yellow lamp and the cool blue window light, like time the kitchen has refused to put away. End-frame: the same composition of old objects and the daughter's lowered gaze.
  \item \texttt{seg\_shot03\_time\_traces\_01} (11\,s): From the framing in which the old photo, the stopped clock and the enamel cup press behind them, the lens pans back to the counter and the daughter's lowered profile. She does not lift her face; she only pushes the wrapped dumplings one by one to the edge of the plate, lining them straighter and straighter, the flour on her hands leaving fine white traces along the edge of the old ceramic plate, as if she were laying out, item by item, the leaks, the steep stairs, and the danger of living alone. The father keeps wrapping with his head down, neither protesting nor looking at her; under the warm yellow lamp the two are close in space yet the air is clearly cold. End-frame: the daughter's tight mouth and the row of neat dumplings.
\end{itemize}

\paragraph{Shot~4: \texttt{shot04\_cold\_voice\_crack} (6\,s).} \emph{Story goal:} Let the argument reach its coldest moment: the daughter says small repairs cannot fix the problem, and the father's retreat appears as a torn dumpling skin.

\begin{figure}[H]
  \centering
  \includegraphics[width=\linewidth]{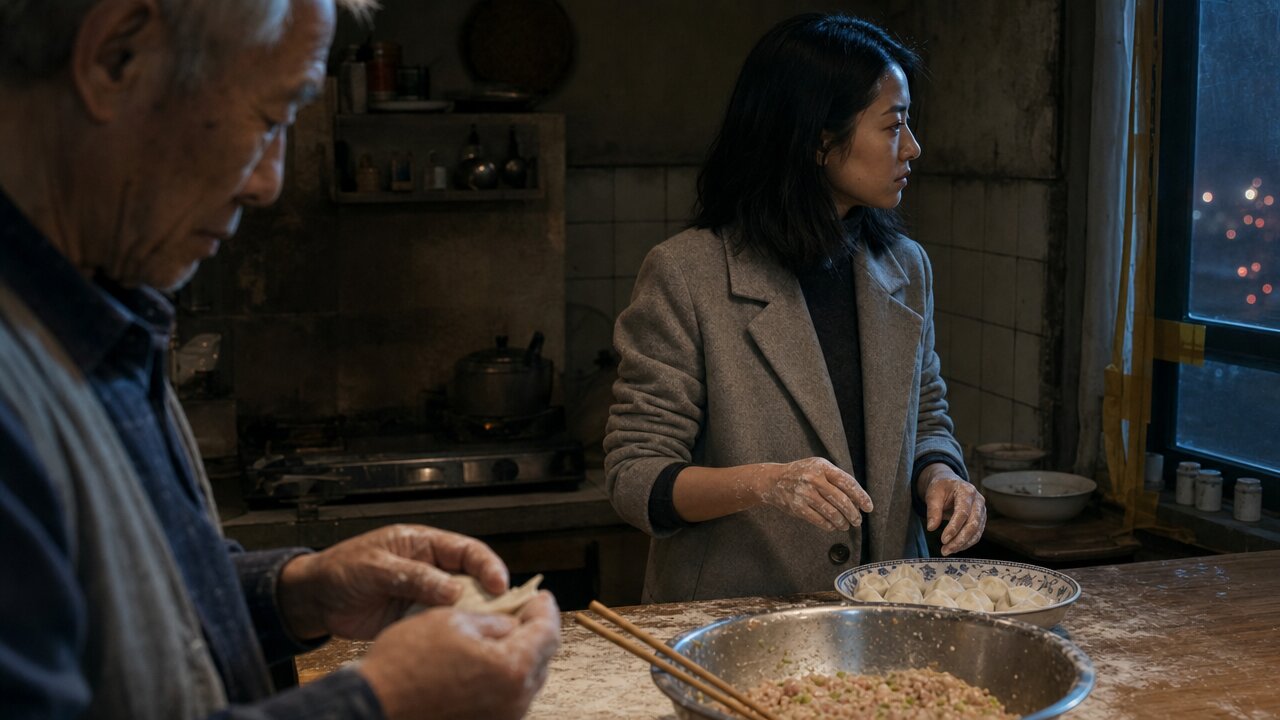}
  \caption{Starting anchor frame \texttt{a04\_cold\_voice\_crack\_start} for Dumplings shot~4.}
  \label{fig:demo-dumplings-a04_cold_voice_crack_start}
\end{figure}

\noindent\emph{Segments.}
\begin{itemize}[leftmargin=1.2em,topsep=2pt,itemsep=2pt,parsep=0pt]
  \item \texttt{seg\_shot04\_cold\_voice\_crack\_00} (6\,s): From the cracked dumpling skin in the father's palm, the handheld lens is pushed very close, so that flour clings to his coarse knuckles and the white fine fissures along the rip stand out. Cool blue window light leaks through the gap in the tape behind the daughter, briefly overpowering the warm yellow lamp, and the air seems to freeze. The father offers no defence; he only loosens his fingers slowly and lays the cracked skin flat in his palm again; the lens calmly racks focus to the daughter's half-lit profile. End-frame: the father lays the dumpling skin flat with his head down, the daughter's hands frozen at the edge of the plate.
\end{itemize}

\paragraph{Shot~5: \texttt{shot05\_mothers\_folds} (18\,s).} \emph{Story goal:} Bring the mother's memory back into the room through the pleat pattern of a dumpling, prompting the daughter's first move from refusal to active mending.

\begin{figure}[H]
  \centering
  \includegraphics[width=\linewidth]{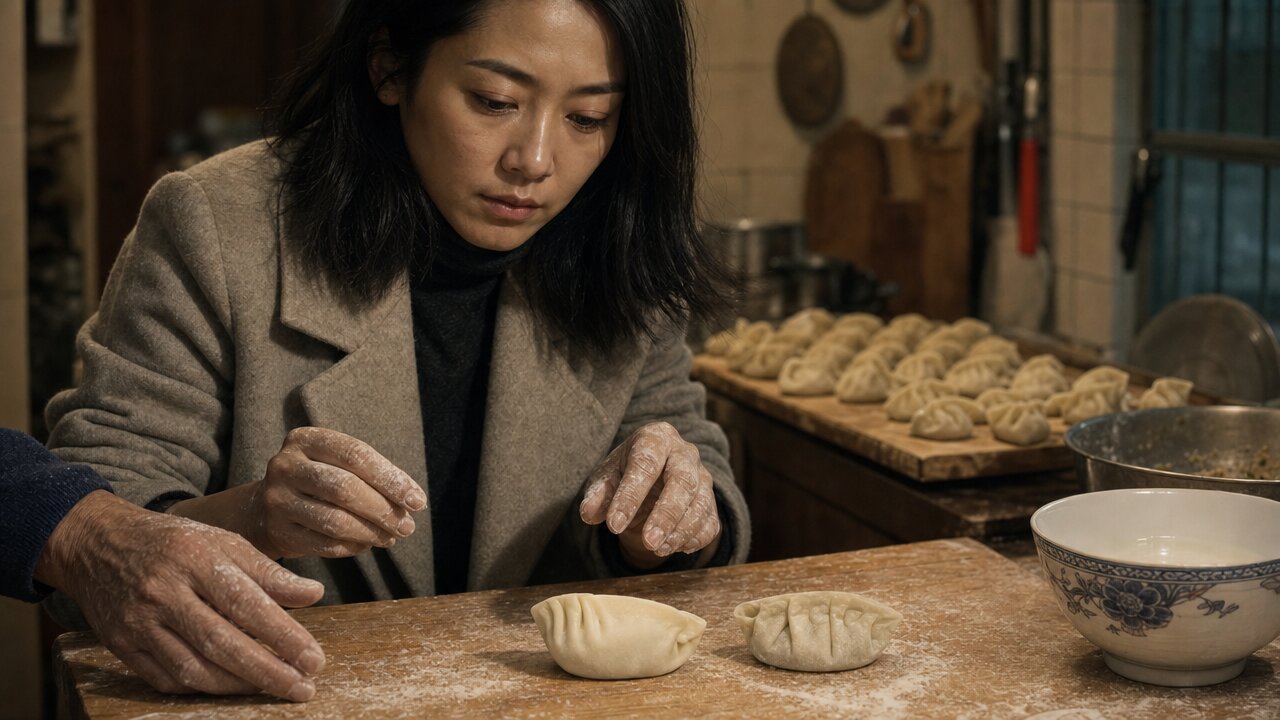}
  \caption{Starting anchor frame \texttt{a05\_mothers\_folds\_start} for Dumplings shot~5.}
  \label{fig:demo-dumplings-a05_mothers_folds_start}
\end{figure}

\noindent\emph{Segments.}
\begin{itemize}[leftmargin=1.2em,topsep=2pt,itemsep=2pt,parsep=0pt]
  \item \texttt{seg\_shot05\_mothers\_folds\_00} (10\,s): From a near shot of the neat dumpling and the misshapen one, the daughter looks down at them, both her cuff and fingertips dusted with flour, while the father's released hand stops at the edge of the frame. The lens pushes very slowly toward the pleats, the focus on the fine pattern (three pleats on the left, a closed edge on the right, a soft press in the middle of the neat dumpling), with the warm yellow light picking out the flour grains and the imperfect handmade edge. The daughter's hand lowers slowly from mid-air; she does not speak; she only takes the earlier misshapen dumpling carefully back between her fingertips while the father's hand pauses at her side. End-frame: the misshapen dumpling held in her fingertips, with the neat dumpling beside it as a reference.
  \item \texttt{seg\_shot05\_mothers\_folds\_01} (8\,s): Carrying on from the daughter holding the misshapen dumpling, she still keeps her head down, flour on her fingertip-pads, and the father's hand at the edge moves still more slowly. The lens stays on her fingers and the pleats: the daughter, following the pleating of the neat dumpling beside her, dips a touch of water on the warped edge, then closes it tight; the pleat does not aim for perfection but clearly shows the order of three left presses, the right closure and the soft middle press. The warm yellow light makes both dumplings' pleats appear like the same old habit lined up side by side. End-frame: she places the mended dumpling beside the neat one; the father's hand pauses still more slowly.
\end{itemize}

\paragraph{Shot~6: \texttt{shot06\_height\_mark\_touch} (7\,s).} \emph{Story goal:} Through the height marks on the door frame, let the daughter touch evidence that she once belonged to this home, so the question of selling begins to loosen.

\begin{figure}[H]
  \centering
  \includegraphics[width=\linewidth]{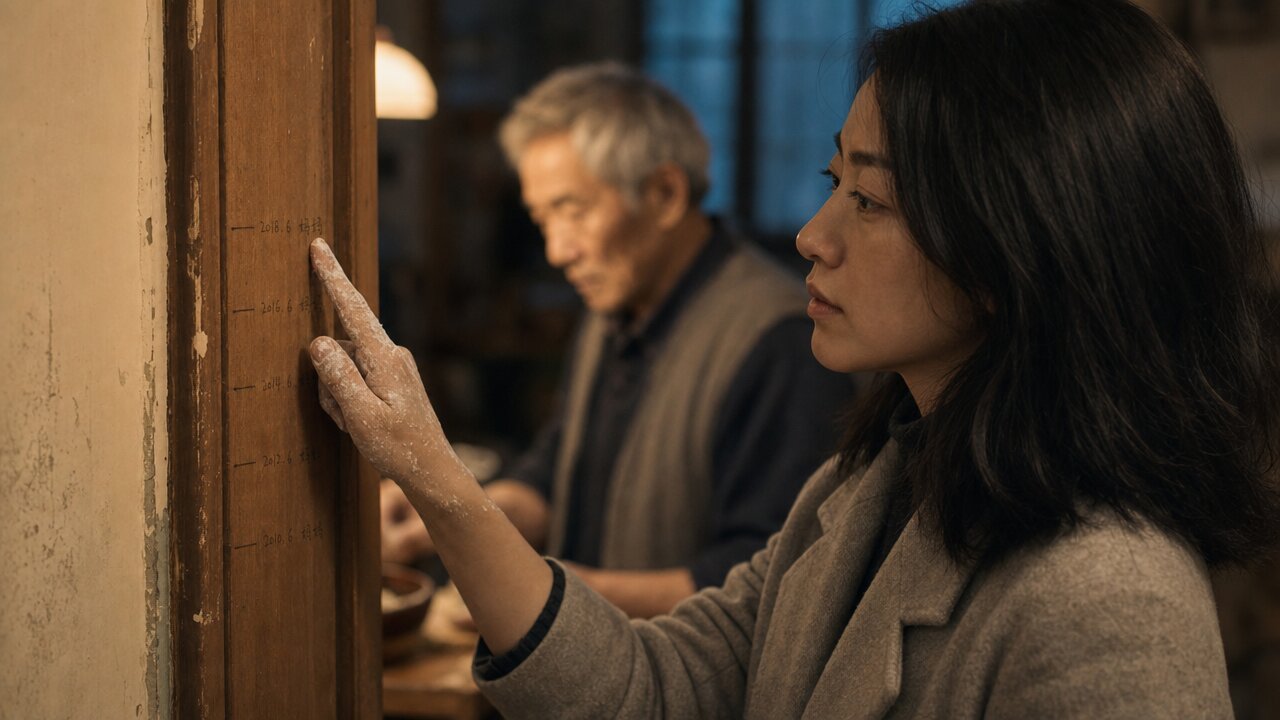}
  \caption{Starting anchor frame \texttt{a06\_height\_mark\_touch\_start} for Dumplings shot~6.}
  \label{fig:demo-dumplings-a06_height_mark_touch_start}
\end{figure}

\noindent\emph{Segments.}
\begin{itemize}[leftmargin=1.2em,topsep=2pt,itemsep=2pt,parsep=0pt]
  \item \texttt{seg\_shot06\_height\_mark\_touch\_00} (7\,s): The daughter's flour-dusted finger stops beside the highest mark on the old door frame, her mother's handwriting just under her fingertip, the beige cuff still rolled up. The lens is still; she only presses her finger pad onto the line very lightly; the warm yellow light picks out the flour grains and the worn paint; an early firework outside makes the glass shake faintly, while the cool blue TV light in the background flashes once and the father stops, blurred, by the counter. End-frame: she leaves a faint flour print beside the highest mark, her profile quiet, the father still standing silently behind her.
\end{itemize}

\paragraph{Shot~7: \texttt{shot07\_boiling\_together} (22\,s).} \emph{Story goal:} Let the father and daughter shift from standing side by side in silence to actually co-operating in the concrete acts of boiling the dumplings.

\begin{figure}[H]
  \centering
  \includegraphics[width=\linewidth]{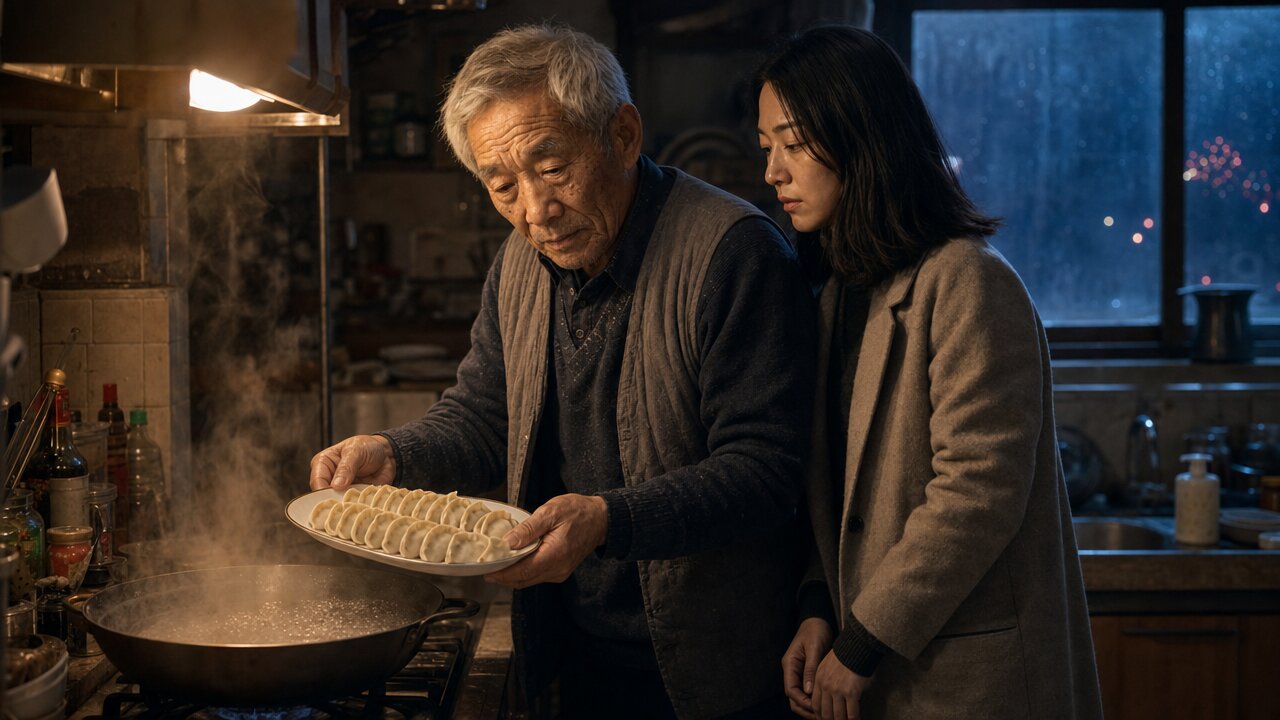}
  \caption{Starting anchor frame \texttt{a07\_boiling\_together\_start} for Dumplings shot~7.}
  \label{fig:demo-dumplings-a07_boiling_together_start}
\end{figure}

\noindent\emph{Segments.}
\begin{itemize}[leftmargin=1.2em,topsep=2pt,itemsep=2pt,parsep=0pt]
  \item \texttt{seg\_shot07\_boiling\_together\_00} (12\,s): From the side of the stove with the boiling water and the raw dumplings, the father still in his dark blue sweater and grey vest holds the plate carefully and slides the dumplings along the rim of the pot into the water. The lens orbits the stove in a small arc; the focus is on the white fat dumplings dropping into the rolling water one by one, the surface gently parted, the steam thickening under the warm yellow stove lamp, while the taped window lets in the cold blue night. The daughter stands closer, beige cuff rolled up, her free hand reaching slowly toward the handle of the slotted spoon. End-frame: a few dumplings have already scattered in the pot, the father holds an almost-empty plate, and the daughter's hand pauses beside the slotted spoon.
  \item \texttt{seg\_shot07\_boiling\_together\_01} (10\,s): From the daughter's hand pausing beside the slotted spoon, she does not speak; she takes the spoon up and gently moves the dumplings in the water along the bottom of the pot in small, steady motions. The lens keeps orbiting the stove slowly; the steam rises and softens both the wrinkles on the father's face and the tense profile of the daughter. The father sets down the near-empty plate and reaches over to turn the heat down; she follows by tilting the lid against the rim of the pot, letting a thin column of white steam escape from one side. End-frame: she holds the slotted spoon by the rim of the pot, the father's hand has just left the dial, the lid is tilted, and the steam is rising.
\end{itemize}

\paragraph{Shot~8: \texttt{shot08\_quiet\_aftertaste} (16\,s).} \emph{Story goal:} With the daughter proposing to repair the kitchen window first instead of selling, deliver a partial, not-yet-complete reconciliation as the closing aftertaste.

\begin{figure}[H]
  \centering
  \includegraphics[width=\linewidth]{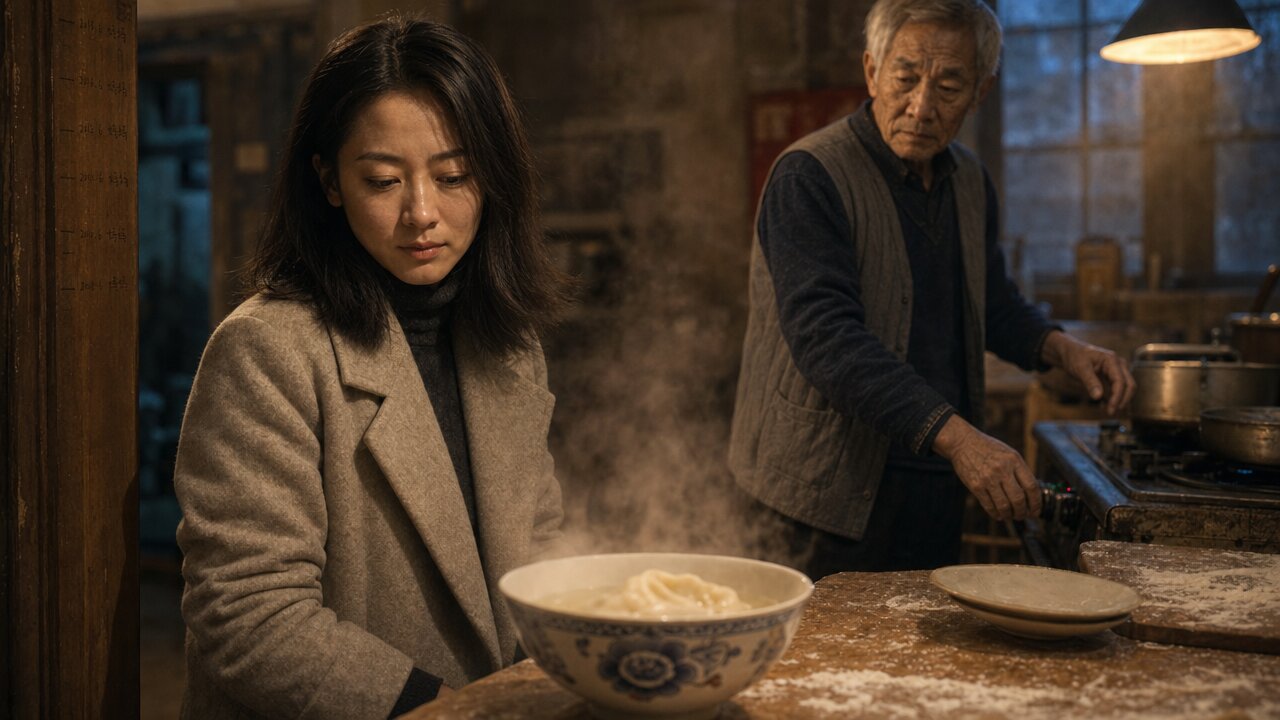}
  \caption{Starting anchor frame \texttt{a08\_quiet\_aftertaste\_start} for Dumplings shot~8.}
  \label{fig:demo-dumplings-a08_quiet_aftertaste_start}
\end{figure}

\noindent\emph{Segments.}
\begin{itemize}[leftmargin=1.2em,topsep=2pt,itemsep=2pt,parsep=0pt]
  \item \texttt{seg\_shot08\_quiet\_aftertaste\_00} (9\,s): From the dumpling that has been re-pleated, lying in the bowl, the white steam of the hot broth softly veils the rim. The daughter looks down at it, her expression loosened though still restrained. The lens dollies out very slowly. The father, beside the stove, hears her say softly that next year they could repair the kitchen window first; he does not press for more; he only lowers the hand resting near the gas dial a little; the warm yellow stove light passes through the steam, the cool blue window light has almost retreated. End-frame: the father's hand close to the dial, the daughter still bent over the bowl, white steam horizontal between them.
  \item \texttt{seg\_shot08\_quiet\_aftertaste\_01} (7\,s): From the silence of the father's hand at the dial and the daughter holding the bowl, the lens keeps pulling back slowly; the dumpling in the bowl and the rising steam recede to the edge of the foreground. They do not embrace and do not explain; they simply stand naturally side by side at the stove, while the old counter, the empty plate with leftover flour, the taped window and the door-frame height marks enter the frame one by one; the steam softens the cracks in the old kitchen, and low-saturation warm yellow wraps the two of them again. End-frame: a wide shot of the old kitchen, with father and daughter standing side by side at the stove and the steam from the pot rising quietly.
\end{itemize}
 
\section{Video Failure Cases}
\label{app:video-failure-cases}

This appendix records video-level failure modes observed in the generated demos and clarifies what ReCA can and cannot repair. These cases are not failures of state allocation alone: several originate inside the frozen video generator or the provider pipeline, so an inference-time controller can only retry or recondition the call.

\paragraph{Backbone-limited visual failures.}
ReCA cannot repair low-level generator failures. ReCA is an inference-time controller that allocates language-side state across a frozen generator's calls; it cannot fix failures that originate inside the generator itself, such as morphological artefacts, blurred motion, identity-preservation breakdowns within a single $5$-second call, or content-safety rejections imposed by the provider's pipeline. When the underlying backbone produces an unusable segment, ReCA's local repair set (\textsc{RepackPrompt}, \textsc{ReanchorState}, \textsc{RegenerateUnit}, \textsc{SplitUnit}) can re-issue the call with adjusted state, but the per-call quality ceiling is set by the generator, not by ReCA.

\begin{figure}[H]
  \centering
  \begin{subfigure}[t]{0.24\linewidth}
    \centering
    \includegraphics[width=\linewidth]{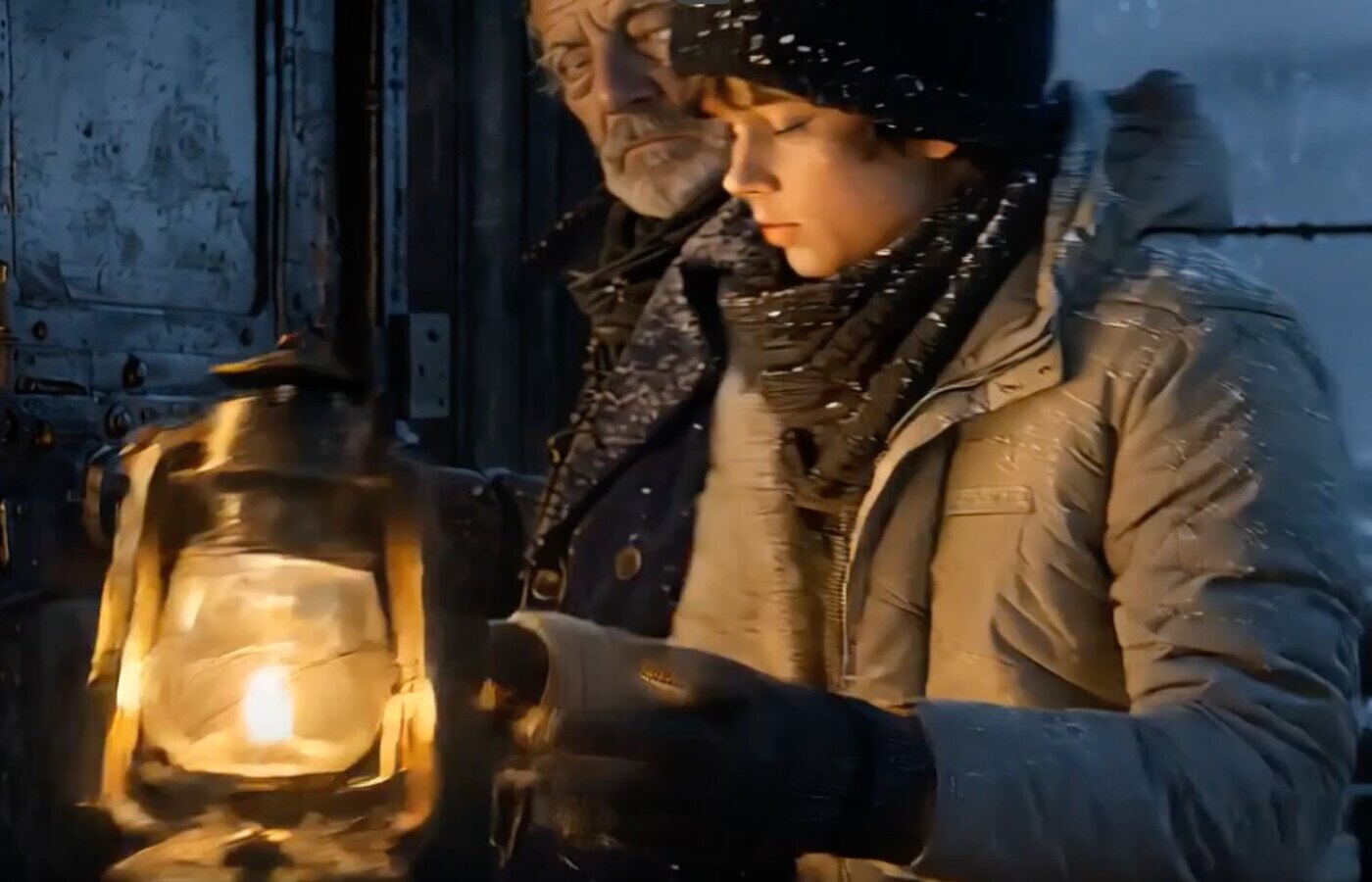}
    \caption{\textbf{Action-state corruption.} A lamp moves by itself rather than obeying the scene dynamics.}
  \end{subfigure}
  \hfill
  \begin{subfigure}[t]{0.24\linewidth}
    \centering
    \includegraphics[width=\linewidth]{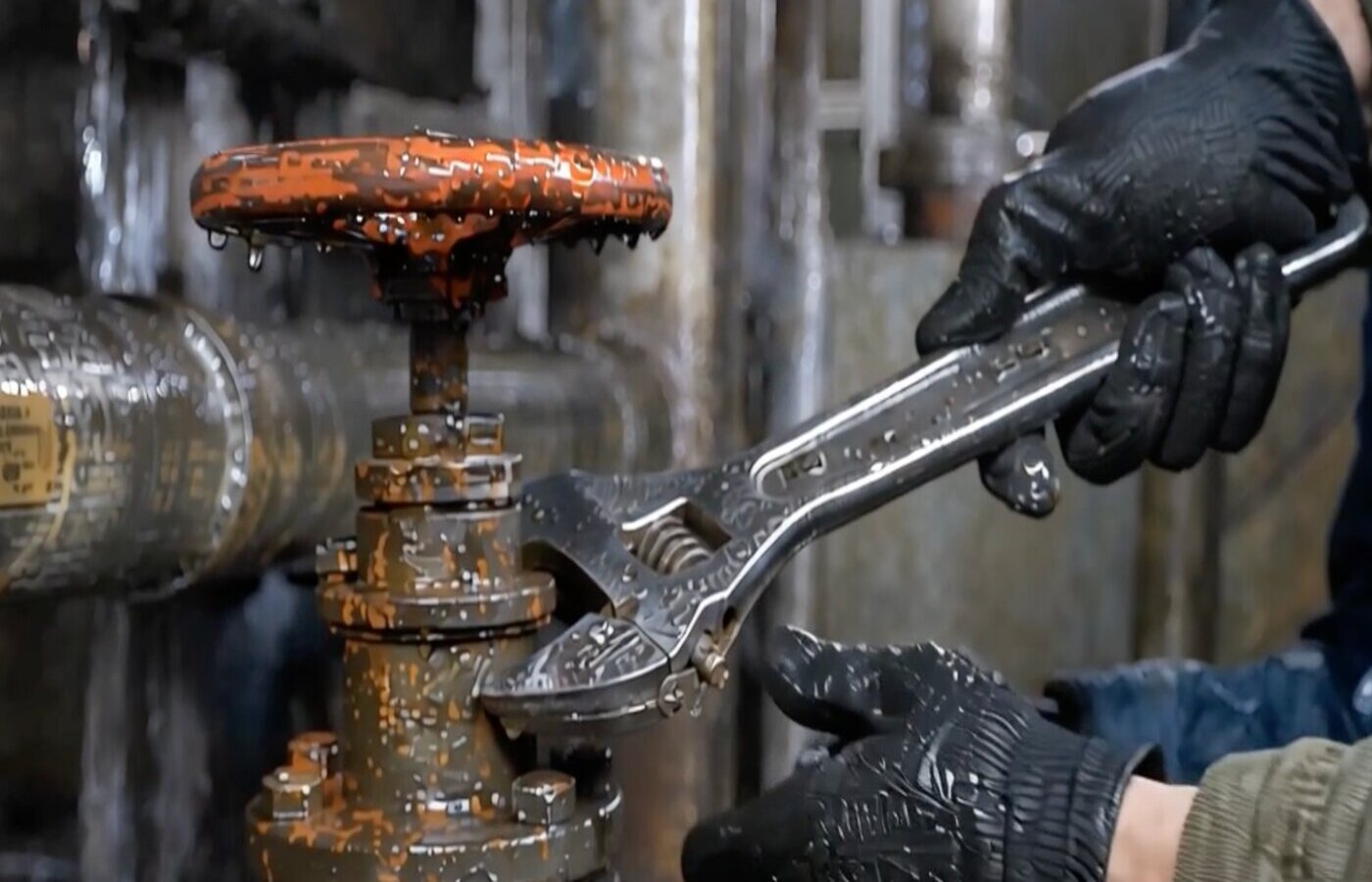}
    \caption{\textbf{Action-state corruption.} The wrench is applied to the wrong contact point.}
  \end{subfigure}
  \hfill
  \begin{subfigure}[t]{0.24\linewidth}
    \centering
    \includegraphics[width=\linewidth]{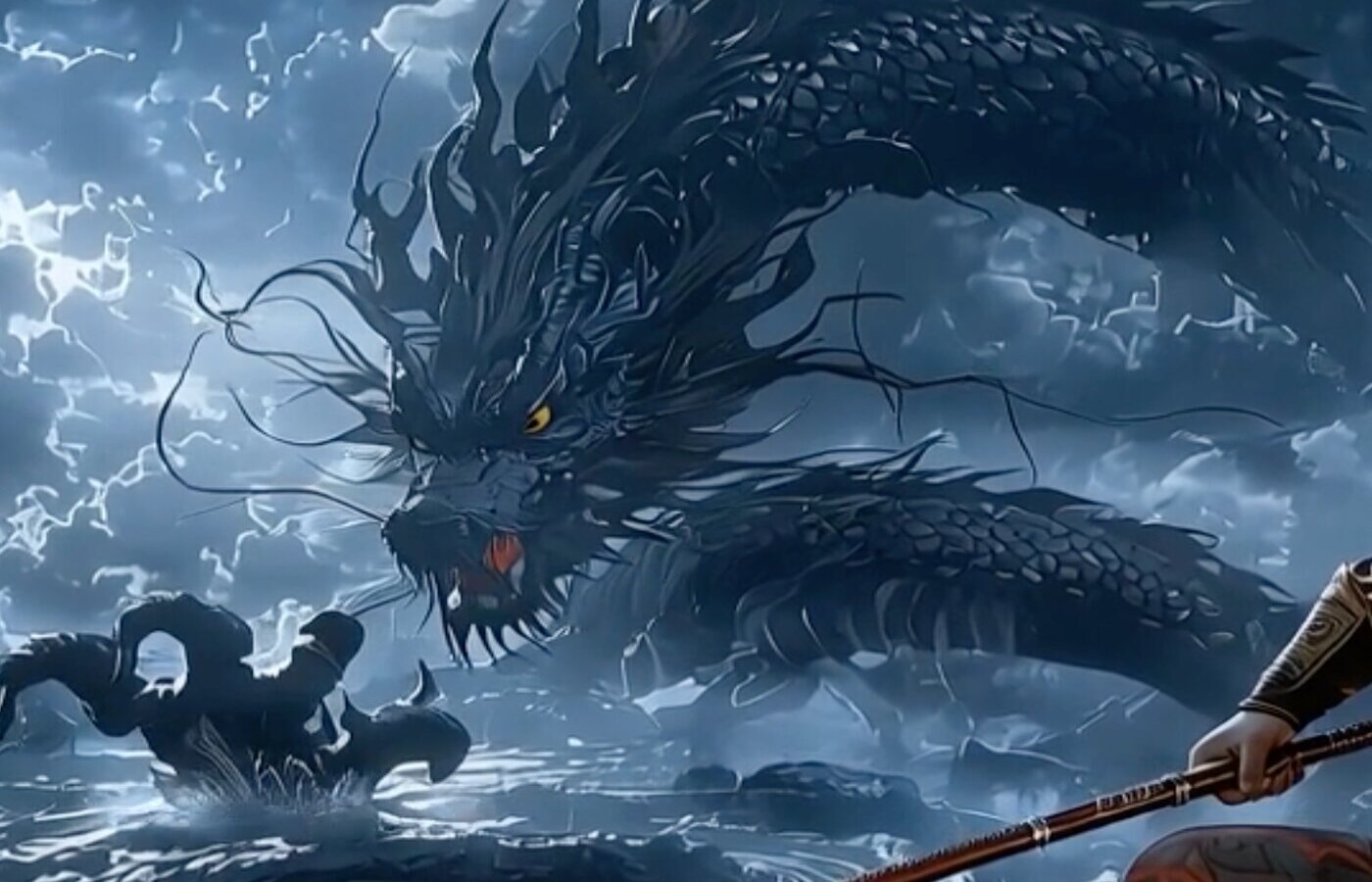}
    \caption{\textbf{Morphology and motion artefact.} The dragon claw emerges from an anatomically implausible location.}
  \end{subfigure}
  \hfill
  \begin{subfigure}[t]{0.24\linewidth}
    \centering
    \includegraphics[width=\linewidth]{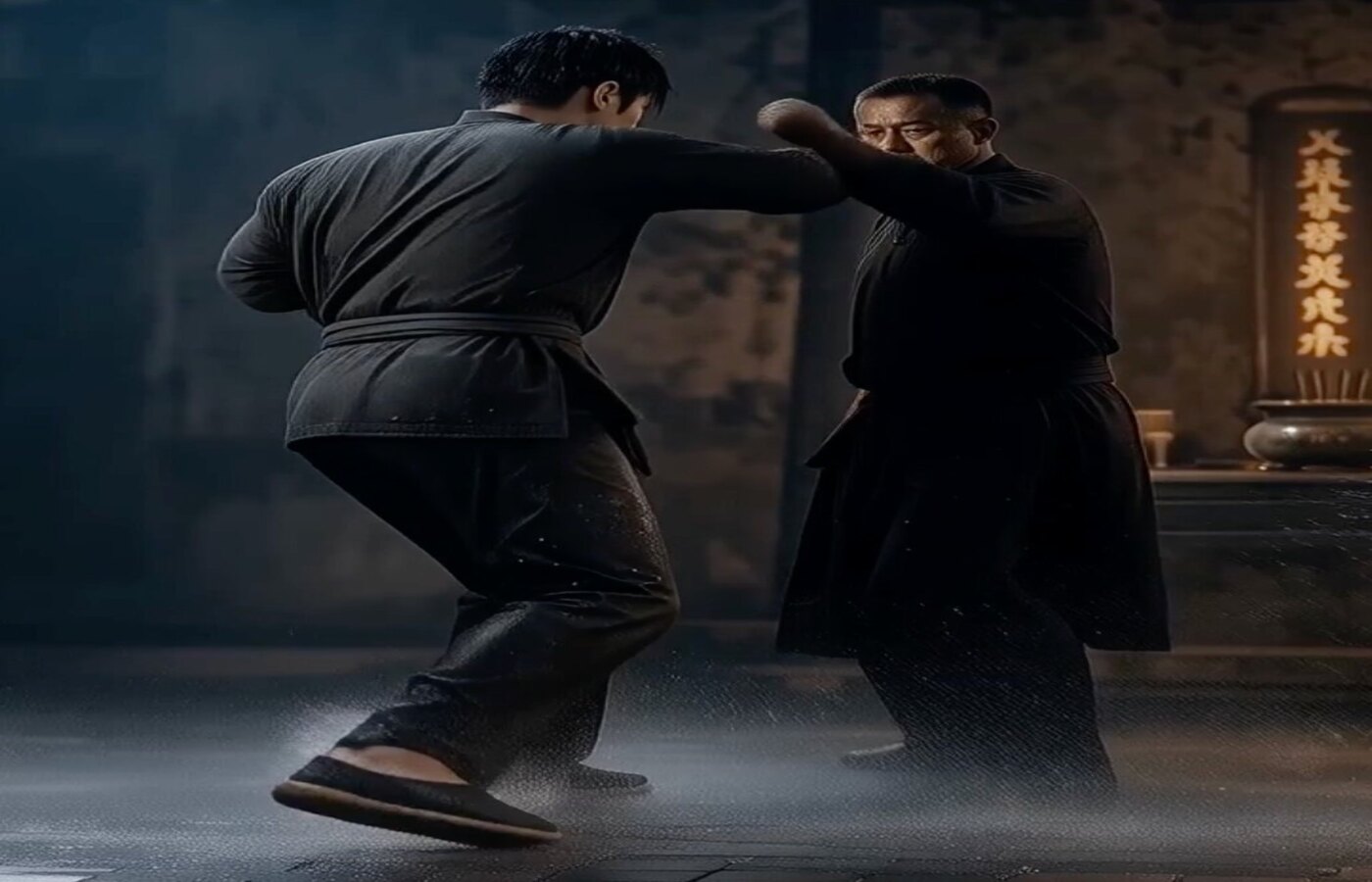}
    \caption{\textbf{Morphology and motion artefact.} The hand is not rendered even though the action requires it.}
  \end{subfigure}
  \caption{\textbf{Representative backbone-limited failure cases.} These examples illustrate two observed categories: action-state corruption and morphology/motion artefacts. ReCA can retry the unit with adjusted language-side state, re-anchor the next call, or split the unit, but the local visual failure itself is bounded by the generator's per-call rendering quality.}
  \label{fig:bad-frame-examples}
\end{figure}

\paragraph{Observed categories.}
We group the residual failures into four categories:
\begin{itemize}[leftmargin=1.2em,topsep=2pt,itemsep=2pt,parsep=0pt]
  \item \textbf{Morphology and motion artefacts.} Hands, faces, weapons, or fast-moving limbs can deform inside a single generated segment. ReCA can shorten or split the unit, but it cannot guarantee that the next call will render the anatomy or motion cleanly.
  \item \textbf{Within-call identity drift.} A character may change clothing, facial structure, scale, or pose identity during one $5$-second generation. Because this happens before the controller receives a refreshed state, the repair options are limited to re-prompting, re-anchoring, or regenerating the unit.
  \item \textbf{Action-state corruption.} The generator may render the wrong contact relation, weapon orientation, prop state, or interaction physics even when the prompt specifies the intended state. ReCA can make the state more explicit in the next prompt, but the local physical plausibility still depends on the generator.
  \item \textbf{Provider-side rejection.} Some calls fail before visual output is produced because the provider's content-safety pipeline rejects the request. The controller can alter wording, split the unit, or regenerate a neighbouring state, but it cannot override the provider policy.
\end{itemize}

\paragraph{Why the distinction matters.}
The main empirical claim is about allocating usable context across prompt, time, and planning axes. A failed segment can therefore have two different causes: the controller may have passed the wrong state, or the generator may have failed despite receiving the right state. ReCA is designed for the first case. For the second case, it supplies retry mechanisms and narrower prompts, but it does not change the frozen model's rendering distribution.
 
\section{Limitations}
\label{app:limitations}

\paragraph{ReCA cannot repair low-level generator failures.}
ReCA is an inference-time controller that allocates language-side state across a frozen generator's calls; it cannot fix failures that originate inside the generator itself, such as morphological artefacts, blurred motion, identity-preservation breakdowns within a single $5$-second call, or content-safety rejections imposed by the provider's pipeline. When the underlying backbone produces an unusable segment, ReCA's local repair set ($\textsc{RepackPrompt}$, $\textsc{ReanchorState}$, $\textsc{RegenerateUnit}$, $\textsc{SplitUnit}$) can re-issue the call with adjusted state, but the per-call quality ceiling is set by the generator, not by ReCA.

\paragraph{Long-video outputs are expensive to evaluate and difficult to compare fairly.}
Each MSVE-Bench instance produces a $3$--$5$~minute clip whose evaluation requires segmenting, aligning to the source clip sequence, and answering source-grounded VLM problems on every segment, which is orders of magnitude more expensive than scoring a $5$-second clip with a single VLM call. Cross-method comparison is further complicated by per-method differences in shot count, shot duration, and prompt rewriting, all of which interact with the segmentation gate and the coverage-gated coherence rule; we therefore cap the user study at $80$ videos and constrain automatic-metric basket size to $5$ scorers (Section~\ref{sec:exp-main}).

\paragraph{Parallel execution benefits depend on API rate limits and dependency structure.}
The wall-clock cost model in Section~\ref{sec:method-parallel}, $\Theta(\lceil N/W\rceil\,\tau_G + D\,\tau_{\mathrm{op}})$, makes the achievable speed-up explicitly dependent on the available concurrency $W$ and the longest state-threaded chain $D$ in the dependency map $\mathrm{prev}$. In practice, $W$ is set by the provider's per-account rate limit on simultaneous T2V requests and the I2V handoff structure of the schedule fixes $D$, so a plan with mostly state-threaded siblings approaches the sequential floor $\Theta(D\,\tau_G)$ and gains little from larger $W$; conversely, a plan dominated by independent siblings benefits linearly with $W$ until the rate limit saturates.

\section{LLM Usage Disclosure}
\label{app:llm-usage-disclosure}

A large language model (OpenAI GPT 5.5) was used solely for grammar and spelling correction; all resulting text was reviewed by the authors.

\appendixtocrecordingfalse

\end{document}